\def\year{2021}\relax
\documentclass[letterpaper]{article} 
\usepackage{aaai21}  
\usepackage{times}  
\usepackage{helvet} 
\usepackage{courier}  
\usepackage[hyphens]{url}  
\usepackage{graphicx} 
\usepackage{natbib}  
\usepackage{caption} 
\usepackage{multirow,multicol}
\usepackage{xcolor}

\usepackage{amsthm}
\newtheorem{theorem}{Theorem}

\newtheorem{dfn}{Definition}

\usepackage{booktabs}
\usepackage{enumitem}
\usepackage{graphics}
\usepackage{subcaption}
\usepackage{paralist}
\usepackage{xspace}
\usepackage{amsfonts}
\usepackage{amsmath}

\usepackage{hyperref}

\newcommand*{\scale}[2][4]{\scalebox{#1}{$#2$}}

\title{Data Augmentation for Graph Neural Networks}
\author {
    Tong Zhao\textsuperscript{\dag}\footnote{This work was done when the author was on internship at Snap Inc.},
    Yozen Liu\textsuperscript{\ddag},
    Leonardo Neves\textsuperscript{\ddag},
    Oliver Woodford\textsuperscript{\ddag},
    Meng Jiang\textsuperscript{\dag},
    Neil Shah\textsuperscript{\ddag}
    \\
}
\affiliations {
    \textsuperscript{\dag} University of Notre Dame, Notre Dame, IN 46556 \\
    \textsuperscript{\ddag} Snap Inc., Santa Monica, CA 90405 \\
    \{tzhao2, mjiang2\}@nd.edu, \{yliu2, lneves, oliver.woodford, nshah\}@snap.com
}
\begin{document}

\maketitle

\newcommand{\methodtwo}{\textsc{GAug-M}\xspace}
\newcommand{\method}{\textsc{GAug-O}\xspace}
\newcommand{\methodshared}{\textsc{GAug}\xspace}

\newcommand{\cora}{\textsc{Cora}\xspace}
\newcommand{\citeseer}{\textsc{Citeseer}\xspace}
\newcommand{\pubmed}{\textsc{Pubmed}\xspace}
\newcommand{\ppi}{\textsc{PPI}\xspace}
\newcommand{\amazon}{\textsc{Amazon}\xspace}
\newcommand{\blogc}{\textsc{BlogCatalog}\xspace}
\newcommand{\flickr}{\textsc{Flickr}\xspace}
\newcommand{\airusa}{\textsc{Air-USA}\xspace}
\newcommand{\ogbn}{\textsc{OGBn-Arxiv}\xspace}

\newcommand{\gcn}{\textsc{GCN}\xspace}
\newcommand{\gsage}{\textsc{GSAGE}\xspace}
\newcommand{\gat}{\textsc{GAT}\xspace}
\newcommand{\jknet}{\textsc{JK-Net}\xspace}

\newcommand{\dropedge}{\textsc{DropEdge}\xspace}
\newcommand{\adaedge}{\textsc{AdaEdge}\xspace}
\newcommand{\bgcn}{\textsc{BGCN}\xspace}

\newcommand\neil[1]{\textcolor{red}{[Neil: #1]}}
\newcommand\ns[1]{\textcolor{red}{[Neil: #1]}}
\newcommand\tong[1]{\textcolor{cyan}{[Tong: #1]}}
\newcommand{\yl}[1]{\textcolor{purple}{[YL: #1]}}

\begin{abstract}
Data augmentation has been widely used to improve generalizability of machine learning models.  However, comparatively little work studies data augmentation for graphs.  This is largely due to the complex, non-Euclidean structure of graphs, which limits possible manipulation operations. Augmentation operations commonly used in vision and language have no analogs for graphs.  Our work studies graph data augmentation for graph neural networks (GNNs) in the context of improving semi-supervised node-classification.  We discuss practical and theoretical motivations, considerations and strategies for graph data augmentation.  Our work shows that neural edge predictors can effectively encode class-homophilic structure to promote intra-class edges and demote inter-class edges in given graph structure, and our main contribution introduces the \methodshared graph data augmentation framework, which leverages these insights to improve performance in GNN-based node classification via edge prediction. Extensive experiments on multiple benchmarks show that augmentation via \methodshared improves performance across GNN architectures and datasets.

\end{abstract}

\section{Introduction}
\label{sec:introduction}
Data driven inference has received a significant boost in generalization capability and performance improvement in recent years from data augmentation techniques. These methods increase the amount of training data available by creating plausible variations of existing data without additional ground-truth labels, and have seen widespread adoption in fields such as computer vision (CV)~\cite{devries2017improved,cubuk2019autoaugment,zhao2019data,ho2019population}, and natural language processing (NLP)~\cite{fadaee2017data,csahin2019data}. Such augmentations allow inference engines to learn to generalize better across those variations and attend to signal over noise.  At the same time, graph neural networks (GNNs)~\cite{hamilton2017inductive,kipf2016semi,velivckovic2017graph,xu2018powerful,zhang2019heterogeneous,chen2018fastgcn,wu2019comprehensive,zhang2018deep,xu2018representation} have emerged as a rising approach for data-driven inference on graphs, achieving promising results on tasks such as node classification, link prediction and graph representation learning.

Despite the complementary nature of GNNs and data augmentation, few works present strategies for combining the two. One major obstacle is that, in contrast to other data, where structure is encoded by position, the structure of graphs is encoded by node connectivity, which is irregular. The hand-crafted, structured, data augmentation operations used frequently in CV and NLP therefore cannot be applied. Furthermore, this irregularity does not lend itself to easily defining new augmentation strategies. The most obvious approaches involve adding or removing nodes or edges.  For node classification tasks, adding nodes poses challenges in labeling and imputing features and connectivity of new nodes, while removing nodes simply reduces the data available. Thus, edge addition and removal appears the best augmentation strategy for graphs. But the question remains, \emph{which} edges to change.

\newcommand{\etal}{et al.}
Three relevant approaches have recently been proposed. \dropedge~\cite{rong2019dropedge} randomly removes a fraction of graph edges before each training epoch, in an approach reminiscent of dropout \cite{srivastava2014dropout}. This, in principle, robustifies test-time inference, but cannot benefit from added edges. In approaches more akin to denoising or pre-filtering, \adaedge~\cite{chen2019measuring} iteratively add (remove) edges between nodes predicted to have the same (different) labels with high confidence in the modified graph.  This ad-hoc, two-stage approach improves inference in general, but is prone to error propagation and greatly depends on training size.  Similarly, \bgcn~\cite{zhang2019bayesian} iteratively trains an assortative mixed membership stochastic block model with predictions of \gcn to produce multiple denoised graphs, and ensembles results from multiple \gcn{s}.  
\bgcn also bears the risk of error propagation. 

\begin{figure*}[t]
    \centering
        \begin{subfigure}[b]{.2\linewidth}
         \includegraphics[width=\textwidth]{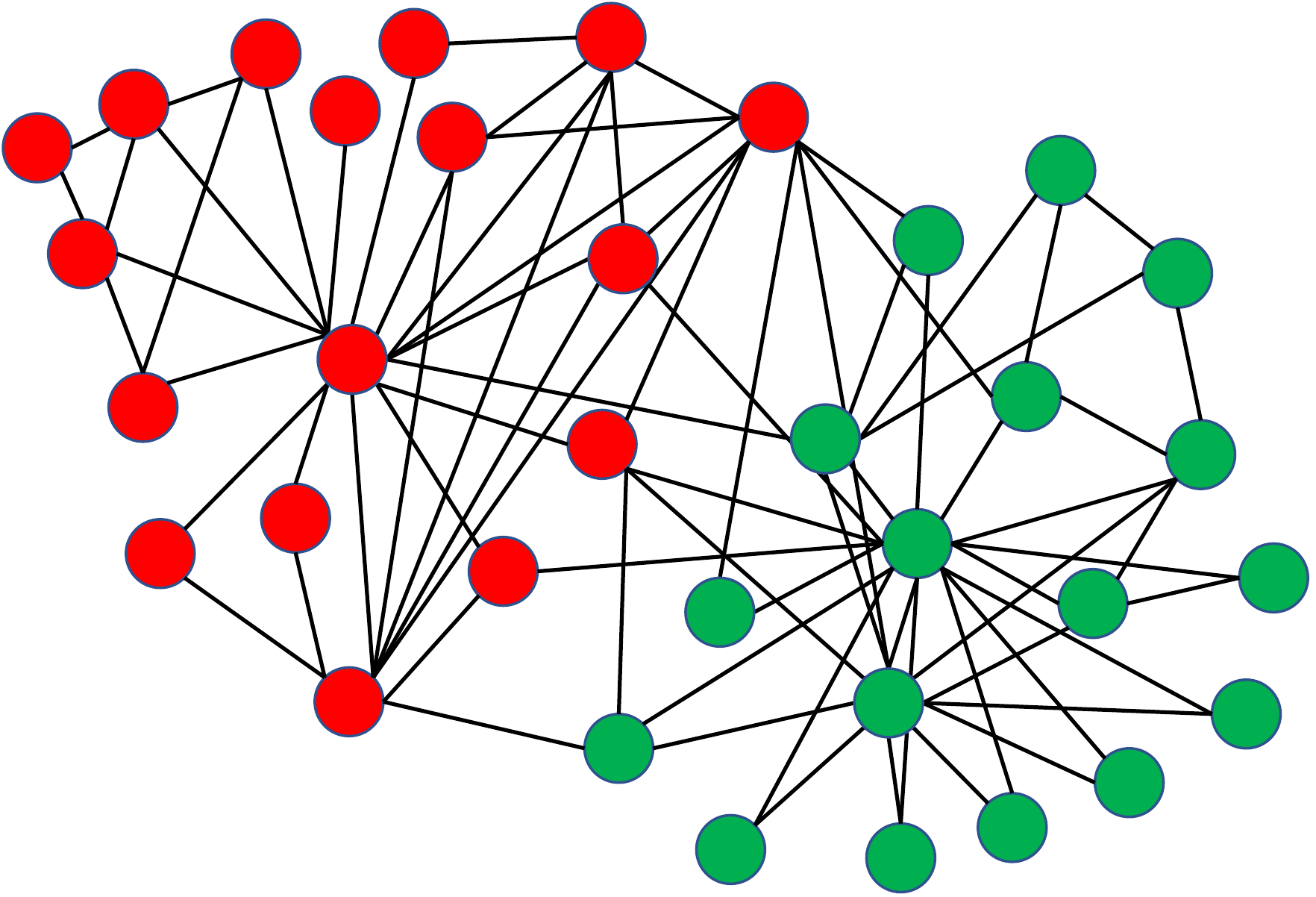}  
         \caption{Original graph. \\ $O: 92.4$ F1}\label{fig:tiny_orig}
        \end{subfigure}\quad
        \begin{subfigure}[b]{.2\linewidth}
         \includegraphics[width=\textwidth]{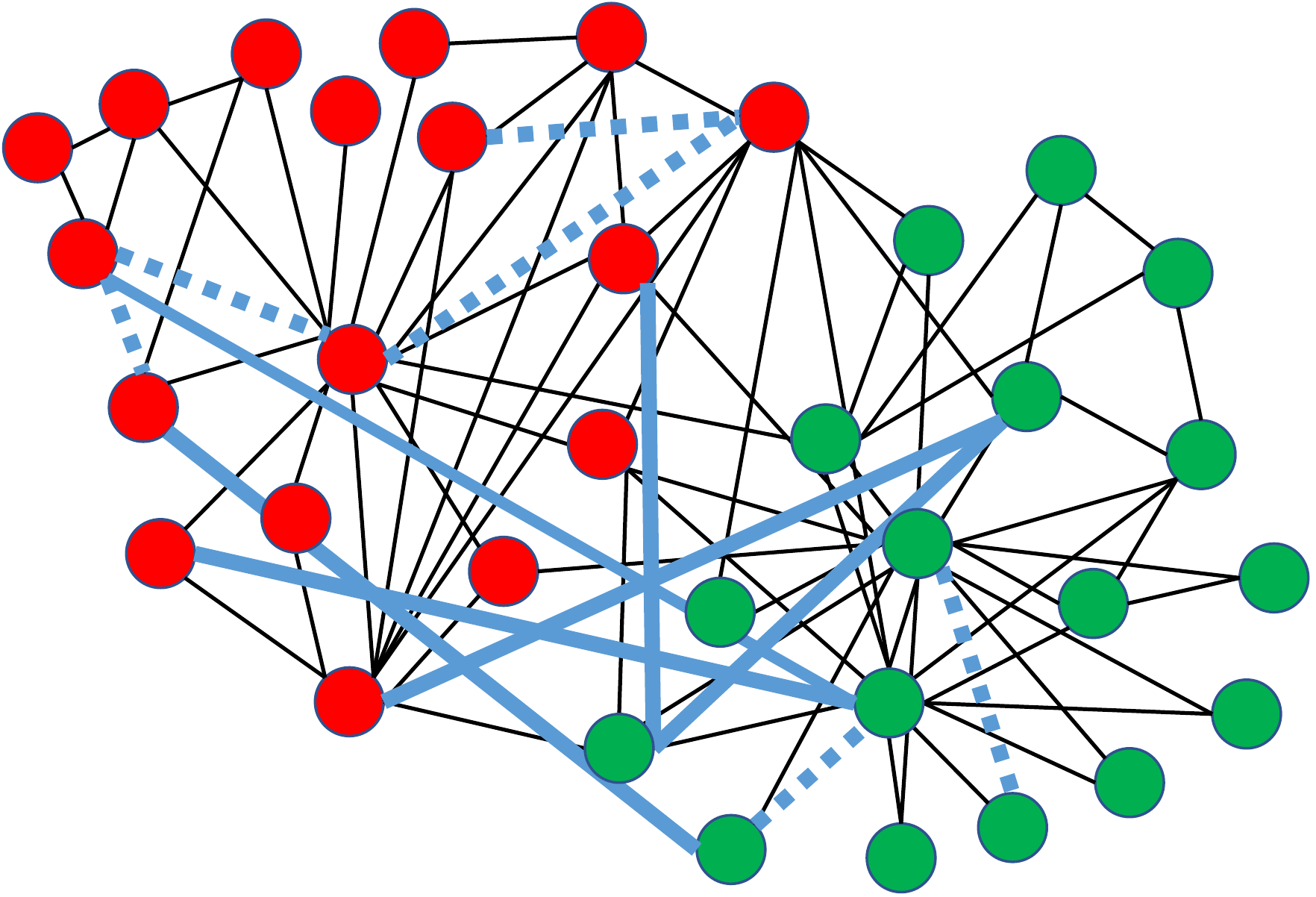}
            \caption{Random mod.\\ $M: 90.4$, $O: 91.0$ F1}\label{fig:tiny_rand}
        \end{subfigure}\quad
        \begin{subfigure}[b]{.2\linewidth}
         \includegraphics[width=\textwidth]{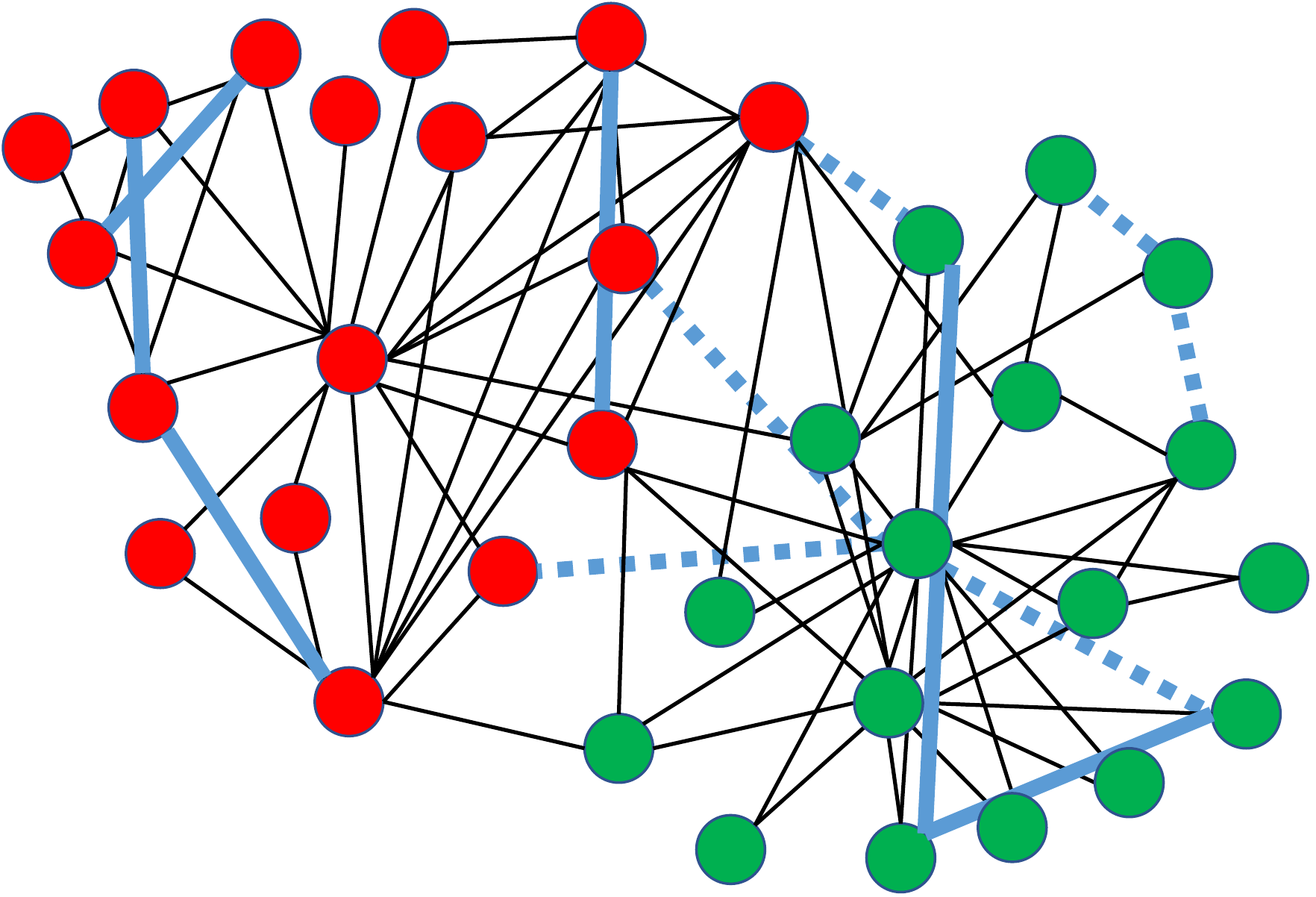}
            \caption{Proposed \methodshared mod.\\ $M: 95.7$, $O: 94.3$ F1}\label{fig:tiny_gaugm}
        \end{subfigure}\quad
        \begin{subfigure}[b]{.2\linewidth}
         \includegraphics[width=\textwidth]{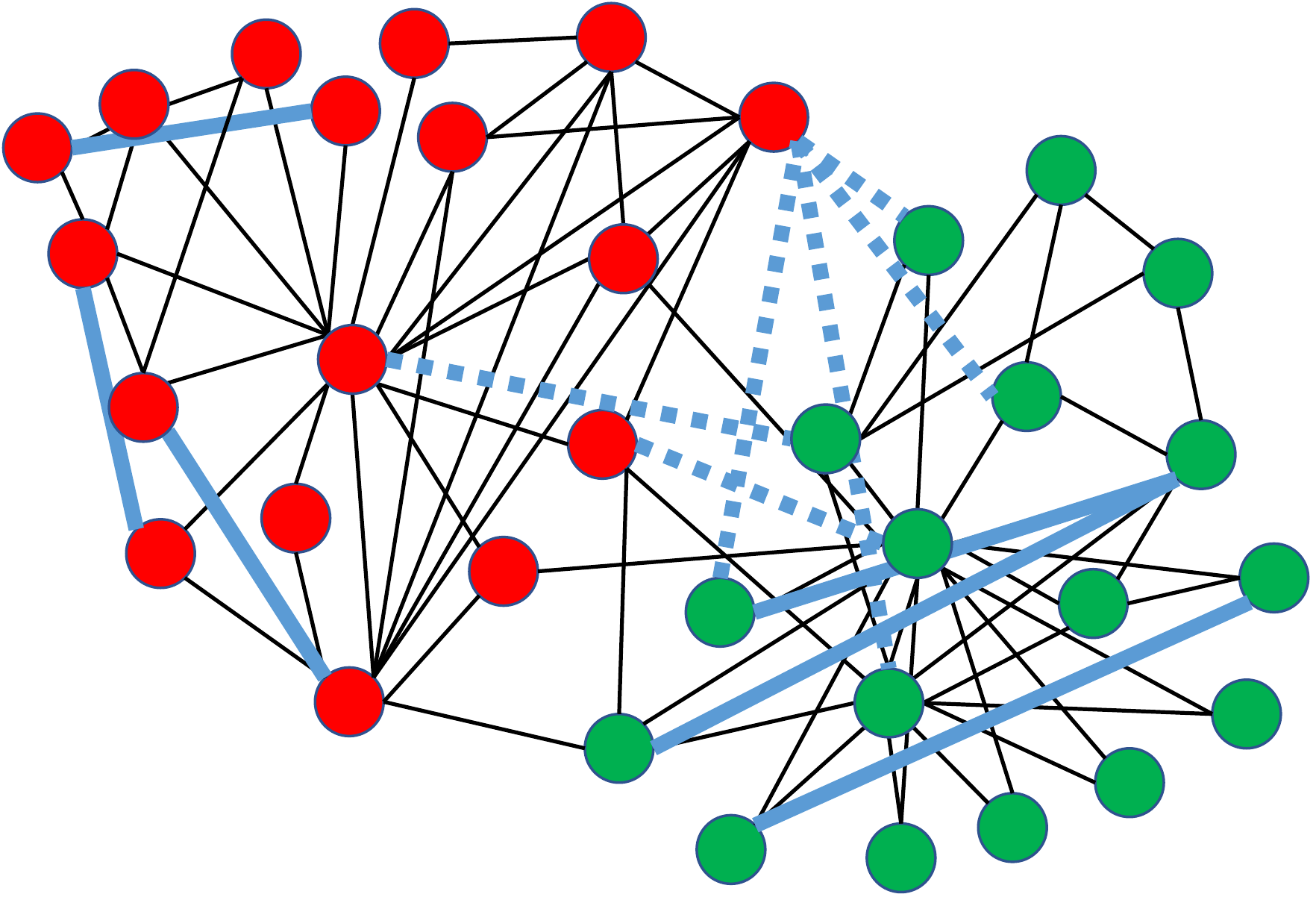}
            \caption{Omniscient mod.\\ $M: 98.6$, $O: 95.6$ F1}\label{fig:tiny_ideal}
        \end{subfigure}
        \caption{\label{fig:tiny} \gcn performance (test micro-F1) on the original Zachary's Karate Club graph in (a), and three augmented graph variants in (b-d), evaluated on both original ($O$) and modified ($M$) graph settings. Black, solid-blue, dashed-blue edges denote original graph connectivity, newly added, and removed edges respectively.  While random graph modification (b) hurts performance, our proposed \methodshared augmentation approaches (c) demonstrate significant relative performance improvements, narrowing the gap to omniscient, class-aware modifications (d).}
\end{figure*}

\noindent \textbf{Present work.} Our work studies new techniques for graph data augmentation to improve node classification.  Section~\ref{sec:edgemanip} introduces motivations and considerations in augmentation via edge manipulation.  Specifically, we discuss how facilitating message passing by removing ``noisy'' edges and adding ``missing'' edges that could exist in the original graph can benefit GNN performance, and its relation to intra-class and inter-class edges.  Figure \ref{fig:tiny} demonstrates, on a toy dataset (a), that while randomly modifying edges (b) can lead to lower test-time accuracy,  strategically choosing ideal edges to add or remove given (unrealistic) omniscience of node class labels (d) can substantially improve it.

Armed with this insight, Section~\ref{sec:proposedmethod} presents our major contribution: the proposed \methodshared framework for graph data augmentation. We show that neural edge predictors like GAE \cite{kipf2016variational} are able to latently learn class-homophilic tendencies in existent edges that are improbable, and nonexistent edges that are probable.  \methodshared leverages this insight in two approaches, \methodtwo and \method, which tackle augmentation in settings where edge manipulation is and is not feasible at inference time.  \methodtwo uses an edge prediction module to fundamentally modify an input graph for future training and inference operations, whereas \method learns to generate plausible edge augmentations for an input graph, which helps node classification 
without any modification at inference time.  In essence, our work tackles the problem of the inherent indeterminate nature of graph data and provides graph augmentations, which can both denoise structure and also mimic variability.  Moreover, its modular design allows augmentation to be flexibly applied to any GNN architecture.  Figure \ref{fig:tiny}(c) shows \methodtwo and \method achieves marked performance improvements over (a-b) on the toy graph.

In Section~\ref{sec:experiment}, we present and discuss an evaluation of \method across multiple GNN architectures and datasets, demonstrating a consistent improvement over the state-of-the-art, and quite large in some scenarios. Our proposed \methodtwo (\method) shows up to 17\% (9\%) absolute F1 performance improvements across datasets and GNN architectures without augmentation, and up to 16\% (9\%) over baseline augmentation strategies. 

\section{Other Related Work}
\label{sec:relatedwork}
As discussed above, relevant literature in data augmentation for graph neural networks is limited \cite{rong2019dropedge, chen2019measuring,zhang2019bayesian}. We discuss other related works in tangent domains below.

\noindent \textbf{Graph Neural Networks.} GNNs enjoy widespread use in modern graph-based machine learning due to their flexibility to incorporate node features, custom aggregations and inductive operation, unlike earlier works which were based on embedding lookups \cite{perozzi2014deepwalk, wang2016structural, tang2015line}. Many GNN variants have been developed in recent years, following the initial idea of convolution based on spectral graph theory \cite{bruna2013spectral}. Many spectral GNNs have since been developed and improved by  \cite{defferrard2016convolutional,kipf2016semi,henaff2015deep,li2018adaptive,levie2018cayleynets,ma2020unified}.  As spectral GNNs generally operate (expensively) on the full adjacency, spatial-based methods which perform graph convolution with neighborhood aggregation became prominent~\cite{hamilton2017inductive,velivckovic2017graph,monti2017geometric,gao2018large,niepert2016learning}, owing to their scalability and flexibility~\cite{ying2018graph}.  Several works propose more advanced architectures which add residual connections to facilitate deep GNN training \cite{xu2018representation, li2019deepgcns,verma2019graphmix}. 
More recently, task-specific GNNs were proposed in different fields such as behavior modeling \cite{wang2020calendar,zhao2020error,yu2020identifying}.

\noindent \textbf{Data Augmentation.} Augmentation strategies for improving generalization have been broadly studied in contexts outside of graph learning.  Traditional point-based classification approaches widely leveraged oversampling, undersampling and interpolation methods \cite{chawla2002smote,barandela2004imbalanced}.  In recent years, variants of such techniques are widely used in natural language processing (NLP) and computer vision (CV). Replacement approaches involving synonym-swapping are common in NLP~\cite{zhang2016synonym}, as are text-variation approaches~\cite{kafle-etal-2017-data} (i.e. for visual question-answering). Backtranslation methods~\cite{sennrich-etal-2016-backtrasnlation,xie2019unsupervised,edunov-etal-2018-bt-at-scale} 
have also enjoyed success.  In CV, historical image transformations in the input space, such as rotation, flipping, color space transformation, translation and noise injection \cite{shorten2019survey}, as well as recent methods such as cutout and random erasure \cite{devries2017improved, zhong2017random} have proven useful.  Recently,  augmentation via photorealistic generation through adversarial networks shows promise in several applications, especially in medicine \cite{antoniou2017data, goodfellow2014generative}.  Most-related to our work is literature on meta-learning based augmentation in CV \cite{lemley2017smart, cubuk2019autoaugment,perez2017effectiveness}, which aim to learn neural image transformation operations via an augmentation network, using a loss from a target network.  While our work is similar in motivation, it fundamentally differs in network structure, and tackles augmentation in the much-less studied graph context.

\section{Graph Data Augmentation via\\ Edge Manipulation}
\label{sec:edgemanip}
In this section, we introduce our key idea of graph data augmentation by manipulating $\mathcal{G}$ via adding and removing edges over the fixed node set.  We discuss preliminaries, practical and theoretical motivations, and considerations in evaluation under a manipulated-graph context.

\subsection{Preliminaries}
Let $\mathcal{G} = (\mathcal{V}, \mathcal{E})$ be the input graph with node set $\mathcal{V}$ and edge set $\mathcal{E}$. Let $N = |\mathcal{V}|$ be the number of nodes. We denote the adjacency matrix as $\mathbf{A} \in \{0, 1\}^{N \times N}$, where $\mathbf{A}_{ij} = 0$ indicates node $i$ and $j$ are not connected. We denote the node feature matrix as $\mathbf{X} \in \mathbb{R}^{N \times F}$, where $F$ is the dimension of the node features and $\mathbf{X}_{i:}$ indicates the feature vector of node $i$ (the $i$th row of $\mathbf{X}$). We define $\mathbf{D}$ as the diagonal degree matrix such that $\mathbf{D}_{ii} = \sum_j \mathbf{A}_{ij}$.

\noindent \textbf{Graph Neural Networks.} In this work, we use the well-known graph convolutional network (\gcn)~\cite{kipf2016semi} as an example when explaining GNNs in the following sections; however, our arguments hold straightforwardly for other GNN architectures. Each \gcn layer (GCL) is defined as:
\begin{equation}
\begin{aligned}
    \mathbf{H}^{(l+1)} &= f_{GCL}(\mathbf{A}, \mathbf{H}^{(l)}; \mathbf{W}^{(l)}) \\
    &= \sigma(\tilde{\mathbf{D}}^{-\frac{1}{2}}\tilde{\mathbf{A}}\tilde{\mathbf{D}}^{-\frac{1}{2}}\mathbf{H}^{(l)}\mathbf{W}^{(l)}),
\end{aligned}
    \label{eq:gcn_layer}
\end{equation}
where $\tilde{\mathbf{A}} = \mathbf{A} + \mathbf{I}$ is the adjacency matrix with added self-loops,  $\tilde{\mathbf{D}}$ is the diagonal degree matrix  $\tilde{D}_{ii} = \sum_j \tilde{A}_{ij}$, and $\sigma(\cdot)$ denotes a nonlinear activation such as ReLU.

\subsection{Motivation}

\noindent \textbf{Practical reasons.} Graphs aim to represent an underlying process of interest.  In reality, a processed or observed graph may not exactly align with the process it intended to model (e.g. ``which users are actually friends?'' vs. ``which users are observed to be friends?'') for several reasons. Many graphs in the real world are susceptible to noise, both adversarial and otherwise (with exceptions, like molecular or biological graphs).  Adversarial noise can manifest via spammers who pollute the space of observed interactions.  Noise can also be induced by partial observation: e.g. a friend recommendation system which never suggests certain friends to an end-user, thus preventing link formation.  Moreover, noise can be created in graph preprocessing, by adding/removing self-loops, removing isolated nodes or edges based on weights.  Finally, noise can occur due to human errors: in citation networks, a paper may omit (include) citation to a highly (ir)relevant paper by mistake.  All these scenarios can produce a gap between the ``observed graph'' and the so-called ``ideal graph'' for a downstream inference task (in our case, node classification).

Enabling an inference engine to bridge this gap suggests the promise of data augmentation via edge manipulation. In the best case, we can produce a graph $G_i$ (ideal connectivity), where supposed (but missing) links are added, and unrelated/insignificant (but existing) links removed. Figure \ref{fig:tiny} shows this benefit realized in the ZKC graph: strategically adding edges between nodes of the same group (intra-class) and removing edges between those in different groups (inter-class) substantially improves node classification test performance, despite using only a single training example per class.  Intuitively, this process encourages smoothness over same-class node embeddings and differentiates other-class node embeddings, improving distinction.

\noindent \textbf{Theoretical reasons.} \label{sec:edgemanip_theory} Strategic edge manipulation to promote intra-class edges and demote inter-class edges makes class differentiation in training trivial with a GNN, when done with label omniscience.  Consider a scenario of extremity where all possible intra-class edges and no possible inter-class edges exists, the graph can be viewed as $k$ fully connected components, where $k$ is the number of classes and all nodes in each component have the same label. 
Then by Theorem~\ref{thm:1} (proof in Appendix~\ref{appn:proof}), GNNs can easily generate distinct node representations between distinct classes, with equivalent representations for all same-class nodes. Under this ``ideal graph'' scenario, learned embeddings can be effortlessly classified.

\begin{theorem}
\label{thm:1}
Let $\mathcal{G}=(\mathcal{V}, \mathcal{E})$ be a undirected graph with adjacency matrix $\mathbf{A}$, and node features $\mathbf{X}$ be any block vector in $\mathbb{R}^{N\times F}$. Let $f: \mathbf{A}, \mathbf{X}; \mathbf{W} \rightarrow \mathbf{H}$ be any GNN layer with a permutation-invariant neighborhood aggregator over the target node and its neighbor nodes $u \cup \mathcal{N}(u)$ (e.g. Eq.~\ref{eq:gcn_layer}) with any parameters $\mathbf{W}$, and $\mathbf{H} = f(\mathbf{A}, \mathbf{X}; \mathbf{W})$ be the resulting embedding matrix. Suppose $\mathcal{G}$ contains $k$ fully connected components. Then we have:
\begin{compactenum}
    \item For any two nodes $i, j \in \mathcal{V}$ that are contained in the same connected component, $\mathbf{H}_{i:} = \mathbf{H}_{j:}$.
    \item For any two nodes $i, j \in \mathcal{V}$ that are contained in different connected components $\mathcal{S}_a, \mathcal{S}_b \subseteq \mathcal{V}$, $\mathbf{H}_{i:} \neq \mathbf{H}_{j:}$ when $\mathbf{W}$ is not all zeros and $\sum_{v \in \mathcal{S}_a}{\mathbf{X}_{v:}} \neq \varepsilon \sum_{u \in \mathcal{S}_b}{\mathbf{X}_{u:}}, \forall \varepsilon \in \mathbb{R}$.
\end{compactenum}
\end{theorem}

This result suggests that with an ideal, \emph{class-homophilic} graph $\mathcal{G}_i$, class differentiation \emph{in training} becomes trivial.  However, it does not imply such results in testing, where node connectivity is likely to reflect $\mathcal{G}$ and not $\mathcal{G}_i$.
We would expect that if modifications in training are too contrived, we risk overfitting to $\mathcal{G}_i$ and performing poorly on $\mathcal{G}$ due to a wide train-test gap.  We later show techniques (Section~\ref{sec:proposedmethod}) for approximating $\mathcal{G}_i$ with a modified graph $\mathcal{G}_m$, and show empirically that these modifications in fact help generalization, both when evaluating on graphs akin to $\mathcal{G}_m$ and $\mathcal{G}$.


\subsection{Modified and Original Graph Settings for Graph Data Augmentation}
\label{sec:3-3}
Prior CV literature \cite{wang2019survey}  considers image data augmentation a two-step process: (1) applying a transformation $f: \mathcal{S} \rightarrow \mathcal{T}$ to input images $\mathcal{S}$ to generate variants $\mathcal{T}$, and (2) utilizing $\mathcal{S} \cup \mathcal{T}$ for model training.  Graph data augmentation is notably different, since typically $|\mathcal{S}| = 1$  for node classification, unlike the image setting where $|\mathcal{S}| \gg 1$.  However, we propose two strategies with analogous, but distinct formalisms: we can either (1) apply one or multiple graph transformation operation $f: \mathcal{G} \rightarrow \mathcal{G}_m$, such that $\mathcal{G}_m$  replaces $\mathcal{G}$ for both training and inference, or (2) apply many transformations $f_i: \mathcal{G} \rightarrow \mathcal{G}^i_m$ for $i = 1 \ldots N$, such that $\mathcal{G} \cup \{\mathcal{G}^i_m\}_{i=1}^N$ may be used in training, but only $\mathcal{G}$ is used for inference. We call (1) the \emph{modified-graph} setting, and (2) the \emph{original-graph} setting, based on their inference scenario.

One might ask: when is each strategy preferable?   We reason that the answer stems from the feasibility of applying augmentation during inference to avoid a train-test gap.  The \emph{modified-graph} setting is thus most suitable in cases where a given graph is unchanging during inference.  In such cases, one can produce a single $\mathcal{G}_m$, and simply use this graph for both training and testing.  However, when inferences must be made on a dynamic graph (i.e. for large-scale, latency-sensitive applications) where calibrating new graph connectivity (akin to $\mathcal{G}$) with $\mathcal{G}_m$ during inference is infeasible (e.g. due to latency constraints), augmentation in the \emph{original-graph} setting is more appropriate. 
In such cases,  test statistics on $\mathcal{G}_m$ may be overly optimistic as performance indicators. 
In practice, these loosely align with transductive and inductive contexts in prior GNN literature.

\section{Proposed \methodshared Framework}
\label{sec:proposedmethod}
\begin{figure}[t]
    \centering
        \begin{subfigure}[b]{.43\linewidth}
          \includegraphics[width=\linewidth]{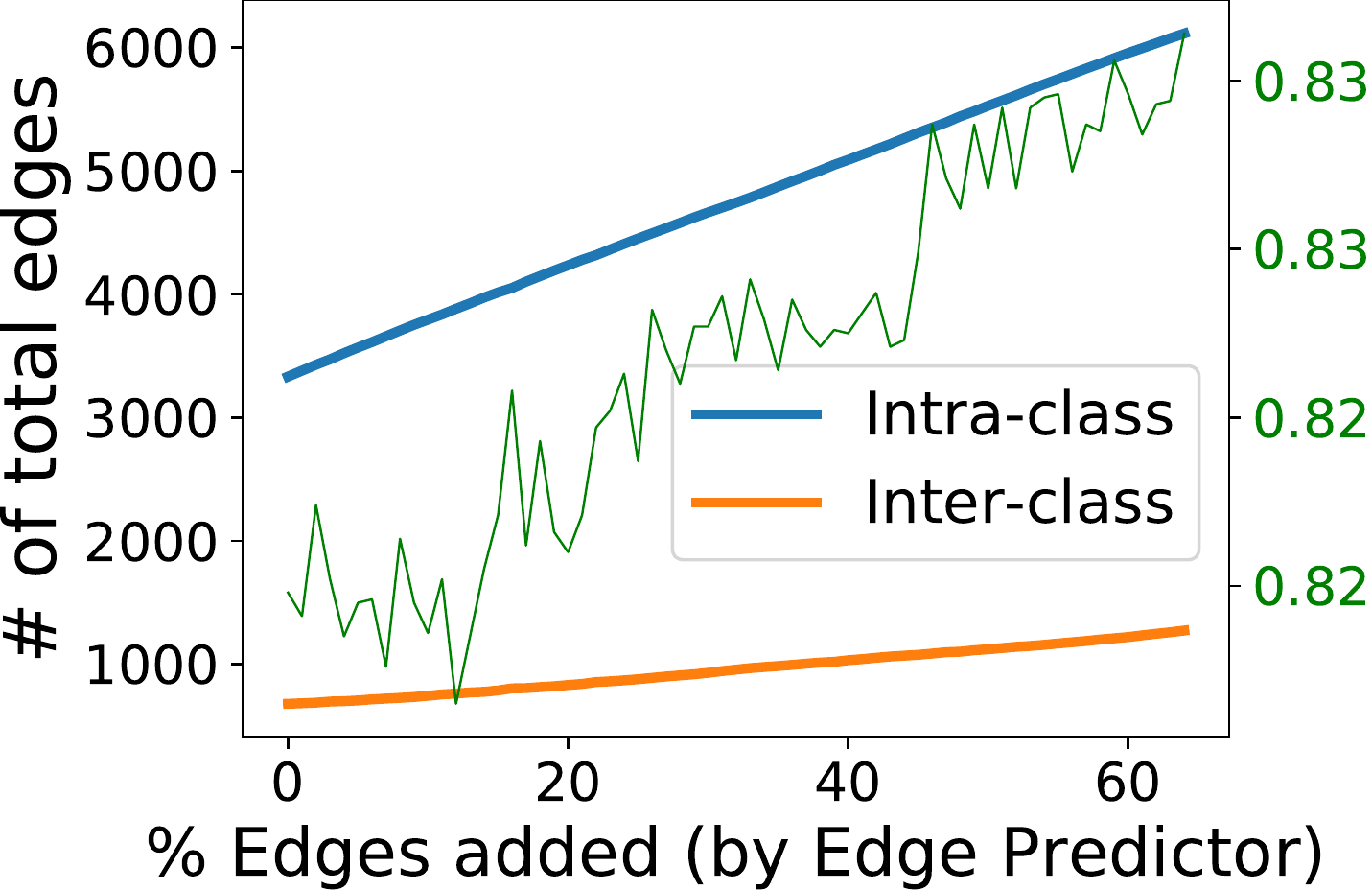}
          \caption{Learned edge +}\label{fig:}
        \end{subfigure}
        \begin{subfigure}[b]{.43\linewidth}
          \includegraphics[width=\linewidth]{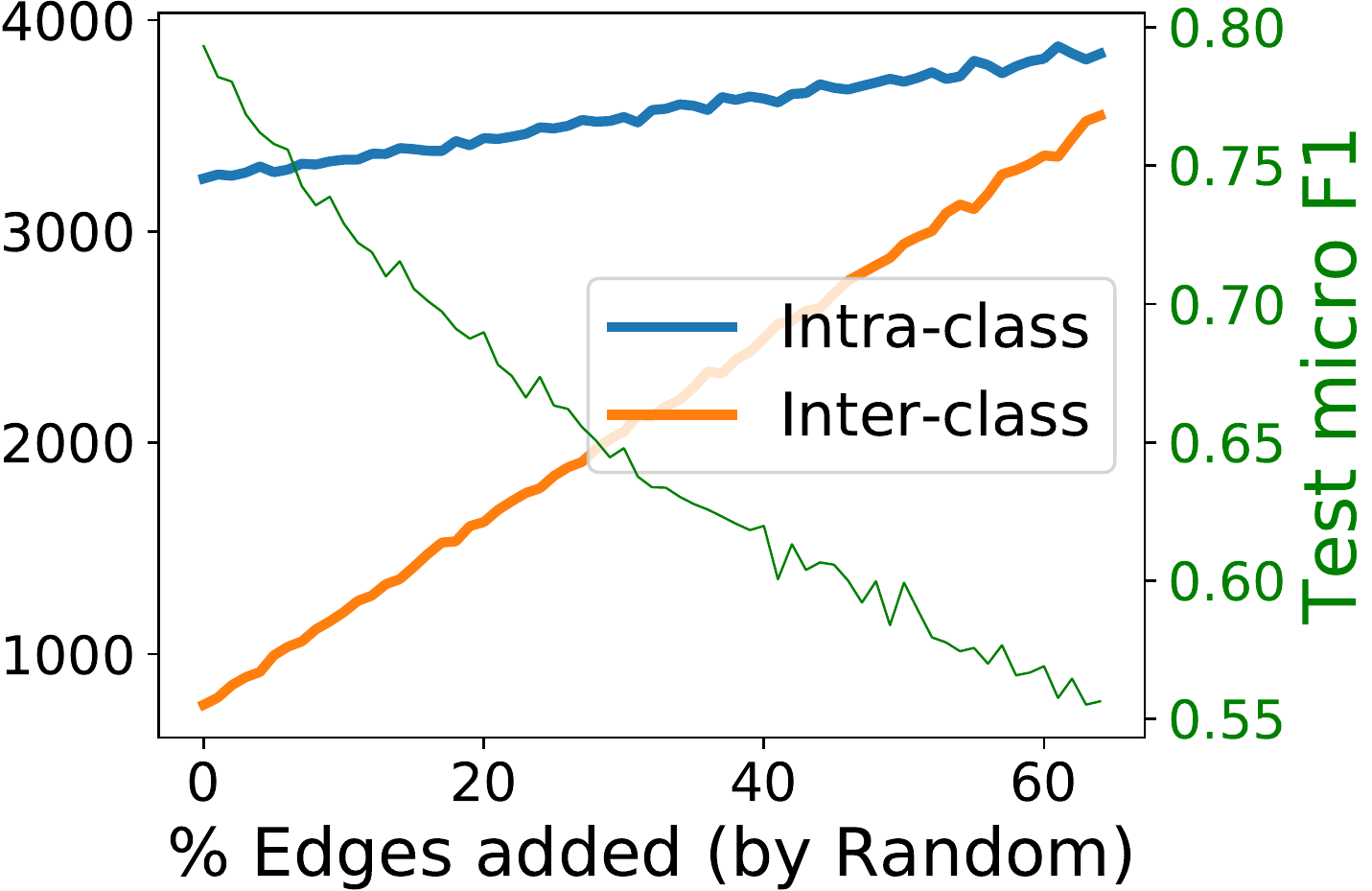}
             \caption{Random edge +}\label{fig:}
        \end{subfigure}
        \begin{subfigure}[b]{.43\linewidth}
            \includegraphics[width=\textwidth]{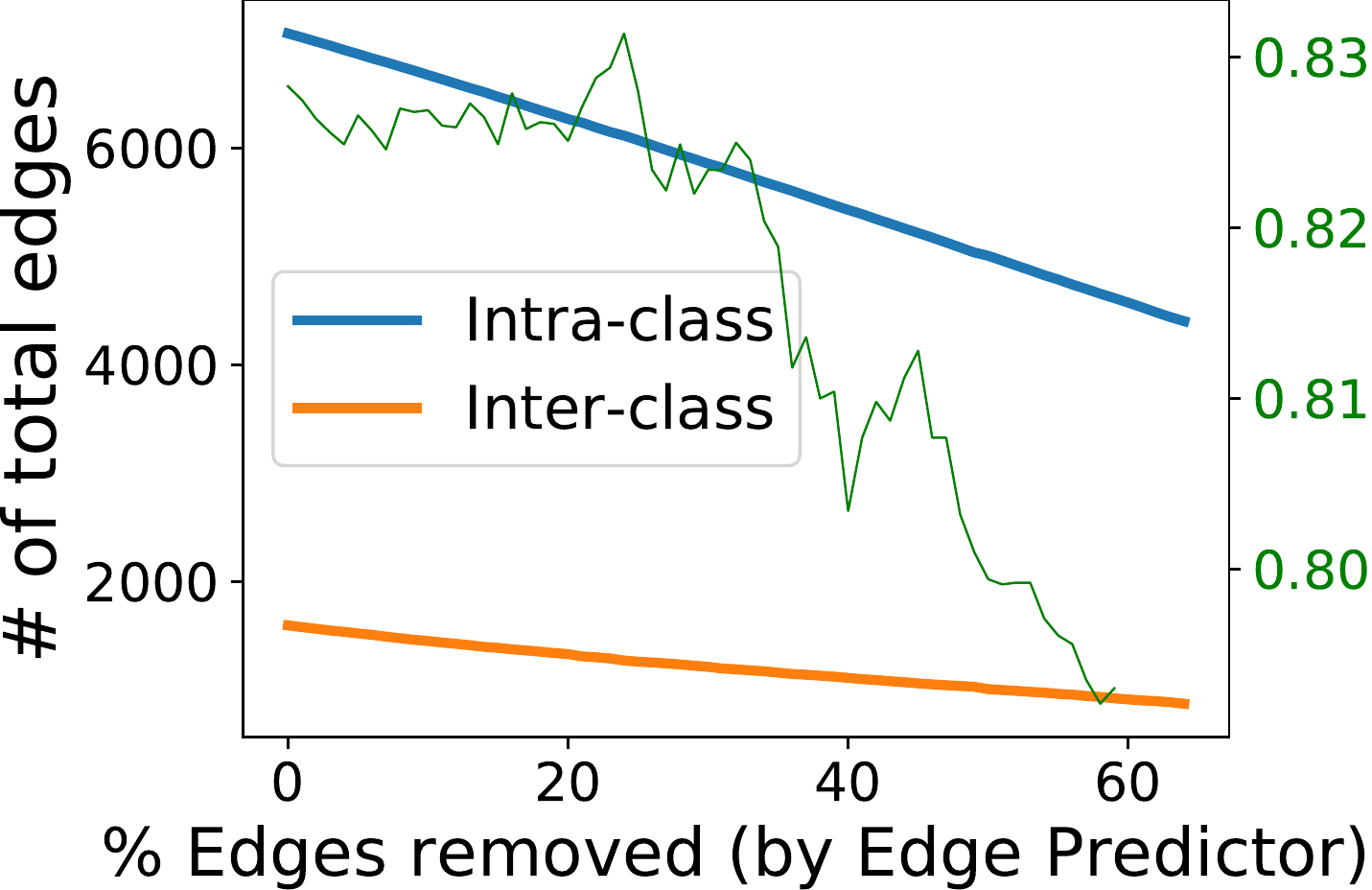}
            \caption{Learned edge -- }\label{fig:}
        \end{subfigure}
        \begin{subfigure}[b]{.43\linewidth}
            \includegraphics[width=\textwidth]{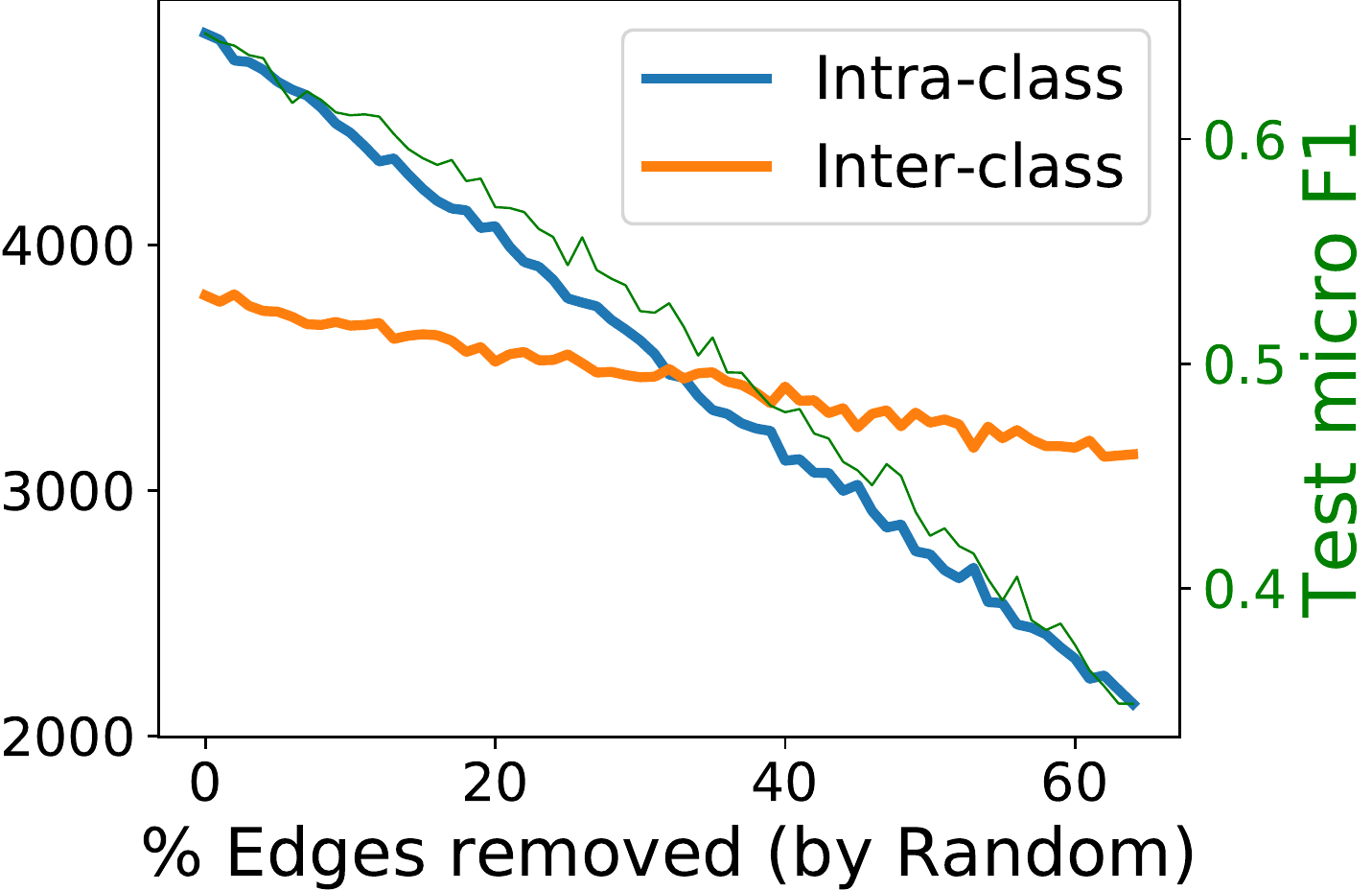}
            \caption{Random edge --}\label{fig:}
        \end{subfigure}
    \caption{\label{fig:2step} \methodtwo uses an edge-predictor module to deterministically modify a graph for future inference. Neural edge-predictors (e.g. GAE) can learn class-homophilic tendencies, promoting intra-class and demoting inter-class edges compared to random edge additions (a-b) and removals (c-d) respectively, leading to node classification performance (test micro-F1) improvements (green).}
\end{figure}


In this section, we introduce the \methodshared framework, covering two approaches for augmenting graph data in the aforementioned modified-graph and original-graph settings respectively.  Our key idea is to leverage information \emph{inherent} in the graph to predict which non-existent edges should likely exist, and which existent edges should likely be removed in $\mathcal{G}$ to produce modified graph(s) $\mathcal{G}_m$ to improve model performance.  As we later show in Section~\ref{sec:experiment}, by leveraging this label-free information, we can consistently realize improvements in test/generalization performance in semi-supervised node classification tasks across augmentation settings, GNN architectures and datasets.   


\subsection{\methodtwo for Modified-Graph Setting}
\label{sec:edgemanip_twostep}




We first introduce \methodtwo, an approach for augmentation in the modified-graph setting which includes two steps: (1) we use an edge predictor function to obtain edge probabilities for all possible and existing edges in $\mathcal{G}$. The role of the edge predictor is flexible and can generally be replaced with any suitable method. (2) Using the predicted edge probabilities, we deterministically add (remove) new (existing) edges to create a modified graph $\mathcal{G}_m$, which is used as input to a GNN node-classifier.

The edge predictor can be defined as any model $f_{ep}: \mathbf{A}, \mathbf{X} \rightarrow \mathbf{M}$, which takes the graph as input, and outputs an edge probability matrix $\mathbf{M}$ where $\mathbf{M}_{uv}$ indicates the predicted probability of an edge between nodes $u$ and $v$.  In this work, we use the graph auto-encoder (GAE)~\cite{kipf2016variational} as the edge predictor module due to its simple architecture and competitive performance. GAE consists of a two layer \gcn encoder and an inner-product decoder:
\begin{equation}
\scale[0.95]{
    \mathbf{M} = \sigma\left(\mathbf{Z}\mathbf{Z}^T\right), \ \mbox{where} \ \ \mathbf{Z} = f_{GCL}^{(1)}\left(\mathbf{A}, f_{GCL}^{(0)}\left(\mathbf{A}, \mathbf{X}\right)\right).
}
\end{equation}
$\mathbf{Z}$ denotes the hidden embeddings learned by the encoder, $\mathbf{M}$ is the predicted (symmetric) edge probability matrix produced by the inner-product decoder, and $\sigma(\cdot)$ is an element-wise sigmoid function.  Let $|\mathcal{E}|$ denote the number of edges in $\mathcal{G}$. Then, using the probability matrix $\mathbf{M}$, 
\methodtwo deterministically adds the top $i|\mathcal{E}|$ non-edges with highest edge probabilities, and removes the $j|\mathcal{E}|$ existing edges with least edge probabilities from $\mathcal{G}$ to produce $\mathcal{G}_m$, where $i, j \in [0, 1]$. This is effectively a denoising step.



Figure \ref{fig:2step} shows the change in intra-class and inter-class edges when adding/removing using GAE-learned edge probabilities and their performance implications compared to a random perturbation baseline on \cora: adding (removing) by learned probabilities results in a much steeper growth (slower decrease) of intra-class edges and much slower increase (steeper decrease) in inter-class edges compared to random.  Notably, these affect classification performance (micro-F1 scores, in green): random addition/removal hurts performance, while learned addition consistently improves performance throughout the range, and learned removal improves performance over part of the range (until ${\sim}20\%$). Importantly, these results show that while we are generally not able to produce the ideal graph $\mathcal{G}_i$ without omniscience (as discussed in Section~\ref{sec:edgemanip_theory}), such capable edge predictors can latently learn to approximate class-homophilic information in graphs and successfully promote intra-class and demote inter-class edges to realize performance gains in practice. 

\begin{figure*}[t]
    \centering
    \includegraphics[width=0.78\textwidth]{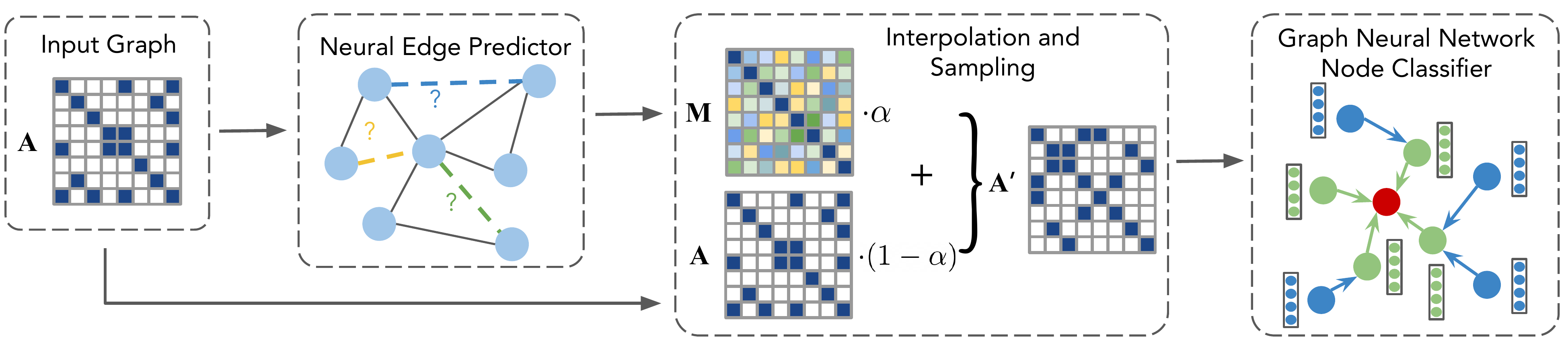}
    \caption{\method is comprised of three main components: (1) a differentiable edge predictor which produces edge probability estimates, (2) an interpolation and sampling step which produces sparse graph variants, and (3) a GNN which learns embeddings for node classification using these variants.  The model is trained end-to-end with both classification and edge prediction losses.
    } \label{fig:gaug}
\end{figure*}

\methodtwo shares the same time and space complexity as its associated GNN architecture during training/inference, while requiring extra disk space to save the dense $O(N^2)$ edge probability matrix $\mathbf{M}$ for manipulation. Note that $\mathbf{M}$'s computation can be trivially parallelized. 


\subsection{\method for Original-Graph Setting}


To complement the above approach, we propose \method for the original-graph setting, where we cannot benefit from graph manipulation at inference time.  \method is reminiscent of the two-step approach in \methodshared in that it also uses an edge prediction module for the benefit of node classification, but also aims to improve model generalization (test performance on $\mathcal{G}$) by generating graph variants $\{\mathcal{G}^i_m\}_{i=1}^N$ via edge prediction and hence improve data diversity.  \method does not require discrete specification of edges to add/remove, is end-to-end trainable, and utilizes both edge prediction and node-classification losses to iteratively improve augmentation capacity of the edge predictor and classification capacity of the node classifier GNN.  Figure~\ref{fig:gaug} shows the overall architecture: each training iteration exposes the node-classifier to a new augmented graph variant.




Unlike \methodtwo's deterministic graph modification step, \method supports a learnable, stochastic augmentation process. 
As such, we again use the graph auto-encoder (GAE) for edge prediction.  To prevent the edge predictor from arbitrarily deviating from original graph adjacency, we interpolate the predicted  $\mathbf{M}$ with the original $\mathbf{A}$ to derive an adjacency $\mathbf{P}$.  In the edge sampling phase, we sparsify $\mathbf{P}$ with Bernoulli sampling on each edge to get the graph variant adjacency $\mathbf{A}'$. For training purposes, we employ a (soft, differentiable) relaxed Bernoulli sampling procedure as a 
Bernoulli approximation. This relaxation is a binary special case of the Gumbel-Softmax reparameterization trick ~\cite{maddison2016concrete, jang2016categorical}.  Using the relaxed  sample, we apply a straight-through (ST) gradient estimator~\cite{bengio2013estimating}, which rounds the relaxed samples in the forward pass, hence sparsifying the adjacency. In the backward pass, gradients are directly passed to the relaxed samples rather than the rounded values, enabling training. Formally, 
\begin{equation}
\begin{aligned}
    &\mathbf{A}_{ij}' = \left\lfloor \frac{1}{1 + e^{-(\log \mathbf{P}_{ij} + G) / \tau}} + \frac{1}{2} \right\rfloor,  \\ 
    & \mbox{where} \quad \mathbf{P}_{ij} = \alpha \mathbf{M}_{ij} + (1-\alpha)\mathbf{A}_{ij} 
\end{aligned}
\end{equation}
where $\mathbf{A}'$ is the sampled adjacency matrix, $\tau$ is the temperature of Gumbel-Softmax distribution, $G \sim Gumbel(0,1)$ is a Gumbel random variate, and $\alpha$ is a hyperparameter mediating the influence of edge predictor on the original graph.

The graph variant adjacency $\mathbf{A}'$ is passed along with node features $\mathbf{X}$ to the GNN node classifier. We then backpropagate using a joint node-classification loss $\mathcal{L}_{nc}$ and edge-prediction loss $\mathcal{L}_{ep}$
\begin{equation}
    \begin{aligned}
    &\mathcal{L} = \mathcal{L}_{nc} + \beta \mathcal{L}_{ep}, \\
    \mbox{where} \quad & \mathcal{L}_{nc} = CE(\mathbf{\hat{y}}, \mathbf{y}) \\ 
    \mbox{and} \quad & \mathcal{L}_{ep} = BCE(\sigma(f_{ep}(\mathbf{A}, \mathbf{X})), \mathbf{A})
\end{aligned}
\end{equation}
where $\beta$ is a hyperparameter to weight the reconstruction loss,  $\sigma(\cdot)$ is an elementwise sigmoid, $\mathbf{y}, \mathbf{\hat{y}}$ denote ground-truth node class labels and predicted probabilities, and $BCE/CE$ indicate standard (binary) cross-entropy loss. We train using $\mathcal{L}_{ep}$ in addition to $\mathcal{L}_{nc}$ to control potentially excessive drift in edge prediction performance. The node-classifier GNN is then directly used for inference, on $\mathcal{G}$. 

During training, \method has a space complexity of $O(N^2)$ in full-batch setting due to backpropagation through all entries of the adjacency matrix.  Fortunately, we can easily adapt the graph mini-batch training introduced by Hamilton et al.~\cite{hamilton2017inductive} to achieve an acceptable space complexity of $O(M^2)$, where $M$ is the batch size. Appendix~\ref{appn:implementation_gaug} further details (pre)training, mini-batching, and implementation choices.

\section{Evaluation}
\label{sec:experiment}
In this section, we evaluate the performance of \methodtwo and \method across architectures and datasets, and over alternative strategies for graph data augmentation.  We also showcase their abilities to approximate class-homophily via edge prediction and  sensitivity to supervision.

\begin{table*}[t]
\small
  \caption{Summary statistics and experimental setup for the six evaluation datasets.}
  \label{tab:datasets}
  \centering
  \scale[0.9]{\begin{tabular}{lccccccc}
    \toprule
     & \cora & \citeseer & \ppi & \blogc & \flickr & \airusa \\
    \midrule
    \# Nodes & 2,708 & 3,327 & 10,076 & 5,196 & 7,575 & 1,190 \\
    \# Edges & 5,278 & 4,552 & 157,213 & 171,743 & 239,738 & 13,599 \\
    \# Features & 1,433 & 3,703 & 50 & 8,189 & 12,047 & 238 \\
    \# Classes & 7 & 6 &  121 & 6 & 9 & 4 \\
    \# Training nodes & 140 & 120 & 1,007 & 519 & 757 & 119 \\
    \# Validation nodes & 500 & 500 & 2,015 & 1,039 & 1,515 & 238 \\
    \# Test nodes & 1,000 & 1,000 & 7,054 & 3,638 & 5,303 & 833 \\
    \bottomrule
  \end{tabular}}
\end{table*}

\begin{table*}[t!]
\setlength{\tabcolsep}{7pt}
\small
  \caption{\methodshared performance across GNN architectures and six benchmark datasets. }
  \label{tab:results}
  \centering
  \scale[0.9]{\begin{tabular}{clcccccc}
    \toprule
    GNN Arch. & Method & \cora & \citeseer & \ppi & \textsc{BlogC} & \flickr & \airusa \\
    \midrule
    \multirow{6}*{\gcn} 
    & Original & 81.6$\pm$0.7 & 71.6$\pm$0.4 & 43.4$\pm$0.2 & 75.0$\pm$0.4 & 61.2$\pm$0.4 & 56.0$\pm$0.8 \\
    \cmidrule(lr){2-8}
    & +\bgcn & 81.2$\pm$0.8 & 72.4$\pm$0.5 & -- & 72.0$\pm$2.3 & 52.7$\pm$2.8 & 56.5$\pm$0.9 \\
    & +\adaedge & 81.9$\pm$0.7 & 72.8$\pm$0.7 & 43.6$\pm$0.2 & 75.3$\pm$0.3 & 61.2$\pm$0.5 & 57.2$\pm$0.8 \\
    & +\methodtwo & 83.5$\pm$0.4 & 72.3$\pm$0.4 & 43.5$\pm$0.2 & \textbf{77.6$\pm$0.4} & \textbf{68.2$\pm$0.7} & 61.2$\pm$0.5 \\
    \cmidrule(lr){2-8}
    & +\dropedge & 82.0$\pm$0.8 & 71.8$\pm$0.2 & 43.5$\pm$0.2 & 75.4$\pm$0.3 & 61.4$\pm$0.7 & 56.9$\pm$0.6 \\
    & +\method & \textbf{83.6$\pm$0.5} & \textbf{73.3$\pm$1.1} & \textbf{46.6$\pm$0.3} & 75.9$\pm$0.2 & 62.2$\pm$0.3 & \textbf{61.4$\pm$0.9} \\
    \midrule
    \multirow{6}*{\gsage} 
    & Original & 81.3$\pm$0.5 & 70.6$\pm$0.5 & 40.4$\pm$0.9 & 73.4$\pm$0.4 & 57.4$\pm$0.5 & 57.0$\pm$0.7 \\
    \cmidrule(lr){2-8}
    & +\bgcn & 80.5$\pm$0.1 & 70.8$\pm$0.1 & -- & 73.2$\pm$0.2 & 58.1$\pm$0.3 & 53.5$\pm$0.3 \\
    & +\adaedge & 81.5$\pm$0.6 & 71.3$\pm$0.8 & 41.6$\pm$0.8 & 73.6$\pm$0.4 & 57.7$\pm$0.7 & 57.1$\pm$0.5 \\
    & +\methodtwo & \textbf{83.2$\pm$0.4} & 71.2$\pm$0.4 & 41.1$\pm$1.0 & \textbf{77.0$\pm$0.4} & \textbf{65.2$\pm$0.4} & \textbf{60.1$\pm$0.5} \\
    \cmidrule(lr){2-8}
    & +\dropedge & 81.6$\pm$0.5 & 70.8$\pm$0.5 & 41.1$\pm$1.0 & 73.8$\pm$0.4 & 58.4$\pm$0.7 & 57.1$\pm$0.5 \\
    & +\method & 82.0$\pm$0.5 & \textbf{72.7$\pm$0.7} & \textbf{44.4$\pm$0.5} & 73.9$\pm$0.4 & 56.3$\pm$0.6 & 57.1$\pm$0.7 \\
    \midrule
    \multirow{6}*{\gat} 
    & Original & 81.3$\pm$1.1 & 70.5$\pm$0.7 & 41.5$\pm$0.7 & 63.8$\pm$5.2 & 46.9$\pm$1.6 & 52.0$\pm$1.3 \\
    \cmidrule(lr){2-8}
    & +\bgcn & 80.8$\pm$0.8 & 70.8$\pm$0.6 & -- & 61.4$\pm$4.0 & 46.5$\pm$1.9 & 54.1$\pm$3.2 \\
    & +\adaedge & 82.0$\pm$0.6 & 71.1$\pm$0.8 & 42.6$\pm$0.9 & 68.2$\pm$2.4 & 48.2$\pm$1.0 & 54.5$\pm$1.9 \\
    & +\methodtwo & 82.1$\pm$1.0 & 71.5$\pm$0.5 & 42.8$\pm$0.9 & 70.8$\pm$1.0 & \textbf{63.7$\pm$0.9} & \textbf{59.0$\pm$0.6} \\
    \cmidrule(lr){2-8}
    & +\dropedge & 81.9$\pm$0.6 & 71.0$\pm$0.5 & \textbf{45.9$\pm$0.3} & 70.4$\pm$2.4 & 50.0$\pm$1.6 & 52.8$\pm$1.7 \\
    & +\method & \textbf{82.2$\pm$0.8} & \textbf{71.6$\pm$1.1} & 44.9$\pm$0.9 & \textbf{71.0$\pm$1.1} & 51.9$\pm$0.5 & 54.6$\pm$1.1 \\
    \midrule
    \multirow{6}*{\jknet} 
    & Original & 78.8$\pm$1.5 & 67.6$\pm$1.8 & 44.1$\pm$0.7 & 70.0$\pm$0.4 & 56.7$\pm$0.4 & 58.2$\pm$1.5 \\
    \cmidrule(lr){2-8}
    & +\bgcn & 80.2$\pm$0.7 & 69.1$\pm$0.5 & -- & 65.7$\pm$2.2 & 53.6$\pm$1.7 & 55.9$\pm$0.8 \\
    & +\adaedge & 80.4$\pm$1.4 & 68.9$\pm$1.2 & 44.8$\pm$0.9 & 70.7$\pm$0.4 & 57.0$\pm$0.3 & 59.4$\pm$1.0 \\
    & +\methodtwo & \textbf{81.8$\pm$0.9} & 68.2$\pm$1.4 & 47.4$\pm$0.6 & \textbf{71.9$\pm$0.5} & \textbf{65.7$\pm$0.8} & 60.2$\pm$0.6 \\
    \cmidrule(lr){2-8}
    & +\dropedge & 80.4$\pm$0.7 & 69.4$\pm$1.1 & 46.3$\pm$0.2 & 70.9$\pm$0.4 & 58.5$\pm$0.7 & 59.1$\pm$1.1 \\
    & +\method & 80.5$\pm$0.9 & \textbf{69.7$\pm$1.4} & \textbf{53.1$\pm$0.3} & 71.0$\pm$0.6 & 55.7$\pm$0.5 & \textbf{60.4$\pm$1.0} \\
    \bottomrule
  \end{tabular}}
\end{table*}

\subsection{Experimental Setup}
We evaluate using 6 benchmark datasets across domains: citation networks (\cora, \citeseer \cite{kipf2016semi}), protein-protein interactions (\ppi \cite{hamilton2017inductive}), social networks (\blogc, \flickr \cite{huang2017label}), and air traffic (\airusa \cite{wu2019net}). 
Statistics for each dataset are shown in Table~\ref{tab:datasets}, with more details in Appendix~\ref{appn:data}.
We follow the semi-supervised setting in most GNN literature~\cite{kipf2016semi, velivckovic2017graph} for train/validation/test splitting on \cora and \citeseer, and a 10/20/70\% split on other datasets due to varying choices in prior work.  We evaluate \methodtwo and \method using 4 widely used GNN architectures:
\gcn~\cite{kipf2016semi}, \gsage~\cite{hamilton2017inductive}, \gat~\cite{velivckovic2017graph} and \jknet~\cite{xu2018representation}. We compare our \methodtwo (modified-graph) and \method (original-graph) performance with that achieved by standard GNN performance, 
as well as three state-of-the-art baselines: \adaedge~\cite{chen2019measuring} (modified-graph), \bgcn~\cite{zhang2019bayesian} (modified-graph), and \dropedge~\cite{rong2019dropedge} (original-graph) evaluating on $\mathcal{G}_m$ and $\mathcal{G}$, respectively. We also show results of proposed \methodshared methods on large graphs~\cite{hu2020open} in Appendix~\ref{appn:exp_large} to show their ability of mini-batching. We report test micro-F1 scores over 30 runs, employing Optuna~\cite{akiba2019optuna} for efficient hyperparameter search. Note that for classification tasks which every object is guaranteed to be assigned to exactly one ground truth class (all datasets except \ppi), micro-F1 score is mathematically equivalent to accuracy (proof in Appendix~\ref{appn:proof2}).
Our implementation is made publicly available\footnote{\url{https://github.com/zhao-tong/GAug}}.

\begin{figure}[t]
    \centering
        \begin{subfigure}[b]{.35\linewidth}
          \includegraphics[width=\linewidth]{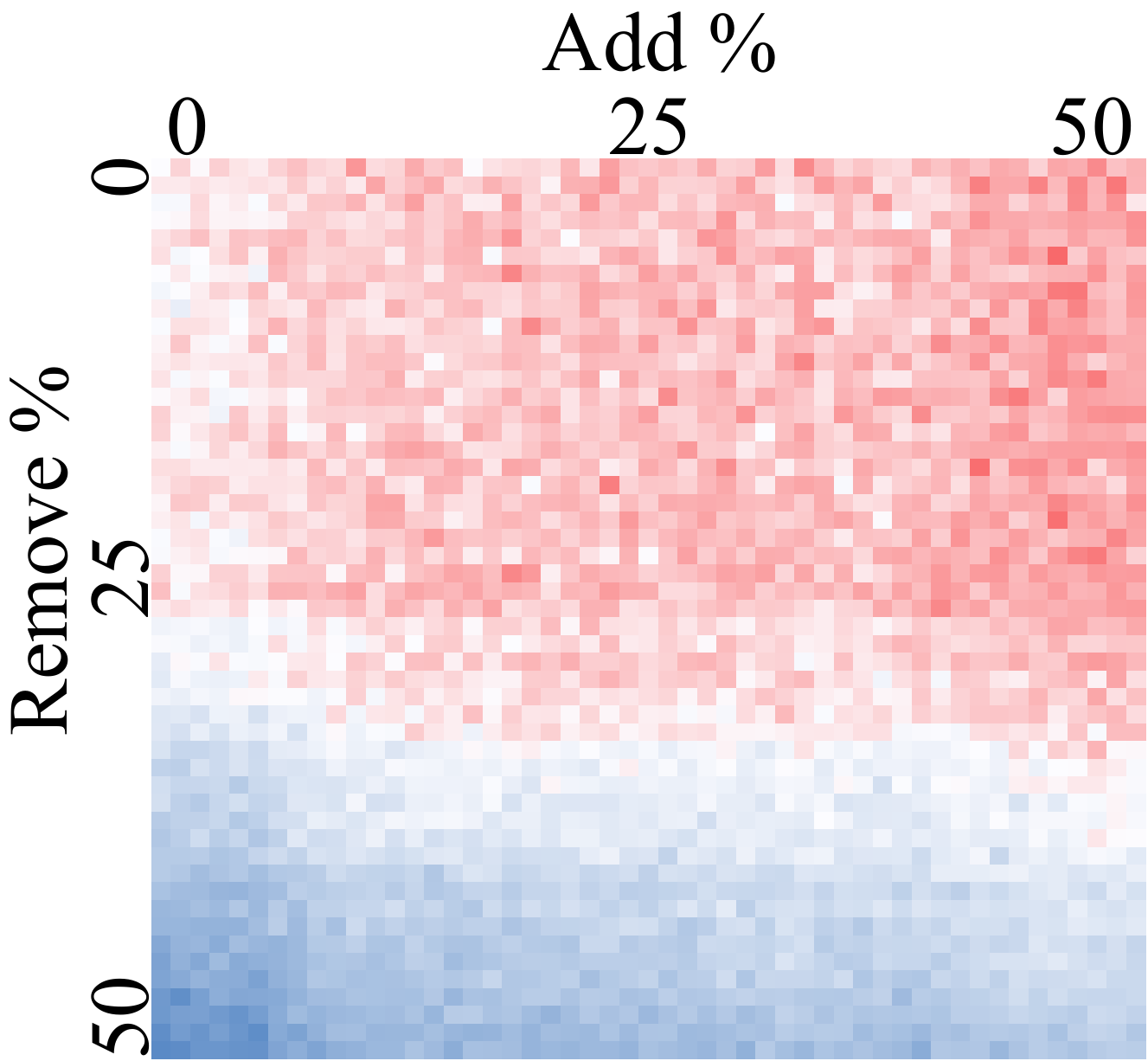}
            \caption{\cora}\label{fig:}
        \end{subfigure}
        \quad
        \begin{subfigure}[b]{.35\linewidth}
          \includegraphics[width=\linewidth]{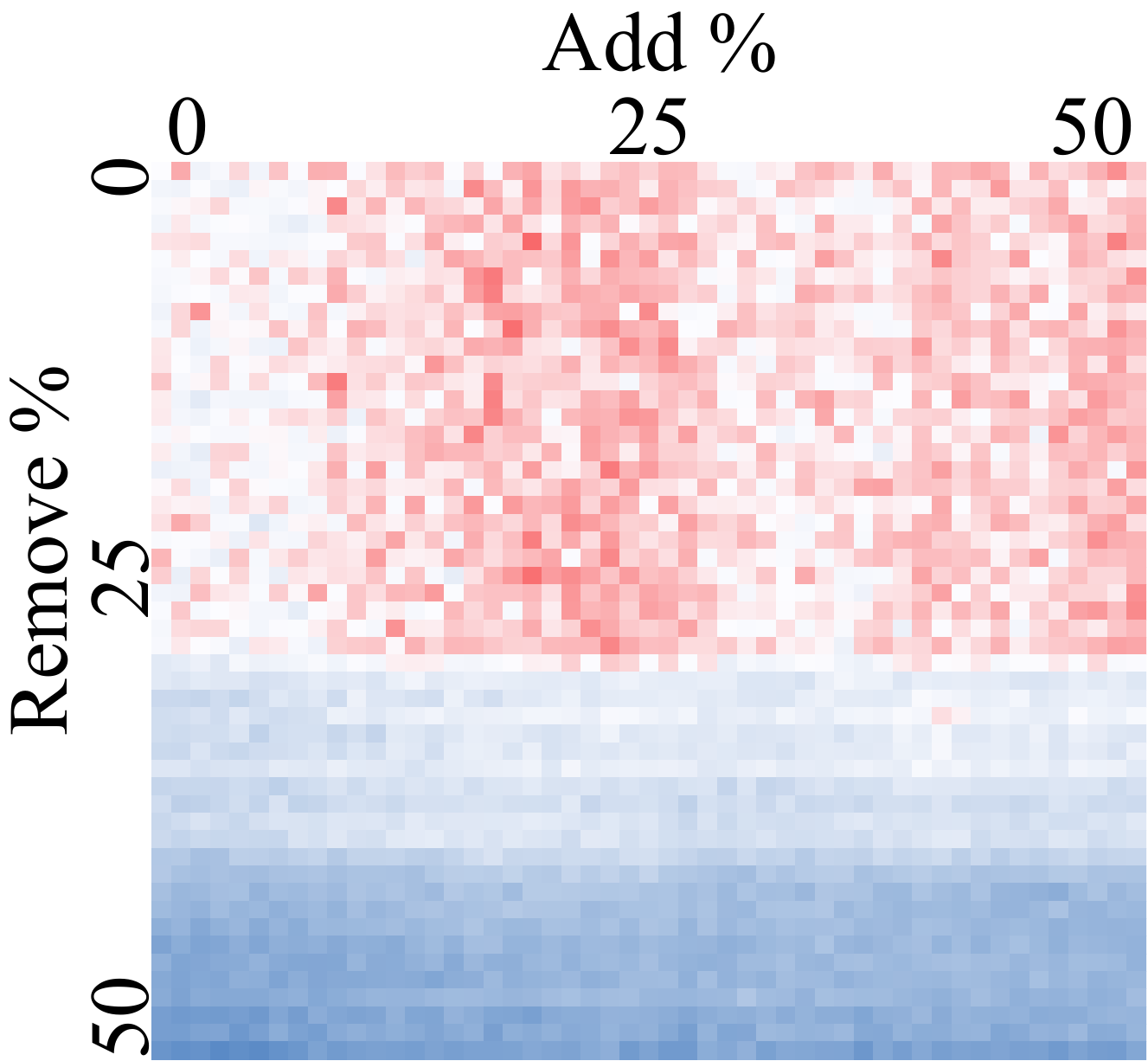}
            \caption{\citeseer}\label{fig:}
        \end{subfigure}
        \quad
        \begin{subfigure}[b]{.35\linewidth}
            \includegraphics[width=\textwidth]{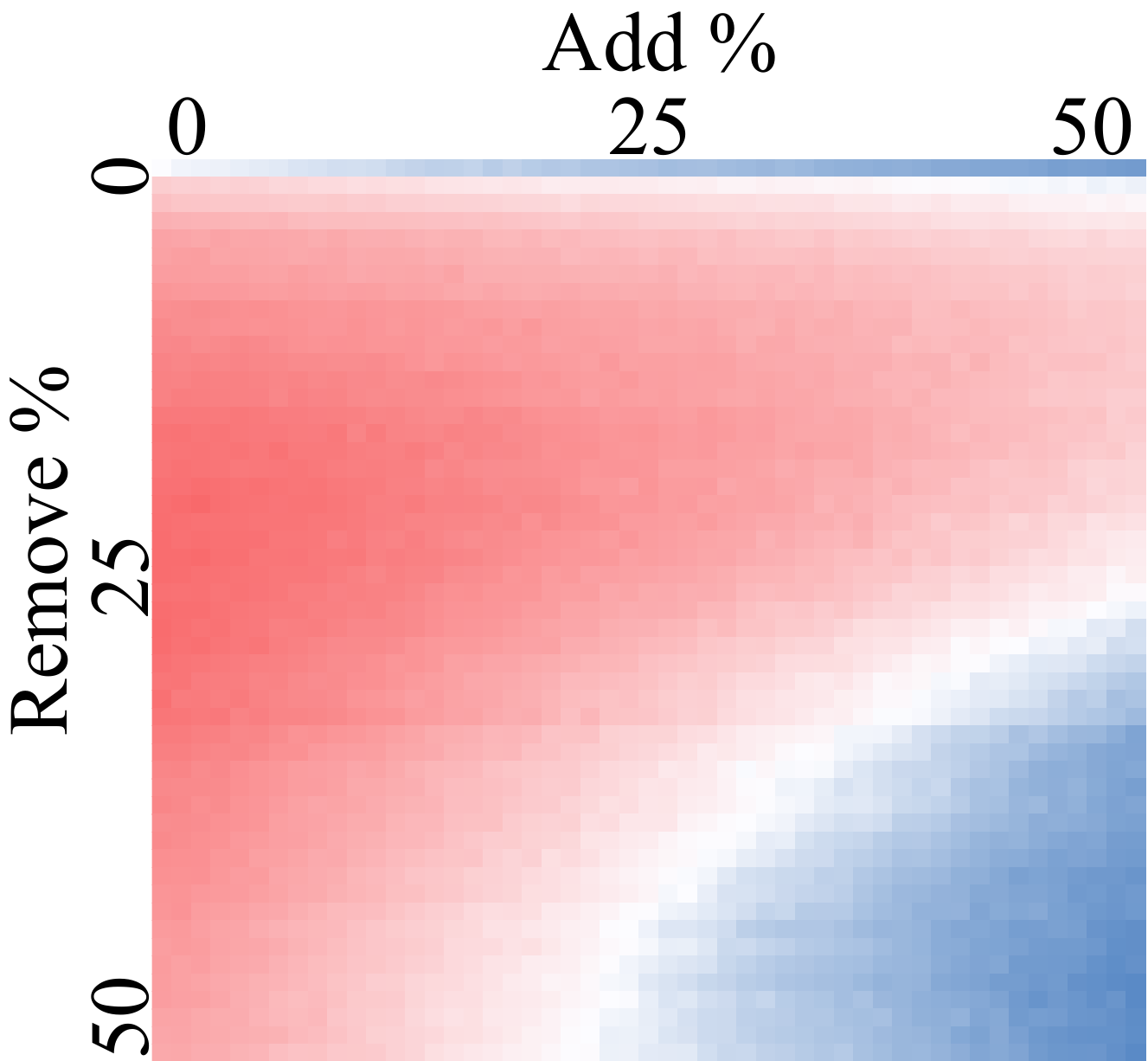}
            \caption{\flickr}\label{fig:}
        \end{subfigure}
        \quad
        \begin{subfigure}[b]{.35\linewidth}
            \includegraphics[width=\textwidth]{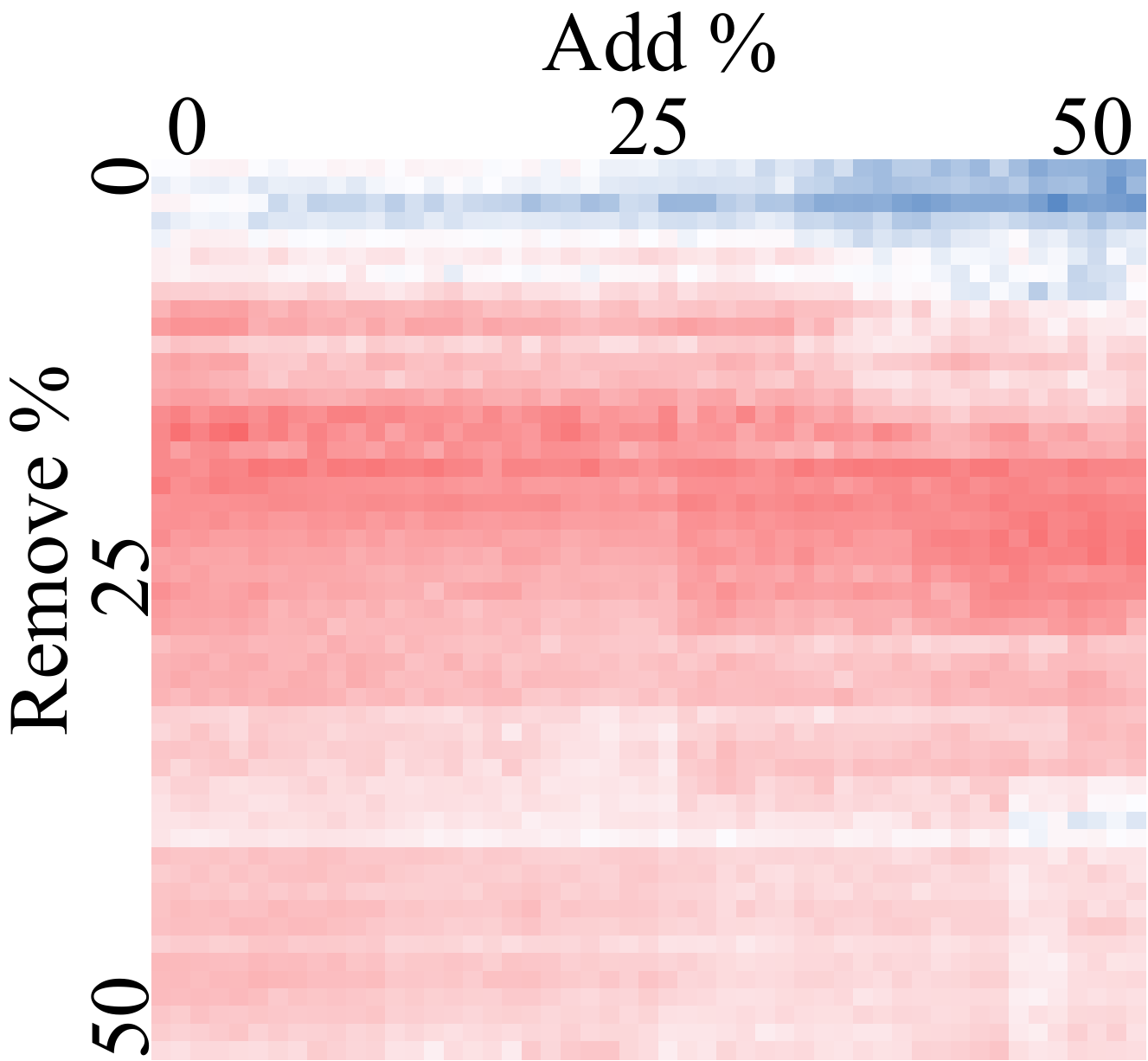}
            \caption{\airusa}\label{fig:}
        \end{subfigure}
    \caption{Classification (test) performance heatmaps of \methodtwo on various datasets when adding/dropping edges. Red-white-blue indicate outperformance, at-par, and underperformance w.r.t. \gcn on $\mathcal{G}$. Pixel $(0,0)$ indicates $\mathcal{G}$, and $x$ ($y$) axes show \% edges added (removed). 
    }\label{fig:heat}
\end{figure}

\subsection{Experimental Results}

We show comparative results against current baselines in Table \ref{tab:results}.  Table \ref{tab:results} is organized per architecture (row), per dataset (column), and original-graph and modified-graph settings (within-row). Note that results of \bgcn on \ppi are missing due to CUDA out of memory error when running the code package from the authors. We bold best-performance per architecture and dataset, but not per augmentation setting for visual clarity.  In short, \method and \methodtwo consistently improve over GNN architectures, datasets and alternatives, with a single exception for \gat on \ppi, on which \dropedge performs the best.  

    


\noindent \textbf{Improvement across GNN architectures.} \methodshared achieves improvements over all 4 GNN architectures (averaged across datasets): \methodtwo improves 4.6\% (\gcn),  4.8\% (\gsage), 10.9\% (\gat) and 5.7\% (\jknet). \method improves 4.1\%, 2.1\%, 6.3\% and 4.9\%, respectively.  We note that augmentation especially improves \gat performance, as self-attention based models are sensitive to connectivity.

\noindent \textbf{Improvements across datasets.} \methodshared also achieves improvements over all 6 datasets (averaged across architectures): \methodtwo improves 2.4\%, 1.0\%, 3.1\%, 5.5\%, 19.2\%, 7.9\% for each dataset (left to right in Table \ref{tab:results}).  Figure \ref{fig:heat} shows \methodtwo (with \gcn) classification performance heatmaps on 4 datasets when adding/removing edges according to various $i,j$ (Section~\ref{sec:edgemanip_twostep}).  Notably, while improvements(red) over original \gcn on $\mathcal{G}$ differ over $i,j$ and by dataset, they are feasible in all cases. These improvements are not necessarily monotonic with edge addition(row) or removal(column), and can encounter transitions.  Empirically, we notice these boundaries correspond to excessive class mixing (addition) or graph shattering (removal).  \method improves 1.6\%, 2.5\%, 11.5\%, 3.6\%, 2.2\%, 4.7\%.  We note that both methods achieves large improvements in social data (\blogc and \flickr) where noisy edges may be prominent due to spam or bots (supporting intuition from Section~\ref{sec:edgemanip_theory}): Figure \ref{fig:heat}(c) shows substantial edge removal significantly helps performance.

\noindent \textbf{Improvements over alternatives.} \methodshared also outperforms augmentation over \bgcn, \adaedge, and \dropedge (averaged across datasets/architectures): \methodtwo improves 9.3\%, 4.8\%, and 4.1\% respectively, while \method improves 4.9\%, 2.7\%, and 2.0\% respectively. We reason that \methodtwo outperforms \bgcn and \adaedge by avoiding iterative error propagation, as well as directly manipulating edges based on the graph, rather than indirectly through classification results. \method outperforms \dropedge via learned denoising via addition and removal, rather than random edge removal. Note that some baselines have worse performance than vanilla GNNs, as careless augmentation/modification on the graph can hurt performance by removing critical edges and adding incorrect ones. 

\begin{figure}[t]
\centering
  \centering
    \begin{subfigure}[b]{.42\linewidth}
    \includegraphics[width=\linewidth]{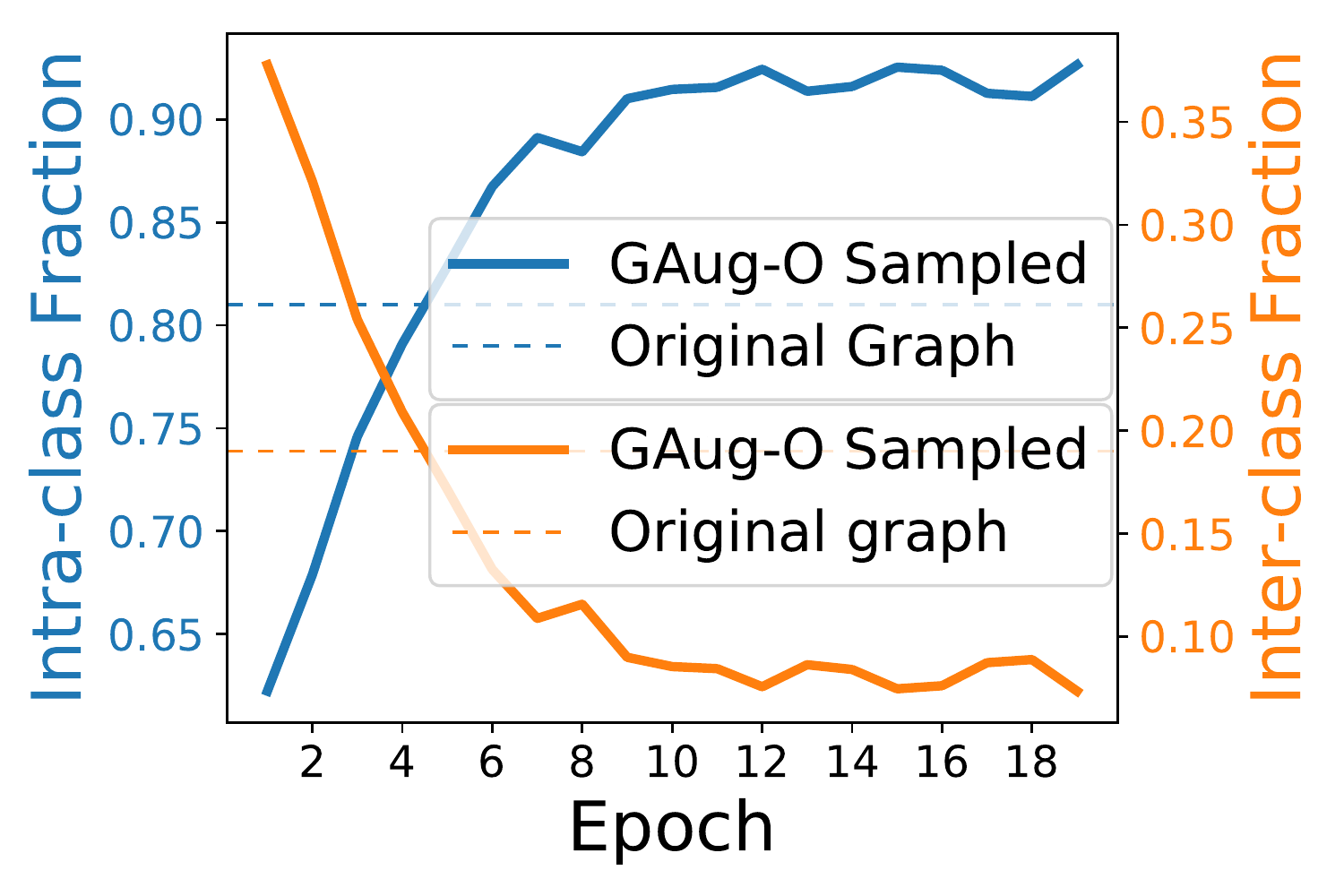}
    \caption{Edge makeup}\label{fig:intra-frac}
    \end{subfigure}
    \begin{subfigure}[b]{.42\linewidth}
    \includegraphics[width=\linewidth]{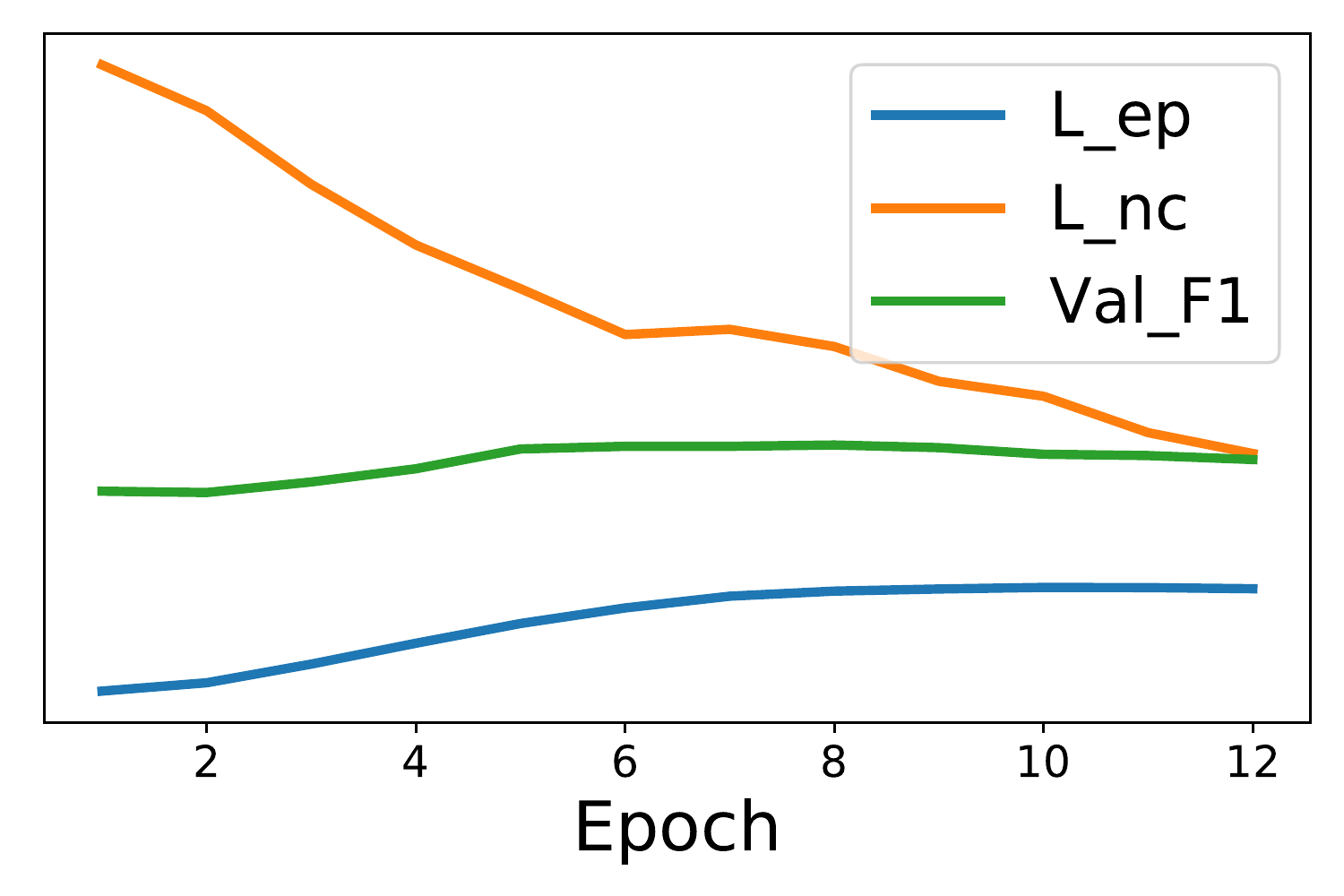}
    \caption{Learning curve}\label{fig:loss}
    \end{subfigure}
    
    \caption{\method promotes class-homophily (a), producing classification improvements (b).}
\end{figure}

\noindent \textbf{Promoting class-homophily.} Figure \ref{fig:intra-frac} shows (on \cora) that the edge predictor in \method learns to promote intra-class edges and demote inter-class ones, echoing results from Figure \ref{fig:2step} on \methodtwo, facilitating message passing and improving performance.  Figure \ref{fig:loss} shows that $\mathcal{L}_{nc}$ decreases and validation F1 improves over the first few epochs, while $\mathcal{L}_{ep}$ increases to reconcile with supervision from $\mathcal{L}_{nc}$.  Later on, the $\mathcal{L}_{nc}$ continues to decrease while intra-class ratio increases (overfitting).

\begin{figure}[t]
  \centering
  \begin{subfigure}[b]{.42\linewidth}
    \includegraphics[width=\linewidth]{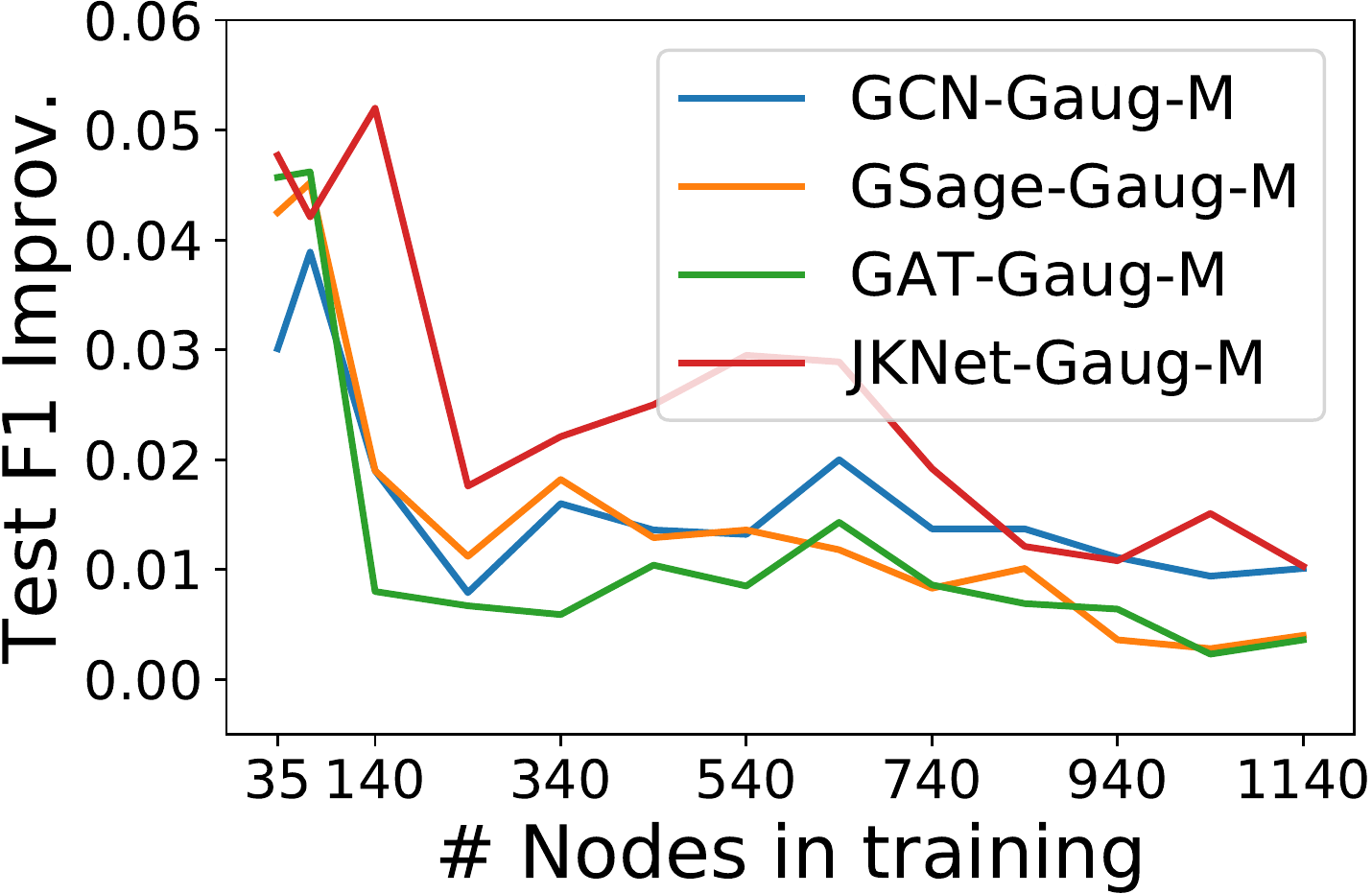}
    \caption{\methodtwo}
    \end{subfigure}
    \begin{subfigure}[b]{.42\linewidth}
    \includegraphics[width=\linewidth]{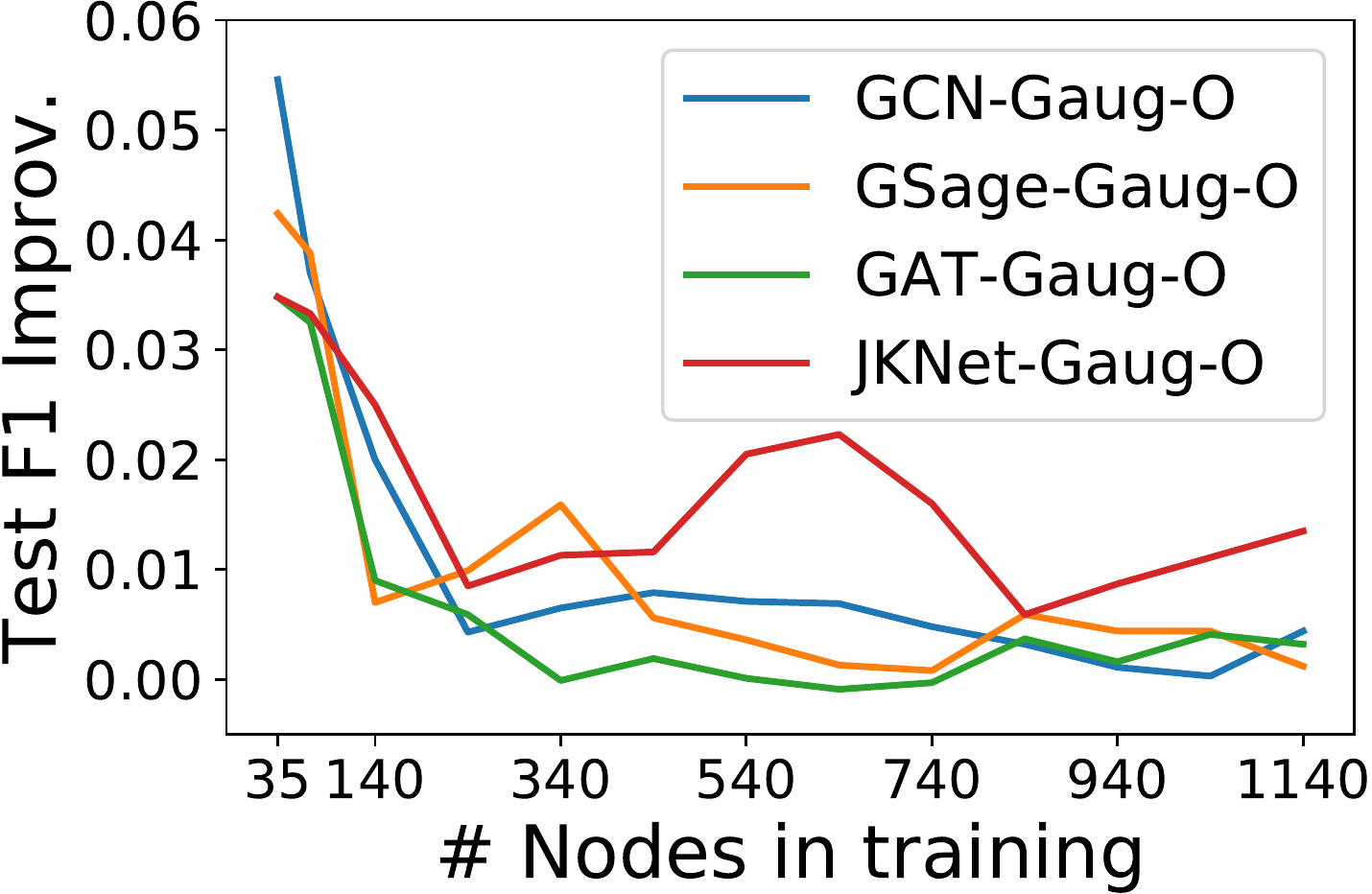}
    \caption{\method}
    \end{subfigure}
    \caption{\methodshared augmentation especially improves performance under weak supervision.}
    \label{fig:trainsize}
\end{figure}

\noindent \textbf{Sensitivity to supervision.}
Figure \ref{fig:trainsize} shows that both \methodshared is especially powerful under weak supervision, producing large F1 improvements with few labeled samples.  Moreover, augmentation helps achieve equal performance w.r.t standard methods with fewer training samples. Naturally, improvements shrink in the presence of more supervision.  \methodtwo tends towards slightly larger outperformance compared to \method with more training nodes, since inference benefits from persistent graph modifications in the former but not the latter.

\section{Conclusion}
\label{sec:conclusion}
Data augmentation for facilitating GNN training has unique challenges due to graph irregularity.  Our work tackles this problem by utilizing neural edge predictors as a means of exposing GNNs to likely (but nonexistent) edges and limiting exposure to unlikely (but existent) ones.  We show that such edge predictors can encode class-homophily to promote intra-class edges and inter-class edges. We propose the \methodshared graph data augmentation framework which uses these insights to improve node classification performance in two inference settings. Extensive experiments show our proposed \method and \methodtwo achieve up to 17\% (9\%) absolute F1 performance improvements across architectures and datasets, and 15\% (8\%) over augmentation baselines.

\section*{Ethical Impact}

We do not foresee ethical concerns posed by our method, but concede that both ethical and unethical applications of graph-based machine learning techniques may benefit from the improvements induced by our work. Care must be taken, in general, to ensure positive ethical and societal consequences of machine learning.

\bibliography{ref}

\begin{thebibliography}{61}
\providecommand{\natexlab}[1]{#1}
\providecommand{\url}[1]{\texttt{#1}}
\providecommand{\urlprefix}{URL }
\expandafter\ifx\csname urlstyle\endcsname\relax
  \providecommand{\doi}[1]{doi:\discretionary{}{}{}#1}\else
  \providecommand{\doi}{doi:\discretionary{}{}{}\begingroup
  \urlstyle{rm}\Url}\fi

\bibitem[{Akiba et~al.(2019)Akiba, Sano, Yanase, Ohta, and
  Koyama}]{akiba2019optuna}
Akiba, T.; Sano, S.; Yanase, T.; Ohta, T.; and Koyama, M. 2019.
\newblock Optuna: A next-generation hyperparameter optimization framework.
\newblock In \emph{Proceedings of the 25th ACM SIGKDD}.

\bibitem[{Antoniou, Storkey, and Edwards(2017)}]{antoniou2017data}
Antoniou, A.; Storkey, A.; and Edwards, H. 2017.
\newblock Data augmentation generative adversarial networks.
\newblock \emph{arXiv preprint arXiv:1711.04340} .

\bibitem[{Barandela et~al.(2004)Barandela, Valdovinos, S{\'a}nchez, and
  Ferri}]{barandela2004imbalanced}
Barandela, R.; Valdovinos, R.~M.; S{\'a}nchez, J.~S.; and Ferri, F.~J. 2004.
\newblock The imbalanced training sample problem: Under or over sampling?
\newblock In \emph{Joint IAPR international workshops on SPR and SSPR},
  806--814. Springer.

\bibitem[{Bengio, L{\'e}onard, and Courville(2013)}]{bengio2013estimating}
Bengio, Y.; L{\'e}onard, N.; and Courville, A. 2013.
\newblock Estimating or propagating gradients through stochastic neurons for
  conditional computation.
\newblock \emph{arXiv preprint arXiv:1308.3432} .

\bibitem[{Bruna et~al.(2013)Bruna, Zaremba, Szlam, and
  LeCun}]{bruna2013spectral}
Bruna, J.; Zaremba, W.; Szlam, A.; and LeCun, Y. 2013.
\newblock Spectral networks and locally connected networks on graphs.
\newblock \emph{arXiv preprint arXiv:1312.6203} .

\bibitem[{Chawla et~al.(2002)Chawla, Bowyer, Hall, and
  Kegelmeyer}]{chawla2002smote}
Chawla, N.~V.; Bowyer, K.~W.; Hall, L.~O.; and Kegelmeyer, W.~P. 2002.
\newblock SMOTE: synthetic minority over-sampling technique.
\newblock \emph{Journal of artificial intelligence research} 16.

\bibitem[{Chen et~al.(2019)Chen, Lin, Li, Li, Zhou, and
  Sun}]{chen2019measuring}
Chen, D.; Lin, Y.; Li, W.; Li, P.; Zhou, J.; and Sun, X. 2019.
\newblock Measuring and Relieving the Over-smoothing Problem for Graph Neural
  Networks from the Topological View.
\newblock \emph{arXiv preprint arXiv:1909.03211} .

\bibitem[{Chen, Ma, and Xiao(2018)}]{chen2018fastgcn}
Chen, J.; Ma, T.; and Xiao, C. 2018.
\newblock Fastgcn: fast learning with graph convolutional networks via
  importance sampling.
\newblock \emph{arXiv preprint arXiv:1801.10247} .

\bibitem[{Cubuk et~al.(2019)Cubuk, Zoph, Mane, Vasudevan, and
  Le}]{cubuk2019autoaugment}
Cubuk, E.~D.; Zoph, B.; Mane, D.; Vasudevan, V.; and Le, Q.~V. 2019.
\newblock Autoaugment: Learning augmentation strategies from data.
\newblock In \emph{Proceedings of the IEEE conference on CVPR}.

\bibitem[{Defferrard, Bresson, and
  Vandergheynst(2016)}]{defferrard2016convolutional}
Defferrard, M.; Bresson, X.; and Vandergheynst, P. 2016.
\newblock Convolutional neural networks on graphs with fast localized spectral
  filtering.
\newblock In \emph{NeurIPS}, 3844--3852.

\bibitem[{DeVries and Taylor(2017)}]{devries2017improved}
DeVries, T.; and Taylor, G.~W. 2017.
\newblock Improved regularization of convolutional neural networks with cutout.
\newblock \emph{arXiv preprint arXiv:1708.04552} .

\bibitem[{Edunov et~al.(2018)Edunov, Ott, Auli, and
  Grangier}]{edunov-etal-2018-bt-at-scale}
Edunov, S.; Ott, M.; Auli, M.; and Grangier, D. 2018.
\newblock Understanding Back-Translation at Scale.
\newblock In \emph{Proceedings of the 2018 Conference on EMNLP}, 489--500.

\bibitem[{Fadaee, Bisazza, and Monz(2017)}]{fadaee2017data}
Fadaee, M.; Bisazza, A.; and Monz, C. 2017.
\newblock Data augmentation for low-resource neural machine translation.
\newblock \emph{arXiv preprint arXiv:1705.00440} .

\bibitem[{Gao, Wang, and Ji(2018)}]{gao2018large}
Gao, H.; Wang, Z.; and Ji, S. 2018.
\newblock Large-scale learnable graph convolutional networks.
\newblock In \emph{Proceedings of the 24th ACM SIGKDD}, 1416--1424.

\bibitem[{Goodfellow et~al.(2014)Goodfellow, Pouget-Abadie, Mirza, Xu,
  Warde-Farley, Ozair, Courville, and Bengio}]{goodfellow2014generative}
Goodfellow, I.; Pouget-Abadie, J.; Mirza, M.; Xu, B.; Warde-Farley, D.; Ozair,
  S.; Courville, A.; and Bengio, Y. 2014.
\newblock Generative adversarial nets.
\newblock In \emph{NeurIPS}, 2672--2680.

\bibitem[{Goyal et~al.(2017)Goyal, Doll{\'a}r, Girshick, Noordhuis, Wesolowski,
  Kyrola, Tulloch, Jia, and He}]{goyal2017accurate}
Goyal, P.; Doll{\'a}r, P.; Girshick, R.; Noordhuis, P.; Wesolowski, L.; Kyrola,
  A.; Tulloch, A.; Jia, Y.; and He, K. 2017.
\newblock Accurate, large minibatch sgd: Training imagenet in 1 hour.
\newblock \emph{arXiv preprint arXiv:1706.02677} .

\bibitem[{Hamilton, Ying, and Leskovec(2017)}]{hamilton2017inductive}
Hamilton, W.; Ying, Z.; and Leskovec, J. 2017.
\newblock Inductive representation learning on large graphs.
\newblock In \emph{NeurIPS}.

\bibitem[{Han, Pei, and Kamber(2011)}]{han2011data}
Han, J.; Pei, J.; and Kamber, M. 2011.
\newblock \emph{Data mining: concepts and techniques}.
\newblock Elsevier.

\bibitem[{Henaff, Bruna, and LeCun(2015)}]{henaff2015deep}
Henaff, M.; Bruna, J.; and LeCun, Y. 2015.
\newblock Deep convolutional networks on graph-structured data.
\newblock \emph{arXiv preprint arXiv:1506.05163} .

\bibitem[{Ho et~al.(2019)Ho, Liang, Stoica, Abbeel, and
  Chen}]{ho2019population}
Ho, D.; Liang, E.; Stoica, I.; Abbeel, P.; and Chen, X. 2019.
\newblock Population based augmentation: Efficient learning of augmentation
  policy schedules.
\newblock \emph{arXiv preprint arXiv:1905.05393} .

\bibitem[{Hu et~al.(2020)Hu, Fey, Zitnik, Dong, Ren, Liu, Catasta, and
  Leskovec}]{hu2020open}
Hu, W.; Fey, M.; Zitnik, M.; Dong, Y.; Ren, H.; Liu, B.; Catasta, M.; and
  Leskovec, J. 2020.
\newblock Open graph benchmark: Datasets for machine learning on graphs.
\newblock \emph{arXiv preprint arXiv:2005.00687} .

\bibitem[{Huang, Li, and Hu(2017)}]{huang2017label}
Huang, X.; Li, J.; and Hu, X. 2017.
\newblock Label informed attributed network embedding.
\newblock In \emph{Proceedings of the Tenth ACM International Conference on
  WSDM}, 731--739.

\bibitem[{Jang, Gu, and Poole(2016)}]{jang2016categorical}
Jang, E.; Gu, S.; and Poole, B. 2016.
\newblock Categorical reparameterization with gumbel-softmax.
\newblock \emph{arXiv preprint arXiv:1611.01144} .

\bibitem[{Kafle, Yousefhussien, and Kanan(2017)}]{kafle-etal-2017-data}
Kafle, K.; Yousefhussien, M.; and Kanan, C. 2017.
\newblock Data Augmentation for Visual Question Answering.
\newblock In \emph{Proceedings of the 10th International Conference on Natural
  Language Generation}, 198--202.

\bibitem[{Kipf and Welling(2016{\natexlab{a}})}]{kipf2016semi}
Kipf, T.~N.; and Welling, M. 2016{\natexlab{a}}.
\newblock Semi-supervised classification with graph convolutional networks.
\newblock \emph{arXiv preprint arXiv:1609.02907} .

\bibitem[{Kipf and Welling(2016{\natexlab{b}})}]{kipf2016variational}
Kipf, T.~N.; and Welling, M. 2016{\natexlab{b}}.
\newblock Variational graph auto-encoders.
\newblock \emph{arXiv preprint arXiv:1611.07308} .

\bibitem[{Lemley, Bazrafkan, and Corcoran(2017)}]{lemley2017smart}
Lemley, J.; Bazrafkan, S.; and Corcoran, P. 2017.
\newblock Smart augmentation learning an optimal data augmentation strategy.
\newblock \emph{Ieee Access} 5.

\bibitem[{Levie et~al.(2018)Levie, Monti, Bresson, and
  Bronstein}]{levie2018cayleynets}
Levie, R.; Monti, F.; Bresson, X.; and Bronstein, M.~M. 2018.
\newblock Cayleynets: Graph convolutional neural networks with complex rational
  spectral filters.
\newblock \emph{IEEE Transactions on Signal Processing} 67(1): 97--109.

\bibitem[{Li et~al.(2019)Li, Muller, Thabet, and Ghanem}]{li2019deepgcns}
Li, G.; Muller, M.; Thabet, A.; and Ghanem, B. 2019.
\newblock Deepgcns: Can gcns go as deep as cnns?
\newblock In \emph{Proceedings of the IEEE ICCV}, 9267--9276.

\bibitem[{Li et~al.(2018)Li, Wang, Zhu, and Huang}]{li2018adaptive}
Li, R.; Wang, S.; Zhu, F.; and Huang, J. 2018.
\newblock Adaptive graph convolutional neural networks.
\newblock In \emph{32th AAAI}.

\bibitem[{Ma et~al.(2020)Ma, Liu, Zhao, Liu, Tang, and Shah}]{ma2020unified}
Ma, Y.; Liu, X.; Zhao, T.; Liu, Y.; Tang, J.; and Shah, N. 2020.
\newblock A Unified View on Graph Neural Networks as Graph Signal Denoising.
\newblock \emph{arXiv preprint arXiv:2010.01777} .

\bibitem[{Maddison, Mnih, and Teh(2016)}]{maddison2016concrete}
Maddison, C.~J.; Mnih, A.; and Teh, Y.~W. 2016.
\newblock The concrete distribution: A continuous relaxation of discrete random
  variables.
\newblock \emph{arXiv preprint arXiv:1611.00712} .

\bibitem[{Monti et~al.(2017)Monti, Boscaini, Masci, Rodola, Svoboda, and
  Bronstein}]{monti2017geometric}
Monti, F.; Boscaini, D.; Masci, J.; Rodola, E.; Svoboda, J.; and Bronstein,
  M.~M. 2017.
\newblock Geometric deep learning on graphs and manifolds using mixture model
  cnns.
\newblock In \emph{Proceedings of the IEEE Conference on CVPR}, 5115--5124.

\bibitem[{Niepert, Ahmed, and Kutzkov(2016)}]{niepert2016learning}
Niepert, M.; Ahmed, M.; and Kutzkov, K. 2016.
\newblock Learning convolutional neural networks for graphs.
\newblock In \emph{ICML}.

\bibitem[{Perez and Wang(2017)}]{perez2017effectiveness}
Perez, L.; and Wang, J. 2017.
\newblock The effectiveness of data augmentation in image classification using
  deep learning.
\newblock \emph{arXiv preprint arXiv:1712.04621} .

\bibitem[{Perozzi, Al-Rfou, and Skiena(2014)}]{perozzi2014deepwalk}
Perozzi, B.; Al-Rfou, R.; and Skiena, S. 2014.
\newblock Deepwalk: Online learning of social representations.
\newblock In \emph{Proceedings of the 20th ACM SIGKDD}, 701--710.

\bibitem[{Rong et~al.(2019)Rong, Huang, Xu, and Huang}]{rong2019dropedge}
Rong, Y.; Huang, W.; Xu, T.; and Huang, J. 2019.
\newblock DropEdge: Towards Deep Graph Convolutional Networks on Node
  Classification.
\newblock In \emph{ICLR}.

\bibitem[{{\c{S}}ahin and Steedman(2019)}]{csahin2019data}
{\c{S}}ahin, G.~G.; and Steedman, M. 2019.
\newblock Data Augmentation via Dependency Tree Morphing for Low-Resource
  Languages.
\newblock \emph{arXiv preprint arXiv:1903.09460} .

\bibitem[{Sennrich, Haddow, and
  Birch(2016)}]{sennrich-etal-2016-backtrasnlation}
Sennrich, R.; Haddow, B.; and Birch, A. 2016.
\newblock Improving Neural Machine Translation Models with Monolingual Data.
\newblock In \emph{Proceedings of the 54th ACL}, 86--96.

\bibitem[{Shorten and Khoshgoftaar(2019)}]{shorten2019survey}
Shorten, C.; and Khoshgoftaar, T.~M. 2019.
\newblock A survey on image data augmentation for deep learning.
\newblock \emph{Journal of Big Data} 6(1): 60.

\bibitem[{Srivastava et~al.(2014)Srivastava, Hinton, Krizhevsky, Sutskever, and
  Salakhutdinov}]{srivastava2014dropout}
Srivastava, N.; Hinton, G.; Krizhevsky, A.; Sutskever, I.; and Salakhutdinov,
  R. 2014.
\newblock Dropout: a simple way to prevent neural networks from overfitting.
\newblock \emph{The journal of machine learning research} 15(1): 1929--1958.

\bibitem[{Tang et~al.(2015)Tang, Qu, Wang, Zhang, Yan, and Mei}]{tang2015line}
Tang, J.; Qu, M.; Wang, M.; Zhang, M.; Yan, J.; and Mei, Q. 2015.
\newblock Line: Large-scale information network embedding.
\newblock In \emph{Proceedings of the 24th WWW}, 1067--1077.

\bibitem[{Veli{\v{c}}kovi{\'c} et~al.(2017)Veli{\v{c}}kovi{\'c}, Cucurull,
  Casanova, Romero, Lio, and Bengio}]{velivckovic2017graph}
Veli{\v{c}}kovi{\'c}, P.; Cucurull, G.; Casanova, A.; Romero, A.; Lio, P.; and
  Bengio, Y. 2017.
\newblock Graph attention networks.
\newblock \emph{arXiv preprint arXiv:1710.10903} .

\bibitem[{Verma et~al.(2019)Verma, Qu, Lamb, Bengio, Kannala, and
  Tang}]{verma2019graphmix}
Verma, V.; Qu, M.; Lamb, A.; Bengio, Y.; Kannala, J.; and Tang, J. 2019.
\newblock Graphmix: Regularized training of graph neural networks for
  semi-supervised learning.
\newblock \emph{arXiv preprint arXiv:1909.11715} .

\bibitem[{Wang, Cui, and Zhu(2016)}]{wang2016structural}
Wang, D.; Cui, P.; and Zhu, W. 2016.
\newblock Structural deep network embedding.
\newblock In \emph{Proceedings of the 22nd ACM SIGKDD}, 1225--1234.

\bibitem[{Wang et~al.(2020)Wang, Jiang, Syed, Conway, Juneja, Subramanian, and
  Chawla}]{wang2020calendar}
Wang, D.; Jiang, M.; Syed, M.; Conway, O.; Juneja, V.; Subramanian, S.; and
  Chawla, N.~V. 2020.
\newblock Calendar Graph Neural Networks for Modeling Time Structures in
  Spatiotemporal User Behaviors.
\newblock In \emph{Proceedings of the 26th ACM SIGKDD}.

\bibitem[{Wang, Wang, and Lian(2019)}]{wang2019survey}
Wang, X.; Wang, K.; and Lian, S. 2019.
\newblock A survey on face data augmentation.
\newblock \emph{arXiv preprint arXiv:1904.11685} .

\bibitem[{Wu, He, and Xu(2019)}]{wu2019net}
Wu, J.; He, J.; and Xu, J. 2019.
\newblock DEMO-Net: Degree-specific graph neural networks for node and graph
  classification.
\newblock In \emph{Proceedings of the 25th ACM SIGKDD}, 406--415.

\bibitem[{Wu et~al.(2019)Wu, Pan, Chen, Long, Zhang, and
  Yu}]{wu2019comprehensive}
Wu, Z.; Pan, S.; Chen, F.; Long, G.; Zhang, C.; and Yu, P.~S. 2019.
\newblock A comprehensive survey on graph neural networks.
\newblock \emph{arXiv preprint arXiv:1901.00596} .

\bibitem[{Xie et~al.(2019)Xie, Dai, Hovy, Luong, and Le}]{xie2019unsupervised}
Xie, Q.; Dai, Z.; Hovy, E.; Luong, M.-T.; and Le, Q.~V. 2019.
\newblock Unsupervised Data Augmentation for Consistency Training.
\newblock \emph{arXiv preprint arXiv:1904.12848} .

\bibitem[{Xu et~al.(2018{\natexlab{a}})Xu, Hu, Leskovec, and
  Jegelka}]{xu2018powerful}
Xu, K.; Hu, W.; Leskovec, J.; and Jegelka, S. 2018{\natexlab{a}}.
\newblock How powerful are graph neural networks?
\newblock \emph{arXiv preprint arXiv:1810.00826} .

\bibitem[{Xu et~al.(2018{\natexlab{b}})Xu, Li, Tian, Sonobe, Kawarabayashi, and
  Jegelka}]{xu2018representation}
Xu, K.; Li, C.; Tian, Y.; Sonobe, T.; Kawarabayashi, K.-i.; and Jegelka, S.
  2018{\natexlab{b}}.
\newblock Representation learning on graphs with jumping knowledge networks.
\newblock \emph{arXiv preprint arXiv:1806.03536} .

\bibitem[{Ying et~al.(2018)Ying, He, Chen, Eksombatchai, Hamilton, and
  Leskovec}]{ying2018graph}
Ying, R.; He, R.; Chen, K.; Eksombatchai, P.; Hamilton, W.~L.; and Leskovec, J.
  2018.
\newblock Graph convolutional neural networks for web-scale recommender
  systems.
\newblock In \emph{Proceedings of the 24th ACM SIGKDD}, 974--983.

\bibitem[{Yu et~al.(2020)Yu, Yu, Zhao, and Jiang}]{yu2020identifying}
Yu, W.; Yu, M.; Zhao, T.; and Jiang, M. 2020.
\newblock Identifying referential intention with heterogeneous contexts.
\newblock In \emph{Proceedings of The Web Conference}.

\bibitem[{Zhang et~al.(2019{\natexlab{a}})Zhang, Song, Huang, Swami, and
  Chawla}]{zhang2019heterogeneous}
Zhang, C.; Song, D.; Huang, C.; Swami, A.; and Chawla, N.~V.
  2019{\natexlab{a}}.
\newblock Heterogeneous graph neural network.
\newblock In \emph{Proceedings of the 25th ACM SIGKDD}, 793--803.

\bibitem[{Zhang, Zhao, and LeCun(2015)}]{zhang2016synonym}
Zhang, X.; Zhao, J.; and LeCun, Y. 2015.
\newblock Character-level Convolutional Networks for Text Classification .

\bibitem[{Zhang et~al.(2019{\natexlab{b}})Zhang, Pal, Coates, and
  Ustebay}]{zhang2019bayesian}
Zhang, Y.; Pal, S.; Coates, M.; and Ustebay, D. 2019{\natexlab{b}}.
\newblock Bayesian graph convolutional neural networks for semi-supervised
  classification.
\newblock In \emph{AAAI}, volume~33, 5829--5836.

\bibitem[{Zhang, Cui, and Zhu(2018)}]{zhang2018deep}
Zhang, Z.; Cui, P.; and Zhu, W. 2018.
\newblock Deep learning on graphs: A survey.
\newblock \emph{arXiv preprint arXiv:1812.04202} .

\bibitem[{Zhao et~al.(2019)Zhao, Balakrishnan, Durand, Guttag, and
  Dalca}]{zhao2019data}
Zhao, A.; Balakrishnan, G.; Durand, F.; Guttag, J.~V.; and Dalca, A.~V. 2019.
\newblock Data augmentation using learned transformations for one-shot medical
  image segmentation.
\newblock In \emph{Proceedings of the IEEE conference on CVPR}, 8543--8553.

\bibitem[{Zhao et~al.(2020)Zhao, Deng, Yu, Jiang, Wang, and
  Jiang}]{zhao2020error}
Zhao, T.; Deng, C.; Yu, K.; Jiang, T.; Wang, D.; and Jiang, M. 2020.
\newblock Error-Bounded Graph Anomaly Loss for GNNs.
\newblock In \emph{Proceedings of the 29th ACM International Conference on
  Information \& Knowledge Management}.

\bibitem[{Zhong et~al.(2017)Zhong, Zheng, Kang, Li, and Yang}]{zhong2017random}
Zhong, Z.; Zheng, L.; Kang, G.; Li, S.; and Yang, Y. 2017.
\newblock Random erasing data augmentation.
\newblock \emph{arXiv preprint arXiv:1708.04896} .

\end{thebibliography}

\clearpage

\appendix
\section{Proofs}
\subsection{Proof of Theorem 1}
\label{appn:proof}
We first reproduce the definition of a permutation-invariant neighborhood aggregator \cite{xu2018powerful} in the context of graph convolution:

\begin{dfn}
A neighborhood aggregator $f:\{\mathbf{X}_{v:}, v \in u\cup\mathcal{N}(u)\} \rightarrow \bar{\mathbf{X}}_{u:}$ is called \textbf{permutation-invariant} when it is invariant to the order of the target node and its neighbor nodes $u\cup\mathcal{N}(u)$, i.e. let $\{x_1, x_2, \dots, x_M\} = \{\mathbf{X}_{v:}, v \in u\cup\mathcal{N}(u)\}$, then for any permutation $\pi : f(\{x_1, x_2, \dots, x_M\}) = f(\{x_{\pi(1)}, x_{\pi(2)}, \dots, x_{\pi(M)}\})$.
\end{dfn}

Next, we prove Theorem~\ref{thm:1}:
\begin{proof}
Let $\tilde{\mathbf{A}}$ be the adjacency matrix with added self loops, i.e., $\tilde{\mathbf{A}}=\mathbf{A}+\mathbf{I}$. Here we denote the calculation process of a GNN layer with a permutation-invariant neighborhood aggregator as: 
\begin{equation}
    \mathbf{H} = f(\mathbf{A}, \mathbf{X}; \mathbf{W}) = \sigma(\bar{\mathbf{A}} \mathbf{X} \mathbf{W})
    \label{eq:gnnlayer_ap}
\end{equation}
where $\sigma$ denotes a nonlinear activation (e.g. ReLU) and $\bar{\mathbf{A}}$ denotes the normalized adjacency matrix according to the design of different GNN architectures. For example, in \gcn layer \cite{kipf2016semi}, $\bar{\mathbf{A}}=\tilde{\mathbf{D}}^{-\frac{1}{2}}\tilde{\mathbf{A}}\tilde{\mathbf{D}}^{-\frac{1}{2}}$; in \gsage layer with GCN aggregator~\cite{hamilton2017inductive}, $\bar{\mathbf{A}}$ is the row L1-normalized $\mathbf{A}$.

For any two nodes $i,j\in \mathcal{V}$ that are contained in the same fully connected component $\mathcal{S} \subseteq \mathcal{V}$, $i$ and $j$ are only connected to all other nodes in $\mathcal{S}$ by definition. Hence $\tilde{\mathbf{A}}_{iv} = \tilde{\mathbf{A}}_{jv} = 1, \forall v \in \mathcal{S}$ and $\tilde{\mathbf{A}}_{iu} = \tilde{\mathbf{A}}_{ju} = 0, \forall u \notin \mathcal{S}$, that is, $\tilde{\mathbf{A}}_{i:} = \tilde{\mathbf{A}}_{j:}$. Moreover, as the degrees of all nodes in the same fully connected component are the the same, we have $\bar{\mathbf{A}}_{i:} = \bar{\mathbf{A}}_{j:}$. Thus by Equation \ref{eq:gnnlayer_ap}, $\mathbf{H}_{i:} = \mathbf{H}_{j:}$.

On the other hand, for any two nodes $i,j\in \mathcal{V}$ that are contained in different fully connected components $\mathcal{S}_a, \mathcal{S}_b \subseteq \mathcal{V}$ respectively ($a \neq b$), $i$ and $j$ are not connected and do not share any neighbors by definition. As all nodes in $\mathcal{S}_a$ have the same degree, all nonzero entries in $\tilde{\mathbf{A}}_{i:}$ would have the same positive value $\bar{a}$. Similarly, all the nonzero entries in $\tilde{\mathbf{A}}_{j:}$ also have the same positive value $\bar{b}$. Then, by Equation \ref{eq:gnnlayer_ap}, the embeddings of $i$ and $j$ after the GNN layer will respectively be 
\begin{equation}
    \mathbf{H}_{i:} = \bar{a}\sum_{v\in\mathcal{S}_a}{\mathbf{X}_{v:}}\mathbf{W}, \qquad \mathbf{H}_{j:} = \bar{b}\sum_{u\in\mathcal{S}_b}{\mathbf{X}_{u:}}\mathbf{W}.
\end{equation}
From the above equation, we can observe that $\mathbf{H}_{i:} \neq \mathbf{H}_{j:}$ when $\mathbf{W}$ is not all zeros and $\sum_{v \in \mathcal{S}_a}{\mathbf{X}_{v:}} \neq \frac{\bar{b}}{\bar{a}} \sum_{u \in \mathcal{S}_b}{\mathbf{X}_{u:}}$.
\end{proof}

\subsection{Micro-F1 and Accuracy}
\label{appn:proof2}
\begin{dfn}
Micro-F1 score is mathematically equivalent to accuracy for classification tasks when every data point is guaranteed to be assigned to exactly one class (one ground truth label for each data point.)
\end{dfn}

\begin{proof}

The $micro$-$F1$ score is defined as following~\cite{han2011data}: 
\begin{equation}
\begin{aligned}
    &micro\mbox{-}precision = TP / (TP + FP) \\
    &micro\mbox{-}recall = TP / (TP + FN) \\
    &micro\mbox{-}F1 = 2 \cdot \frac{micro\mbox{-}precision \cdot micro\mbox{-}recall}{micro\mbox{-}precision + micro\mbox{-}recall}
\end{aligned}
\end{equation}
where $TP$, $TN$, $FP$, and $FN$ are the number of true positives, true negatives, false positives and false negatives for all classes. 

As each data object only has one label, for each mis-classified data object, one $FP$ case for the predicted class and one $FN$ case for the ground truth class are created at the same time. Therefore, $micro$-$precision$ and $micro$-$recall$ will always be the same and we then have:
\begin{equation}
    micro\mbox{-}precision = micro\mbox{-}recall = micro\mbox{-}F1
\end{equation}

Accuracy is defined as the number of correct predictions divided by the number of total cases~\cite{han2011data}. Since the number of correct predictions is the same as $TP$ and the number of incorrect predictions is $FP$, accuracy can also be calculated as following:
\begin{equation}
Accuracy = TP / (TP + FP)
\end{equation}

Thus we have:
\begin{equation}
Accuracy = micro\mbox{-}precision = micro\mbox{-}F1
\end{equation}

\end{proof}

\section{Additional Dataset Details}
\label{appn:data}
In this section, we provide some additional, relevant dataset details. The preprocessed files of all datasets used in this work can be found at \url{https://tinyurl.com/gaug-data}, including graph adjacency matrix, node features, node labels, train/validation/test node ids and predicted edge probabilities for each dataset.

\noindent \textbf{Citation networks.} \cora and \citeseer are citation networks which are used as benchmarks in most GNN-related prior works \cite{kipf2016semi, velivckovic2017graph, rong2019dropedge, chen2019measuring}. In these networks, the nodes are papers published in the field of computer science; the features are bag-of-word vectors of the corresponding paper title; the edges represent the citation relation between papers; the labels are the category of each paper.

\noindent \textbf{Protein-protein interaction network.} \ppi is the combination of multiple protein-protein interaction networks from different human tissue. The node feature contains positional gene sets, motif gene sets and immunological signatures. Gene ontology sets are used as labels (121 in total) \cite{hamilton2017inductive}. The original graph provided by \cite{hamilton2017inductive} contains total of 295 connected components in various sizes, so in this work we took the top 3 largest connected components, forming a graph with 10,076 nodes. 

\noindent \textbf{Social networks.} \blogc is an online blogging community where bloggers can follow each other,  hence forming a social network. The features for each user are generated by the keywords in each bloggers description and the labels are selected from predefined categories of blogger interests \cite{huang2017label}. \flickr is an image and video sharing platform, where users can also follow each other, hence forming a social network. The user-specified list of interest tags are used as user features and the groups that users joined are used as labels \cite{huang2017label}.

\noindent \textbf{Air traffic network.} \airusa is the airport traffic network in the USA, where each node represents an airport and edge indicates the existence of commercial flights between the airports. The node labels are generated based on the label of activity measured by people and flights passed the airports \cite{wu2019net}. The original graph does not have any features, so we used one-hot degree vectors as node features.

\section{Implementation Details \\ and Hyperparameter Tuning}
All experiments were conducted on a virtual machine on Google Cloud\footnote{\url{https://cloud.google.com/}} with 15 vCPUs, 15 Gb of RAM and one NVIDIA Tesla v100 GPU card (16 Gb of RAM at 32Gbps speed).

\subsection{Notes for effectively training \method}
\label{appn:implementation_gaug}

\noindent \textbf{Pretraining the Edge Predictor and Node Classifier.} Since the graph structure of the GNN node classifier largely depends on the edge predictor, we pretrain both components of \method to achieve more stable joint training. Otherwise, a randomly initialized edge predictor can generate very unlikely edge probabilities $\mathbf{M}$, which stunt training.  Empirically, we find that pretraining the edge predictor is more important in producing good performance compared to the node classifier, and excessive pretraining of the node classifier can lead to overfitting and poor optimizer performance.

\noindent \textbf{Learning Rate Warmup for the Edge Predictor.} 
Since the edge predictor is not only trained using $\mathcal{L}_{ep}$, but also by $\mathcal{L}_{nc}$, we adapt the learning rate warmup schema \cite{goyal2017accurate} for the edge predictor to avoid effective undoing of initial pretraining. Specifically, we initialize the edge predictor's learning rate at zero and gradually increase following a sigmoid curve. This empirically helps avoid sudden drift in edge prediction from $\mathcal{L}_{nc}$, and improves results. 
We also incorporate a parameter that narrows down a section of the sigmoid curve to specify how rapid the learning rate warms up.

\noindent \textbf{Mini-batch Training.}
For graphs that are too large and exceed GPU memory in full-batch training with \method, we follow the mini-batch training algorithm proposed by \citet{hamilton2017inductive}. For each batch, we first randomly sample a group of seed nodes from the training nodes. We then populate the batch with the union of seed nodes and their $k$-hop neighbors to form a subgraph, where $k$ is the number of layers in GNNs. Lastly, the subgraph representing the mini-batch, together with their node features is fed as input to the model. $\mathcal{L}_{ep}$ is calculated from the adjacency matrix of this subgraph; $\mathcal{L}_{nc}$ is calculated only with the predictions on seed nodes. 

Note that the sampled graph for each mini-batch would be subtly different from the one during full-batch training, as the model cannot sample any new edges between seed nodes and target nodes outside of the extended subgraph. Nevertheless, this slight difference does not affect the training much as most of the sampled new edges are within the subgraph (within a few hops).

\subsection{Hyperparameters and Search Space}
\label{appn:hyperparamter}
In this section, we describe the parameters of all methods along with the search space of all hyperparameters. All methods were implemented in \verb+Python 3.7.6+ with \verb+PyTorch+. Our implementation can be found at \url{https://github.com/zhao-tong/GAug}.  We further include code for \adaedge \cite{chen2019measuring} and \dropedge \cite{rong2019dropedge} for comparisons. The best hyperparameter choices and searching scripts can be found in the supplementary material.

\noindent \textbf{Original graph neural network architectures.} All original GNN architectures are implemented in \verb+DGL+\footnote{\url{https://www.dgl.ai/}} with Adam optimizer. We search through the basic parameters such as  learning rate and the choice of aggregators (and number of layers only for \jknet) to determine the default settings of each GNN. By default, \gcn, \gsage and \gat have 2 layers, and \jknet has 3 layers due to its unique design.  \gcn, \gsage and \jknet have hidden size 128, and \gat has a hidden size of 16 for each head (have use 8 heads). \gcn, \gsage and \jknet have learning rates of $1\mathrm{e}{-2}$ and \gat has best performance with learning rate of $5\mathrm{e}{-3}$. All methods have weight decay of $5\mathrm{e}{-4}$. \gcn, \gsage and \jknet use feature dropout of 0.5, while \gat uses both feature dropout and attention dropout of 0.6. For \gsage, we use the GCN-style aggregator. For \jknet, we use \gsage layer with GCN-style aggregator as neighborhood aggregation layers and concatenation for the final aggregation layer. To make fair comparisons, these parameters are fixed for all experiments and our hyperparameter searches only search over the new parameters introduced by baselines and our proposed methods.

\noindent \textbf{\adaedge.} We implement \adaedge \cite{chen2019measuring} based on the above mentioned GNNs and the provided pseudo-code in their paper in \verb+PyTorch+, since the author-implemented code was unavailable. We tune the following hyperparameters over ranges: $order \in \{add\_first, remove\_first\}$, $num_+ \in \{0, 1, \dots, |\mathcal{E}|-1\}$, $num_- \in \{0, 1, \dots, |\mathcal{E}|-1\}$, $conf_+ \in [0.5, 1]$, $conf_- \in [0.5, 1]$.

\begin{table*}[ht!]
\small
  \caption{Ablation study of \method on \cora}
  \label{tab:ablative_results}
  \centering
  \begin{tabular}{lcccc}
    \toprule
    Setting & \gcn & \gsage & \gat & \jknet \\
    \midrule
    Original (\gcn) & 81.6$\pm$0.7 & 81.3$\pm$0.5 & 81.3$\pm$1.1 & 78.0$\pm$1.5 \\
    +\method & \textbf{83.6$\pm$0.5} & \textbf{82.0$\pm$0.5} & \textbf{82.2$\pm$0.8} &\textbf{80.5$\pm$0.9} \\
    +\method No Sampling & 82.8$\pm$0.9 & 81.2$\pm$0.8 & 77.8$\pm$2.2 & 76.9$\pm$1.4 \\
    +\method Rounding & 82.5$\pm$0.5 & 81.4$\pm$0.5 & 81.3$\pm$1.1 & 79.5$\pm$1.3 \\
    +\method No $\mathcal{L}_{ep}$ & 82.8$\pm$0.8 & 81.5$\pm$1.1 & 81.9$\pm$0.8 & 79.5$\pm$1.0 \\
    \bottomrule
  \end{tabular}
\end{table*}

\noindent \textbf{\dropedge.} We also implement \dropedge \cite{rong2019dropedge} (adapting the authors' code for easier comparison) based on the above mentioned GNNs, where the GNNs randomly remove $p|\mathcal{E}|$ of the edges and redo the normalization on the adjacency matrix before each training epoch, where $p$ is searched in the range of $[0, 0.99]$.  

\noindent \textbf{\bgcn.}
The \bgcn model consists of two parts: an assortative mixed membership stochastic block model (MMSBM) and a GNN. For MMSBM, we use the code package\footnote{\url{https://github.com/huawei-noah/BGCN}} provided by the authors~\cite{zhang2019bayesian}. For GNNs, we use the above mentioned implementations. We follow the training process provided in the authors' code package.

\noindent \textbf{\methodtwo.} As described in Section \ref{sec:edgemanip_twostep}, \methodtwo has two hyperparameters $i$ and $j$, which are both searched within the range of $\{0, 0.01, 0.02, \dots, 0.8\}$.

\noindent \textbf{\method.} We tune the following hyperparameters over search ranges for \method. The influence of the edge predictor on the original graph: $\alpha \in \{0, 0.01, 0.02, \dots, 1\}$; the weight for $\mathcal{L}_{ep}$ when training the model: $\beta \in \{0, 0.1, 0.2, \dots, 4\}$; the temperature for the relaxed Bernoulli sampling: $temp \in \{0.1, 0.2, \dots, 2\}$; number of pretrain epochs for both edge predictor and node classifier: $n\_pretrain_{ep}, n\_pretrain_{nc} \in \{5, 10, 15, \dots, 300\}$; the parameter for warmup: $warmup \in \{0, 1, \dots, 10\}$.

\noindent \textbf{Mini-batch training.} For vanilla GNNs, we follow the standard mini-batch training proposed by \citet{hamilton2017inductive} and \citet{ying2018graph}. For GNNs with \methodtwo, both the VGAE edge predictor and GNN also followed the same mini-batch training. For GNNs with \method, we used mini-batch training as described in Appx.~\ref{appn:implementation_gaug}.

\section{Discussion and \\ Additional Experimental Results}
\label{appn:discussion}

\subsection{Ablation Study for \method}
\label{sec:ablation}
We extensively study the effect of various design choices in \method to support our decisions. Here, we compare the results of the 4 GNN architectures in combination with baseline and \method applied with different graph sampling and training choices on \cora.  The results are shown in Table \ref{tab:ablative_results}. 

\noindent \textbf{No Sampling:} Instead of interpolating $\mathbf{M}$ and $\mathbf{A}$ and subsequently sampling, we avoid the sampling step and feed the fully dense adjacency into the GNN classifier (every edge is used for convolution, with different weights).  This shows decreased  performance compared to sampling-based solution, likely because the sampling removes noise from many, very weak edges.

\noindent \textbf{Rounding:} Instead of sampling the graph, we deterministically round the edge probabilities to 0 or 1, using the same ST gradient estimator in the backward pass.  Rounding creates the same decision boundary for edge and no-edge at each epoch.  We observe that it hurts performance compared to sampling, likely due to reduced diversity during model training.  

\noindent \textbf{No Edge-Prediction Loss:} Instead of training \method with a positive $\beta$ (coefficient for $\mathcal{L}_{ep}$), we set $\beta = 0$.  Without controlling drift of pre-trained edge predictor by combining its loss($\mathcal{L}_{ep}$) with node classification loss($\mathcal{L}_{nc}$)  in training, we risk generating unrealistic graphs which arbitrarily deviate from the original graph, and producing instability in training.  We find that removing this loss term leads to empirical performance decrease.

\begin{table}[t!]
\small
  \caption{Summary statistics for the large datasets.}
  \label{tab:datasets2}
  \centering
  \begin{tabular}{lcc}
    \toprule
     & \pubmed & \ogbn  \\
    \midrule
    \# Nodes & 19,717 & 169,343  \\
    \# Edges & 44,338 &  1,166,243 \\
    \# Features & 500 & 128 \\
    \# Classes & 3 & 40 \\
    \# Training nodes & 60 & 90,941  \\
    \# Validation nodes & 500 & 29,799  \\
    \# Test nodes & 1000 & 48,603  \\
    \bottomrule
  \end{tabular}
\end{table}

\begin{table}[t!]
\small
  \caption{\methodshared performance with mini-batch training.}
  \label{tab:large_graph}
  \centering
  \begin{tabular}{lcc}
    \toprule
    Methods & \pubmed & \ogbn \\
    \midrule
    \gcn & 78.5$\pm$0.5 & 68.1$\pm$0.3 \\
    \gcn + \methodtwo & \textbf{80.2$\pm$0.3} & 68.2$\pm$0.3 \\
    \gcn + \method & 79.3$\pm$0.4 & \textbf{71.4$\pm$0.5} \\
    \bottomrule
  \end{tabular}
\end{table}

\subsection{\methodshared with Mini-batch Training}
\label{appn:exp_large}
In order to better show the scalability of the proposed methods with mini-batch training, Table~\ref{tab:large_graph} shows that the proposed \methodshared methods are able to perform well when using mini-batch training. More specifically, on \pubmed, augmentation (via \methodtwo) achieves 2.2\% improvement, while on \ogbn, augmentation (via \method) achieves 4.8\% improvement. 
All three methods are trained in the mini-batch setting with the same batch size. Note that the performance using mini-batch training is generally not as good as full-batch training (even for vanilla GNNs), 
hence mini-batch training is only recommended when graphs are too large to fit in GPU as a whole.

\begin{table*}[h!]
\small
  \caption{\methodtwo performance on original and modified graphs. }
  \label{tab:results_orig_mod}
  \centering
  \begin{tabular}{clcccccc}
    \toprule
    Backbone & Method & \cora & \citeseer & \ppi & \textsc{BlogC} & \flickr & \airusa \\
    \midrule
    \multirow{3}*{\gcn} 
    & original & 81.6$\pm$0.7 & 71.6$\pm$0.4 & 43.4$\pm$0.2 & 75.0$\pm$0.4 & 61.2$\pm$0.4 & 56.0$\pm$0.8 \\
    & +\methodtwo & \textbf{83.5$\pm$0.4} & 72.3$\pm$0.4 & \textbf{43.5$\pm$0.2} & \textbf{77.6$\pm$0.4} & \textbf{68.2$\pm$0.7} & \textbf{61.2$\pm$0.5} \\
    & +\methodtwo-\textit{O} & 83.1$\pm$0.5 & \textbf{72.8$\pm$0.5} & \textbf{43.5$\pm$0.2} & 75.6$\pm$0.4 & 61.6$\pm$0.6 & 58.1$\pm$0.6 \\
    \midrule
    \multirow{3}*{\gsage} 
    & Original & 81.3$\pm$0.5 & 70.6$\pm$0.5 & 40.5$\pm$0.9 & 73.4$\pm$0.4 & 57.4$\pm$0.5 & 57.0$\pm$0.7 \\
    & +\methodtwo & \textbf{83.2$\pm$0.4} & 71.2$\pm$0.4 & \textbf{41.1$\pm$1.0} & \textbf{77.0$\pm$0.4} & \textbf{65.2$\pm$0.4} & \textbf{60.1$\pm$0.5} \\
    & +\methodtwo-\textit{O} & 82.4$\pm$0.5 & \textbf{71.6$\pm$0.3} & 41.1$\pm$1.4 & 74.3$\pm$0.3 & 58.1$\pm$0.5 & 58.9$\pm$0.5 \\
    \midrule
    \multirow{3}*{\gat} 
    & Original & 81.3$\pm$1.1 & 70.5$\pm$0.7 & 41.5$\pm$0.7 & 63.8$\pm$5.2 & 49.6$\pm$1.6 & 52.0$\pm$1.3 \\
    & +\methodtwo & \textbf{82.1$\pm$1.0} & \textbf{71.5$\pm$0.5} & 42.8$\pm$0.9 & 70.8$\pm$1.0 & \textbf{63.7$\pm$0.9} & \textbf{59.0$\pm$0.6} \\
    & +\methodtwo-\textit{O} & 82.0$\pm$0.9 & 71.3$\pm$0.7 & \textbf{46.3$\pm$0.2} & \textbf{71.0$\pm$1.3} & 48.5$\pm$1.9 & 53.4$\pm$1.1 \\
    \midrule
    \multirow{3}*{\jknet} 
    & Original & 78.8$\pm$1.5 & 67.6$\pm$1.8 & 44.1$\pm$0.7 & 70.0$\pm$0.4 & 56.7$\pm$0.4 & 58.2$\pm$1.5 \\
    & +\methodtwo & \textbf{81.8$\pm$0.9} & 68.2$\pm$1.4 & 47.4$\pm$0.6 & \textbf{71.9$\pm$0.5} & \textbf{65.7$\pm$0.8} & \textbf{60.2$\pm$0.6} \\
    & +\methodtwo-\textit{O} & 80.6$\pm$1.0 & \textbf{68.3$\pm$1.4} & \textbf{48.6$\pm$0.5} & 71.0$\pm$0.4 & 57.0$\pm$0.4 & 60.2$\pm$0.8 \\
    \bottomrule
  \end{tabular}
\end{table*}

\begin{figure*}[ht!]
    \centering
        \begin{subfigure}[b]{.2\linewidth}
          \includegraphics[width=\linewidth]{figs/cora-gae.pdf}
            \caption{GAE}
        \end{subfigure}
        \begin{subfigure}[b]{.2\linewidth}
          \includegraphics[width=\linewidth]{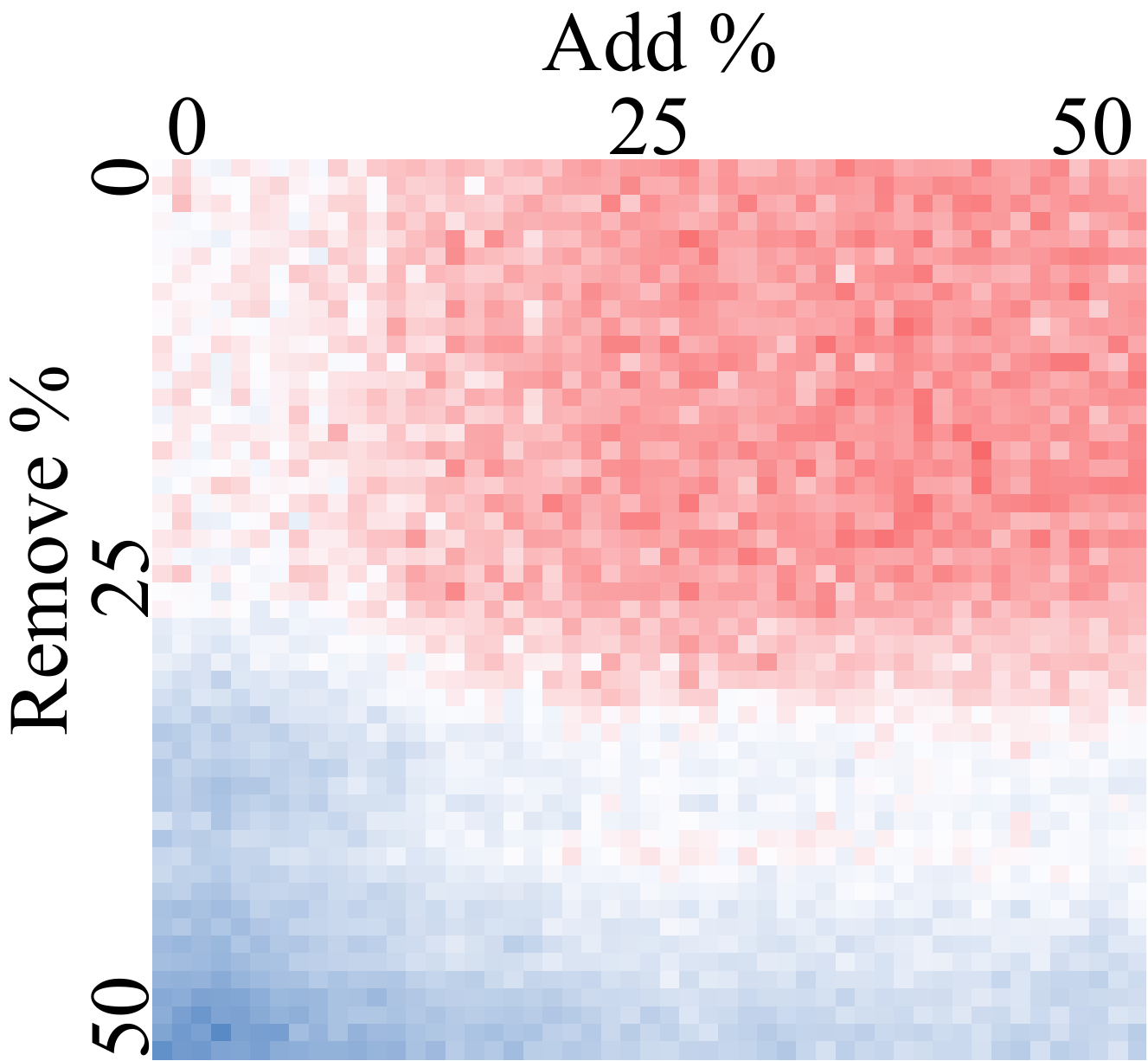}
            \caption{VGAE}
        \end{subfigure}
        \begin{subfigure}[b]{.2\linewidth}
            \includegraphics[width=\textwidth]{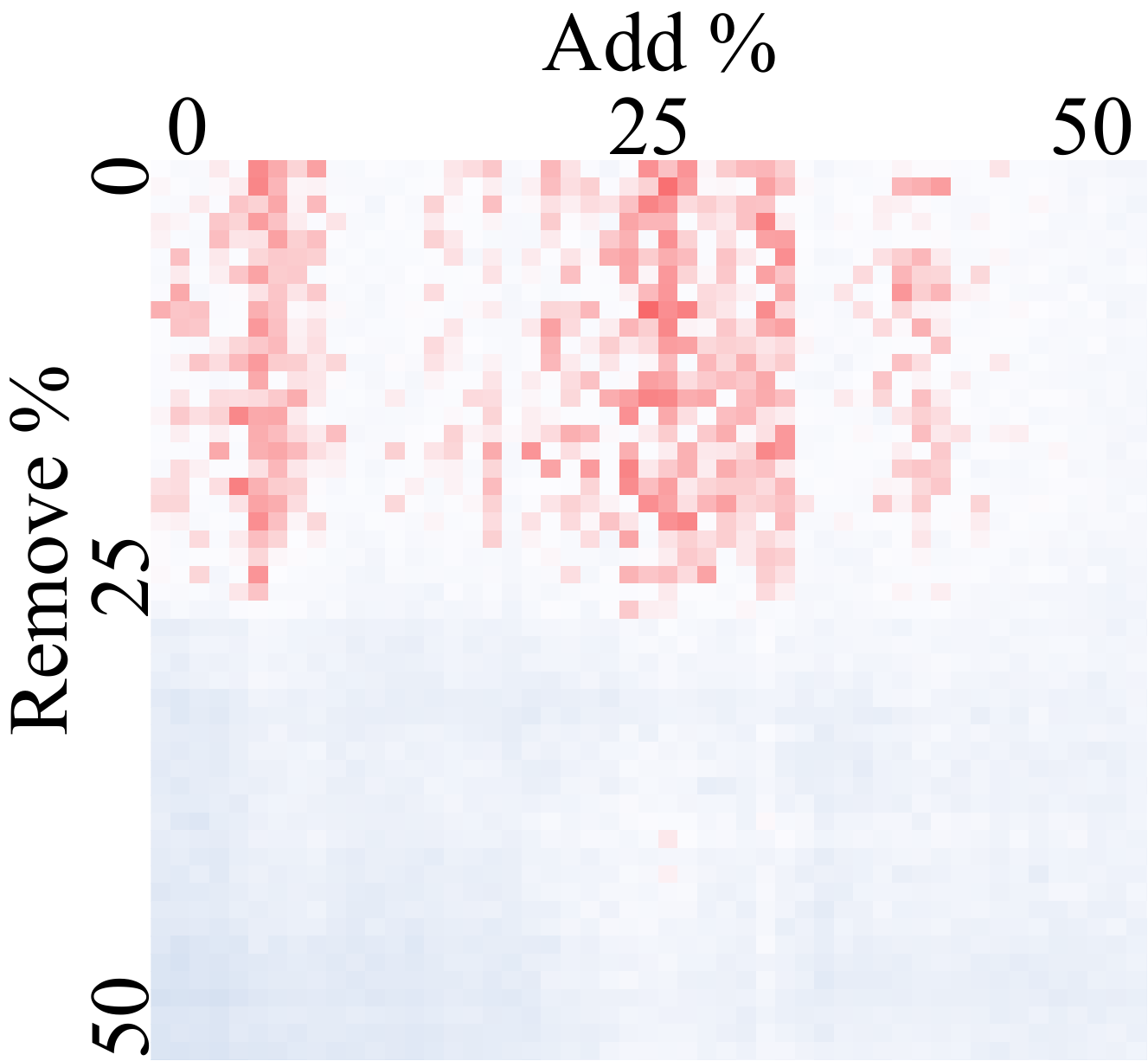}
            \caption{LLHN}
        \end{subfigure}
        \begin{subfigure}[b]{.2\linewidth}
            \includegraphics[width=\textwidth]{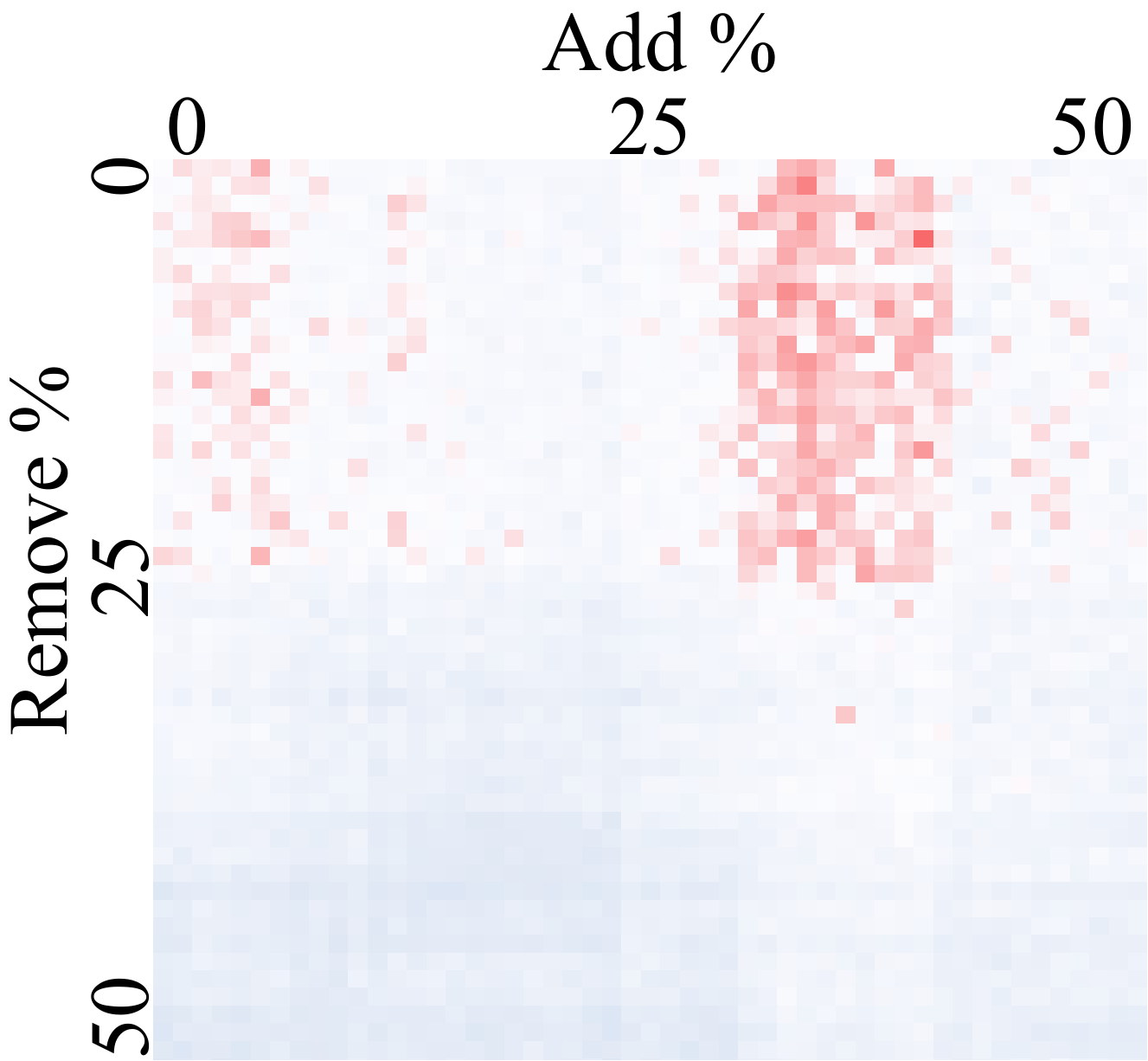}
            \caption{RA}
        \end{subfigure}
        \begin{subfigure}[b]{.2\linewidth}
          \includegraphics[width=\linewidth]{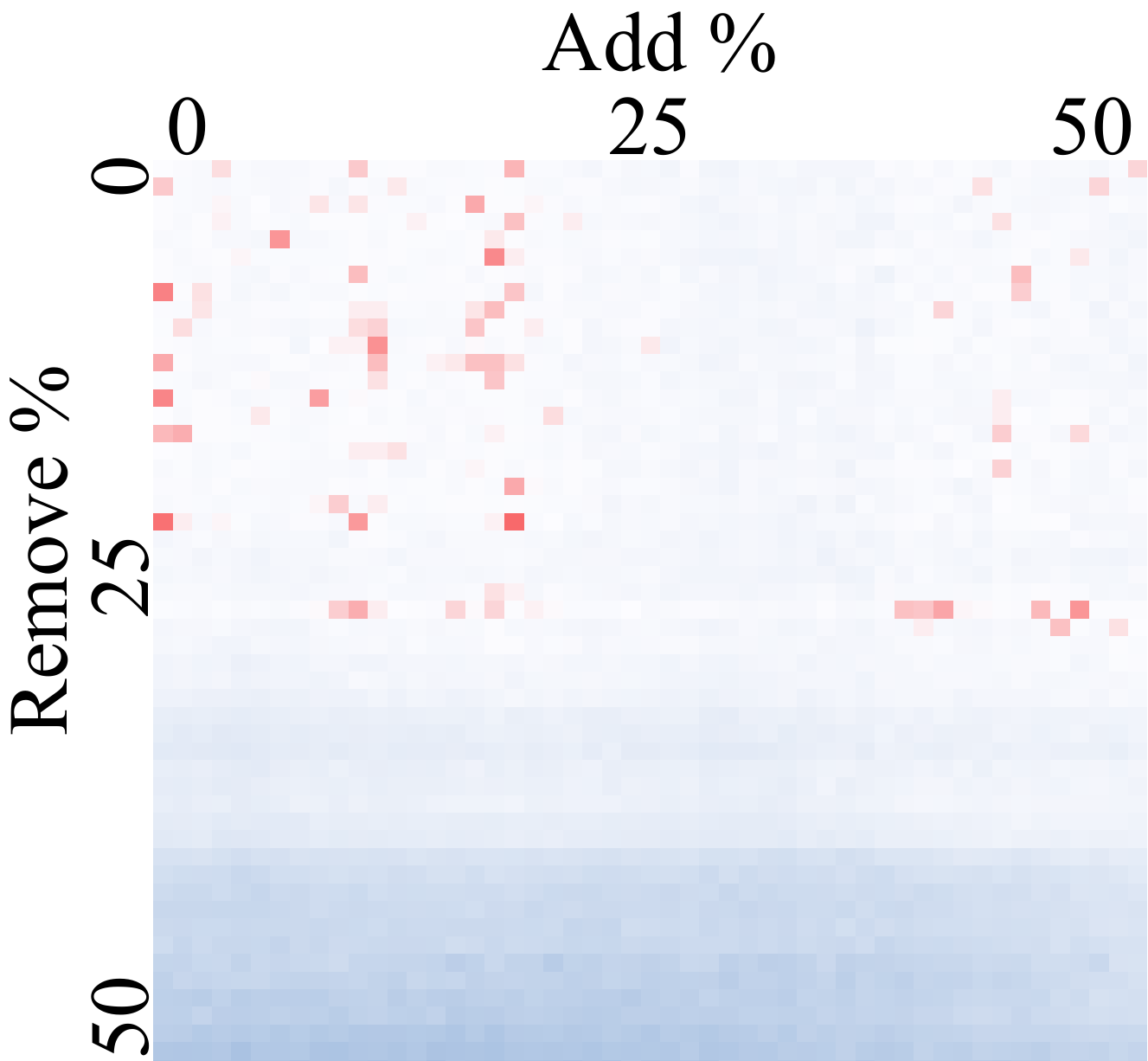}
            \caption{CAR}
        \end{subfigure}
        \begin{subfigure}[b]{.2\linewidth}
            \includegraphics[width=\textwidth]{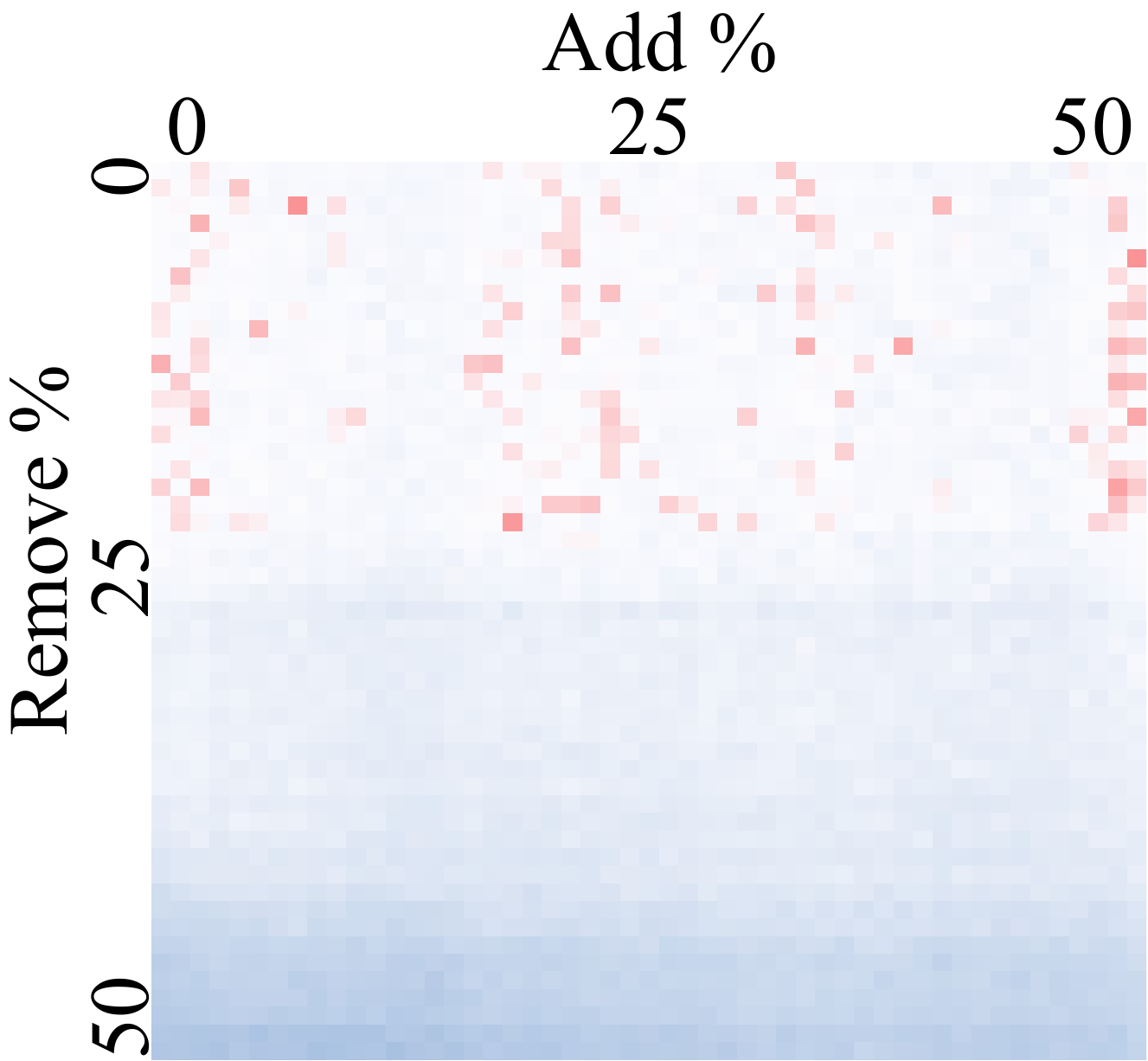}
            \caption{LNB}
        \end{subfigure}
        \begin{subfigure}[b]{.2\linewidth}
            \includegraphics[width=\textwidth]{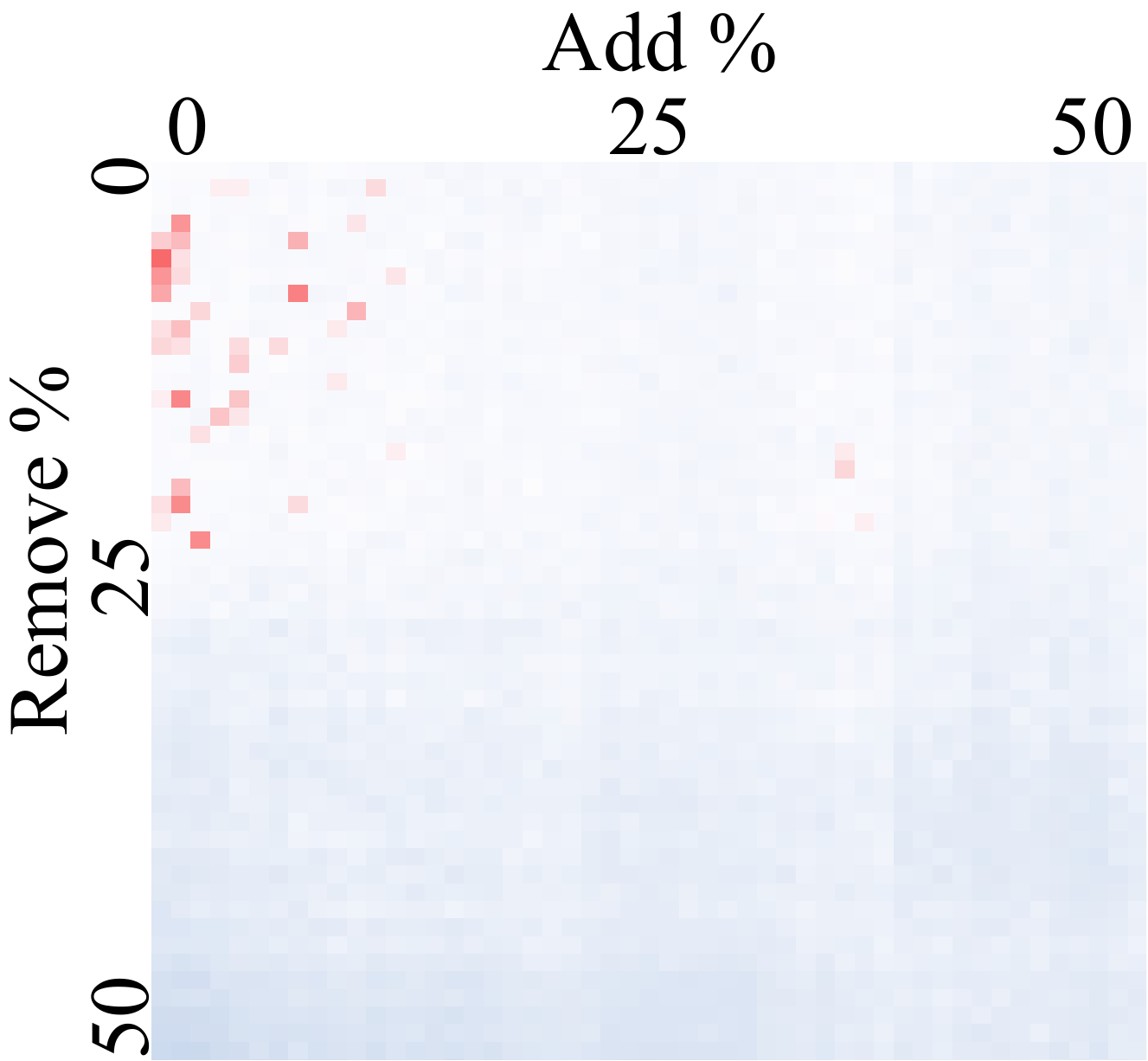}
            \caption{JI}
        \end{subfigure}
    \caption{\methodtwo with \gcn on \cora with different edge prediction heuristics}\label{fig:heatmaps-cora}
\end{figure*}

Similar to \cora and \citeseer, both of the two large graph datasets are citation networks: \pubmed is a commonly used GNN benchmark~\cite{kipf2016semi}, and \ogbn is a standard benchmark provided by the Open Graph Benchmark\footnote{https://ogb.stanford.edu/}~\cite{hu2020open}. Table~\ref{tab:datasets2} summarizes their statistics.

\subsection{Evaluating \methodtwo on original vs. modified graph}
Although \methodtwo is designed for the modified-graph setting, it is still  possible to do inference on $\mathcal{G}$ while training the model on $\mathcal{G}_m$. As previously mentioned in Section \ref{sec:3-3}, inference with \methodtwo on $G$ would result in a train-test gap, which would affect the test performance. Table \ref{tab:results_orig_mod} presents the inference results of \methodtwo with $\mathcal{G}_m$ and $\mathcal{G}$ (\methodtwo-\textit{O} makes inference on $\mathcal{G}$). We can observe that both variants of \methodtwo show performance improvements over the original GNNs across different architectures and datasets. Moreover, inference with \methodtwo on $\mathcal{G}_m$ has equal or better performance in almost cases, which aligns with our intuition in Section \ref{sec:3-3}.  This suggests that \methodtwo actually improves training in a way that helps generalization even on the original graph by better parameter inference, despite modifying the graph during training to achieve this.

\begin{table*}[t!]
\small
  \caption{\methodshared performance for deeper GNNs.}
  \label{tab:n_layers}
  \centering
  \begin{tabular}{clcccc}
    \toprule
    \# of layers & Method & \gcn & \gsage & \gat & \jknet \\
    \midrule
    \multirow{3}*{default} 
    & Original & 81.6$\pm$0.7 & 81.3$\pm$0.5 & 81.3$\pm$1.1 & 78.0$\pm$1.5 \\
    & +\methodtwo & 83.5$\pm$0.4 & {\bf 83.2$\pm$0.4} & 82.1$\pm$1.0 & {\bf 81.8$\pm$0.9} \\
    & +\method & {\bf 83.6$\pm$0.5} & 82.0$\pm$0.5 & {\bf 82.2$\pm$0.8} & 80.5$\pm$0.9 \\
    \midrule
    \multirow{3}*{4 layers} 
    & Original & 74.7$\pm$2.7 & 78.9$\pm$1.4 & 79.8$\pm$1.0 & 79.6$\pm$1.6 \\
    & +\methodtwo & 78.9$\pm$1.0 & {\bf 81.5$\pm$0.8} & {\bf 81.4$\pm$0.8} & {\bf 81.9$\pm$1.2} \\
    & +\method & {\bf 80.6$\pm$0.9} & 80.1$\pm$1.0 & 75.3$\pm$2.1 & 80.9$\pm$0.9 \\
    \midrule
    \multirow{3}*{6 layers} 
    & Original & 57.2$\pm$9.6 & 77.7$\pm$1.3 & 77.8$\pm$1.5 & 79.7$\pm$1.0 \\
    & +\methodtwo & 74.2$\pm$2.4 & {\bf 80.7$\pm$1.1} & {\bf 79.2$\pm$0.8} & {\bf 82.0$\pm$0.8} \\
    & +\method & {\bf 79.8$\pm$1.0} & 79.8$\pm$0.7 & 13.6$\pm$2.8 & 80.9$\pm$0.6 \\
    \midrule
    \multirow{3}*{8 layers} 
    & Original & 25.0$\pm$4.7 & 61.7$\pm$9.9 & 65.0$\pm$6.4 & 79.2$\pm$1.5 \\
    & +\methodtwo & 56.4$\pm$5.7 & {\bf 78.1$\pm$2.0} & {\bf 77.9$\pm$1.5} & {\bf 82.1$\pm$0.8} \\
    & +\method & {\bf 77.4$\pm$1.9} & 76.7$\pm$1.5 & 13.0$\pm$0.0 & 81.1$\pm$1.1 \\
    \bottomrule
  \end{tabular}
\end{table*}

\begin{figure*}[t!]
    \centering
        \begin{subfigure}[b]{.23\linewidth}
          \includegraphics[width=\linewidth]{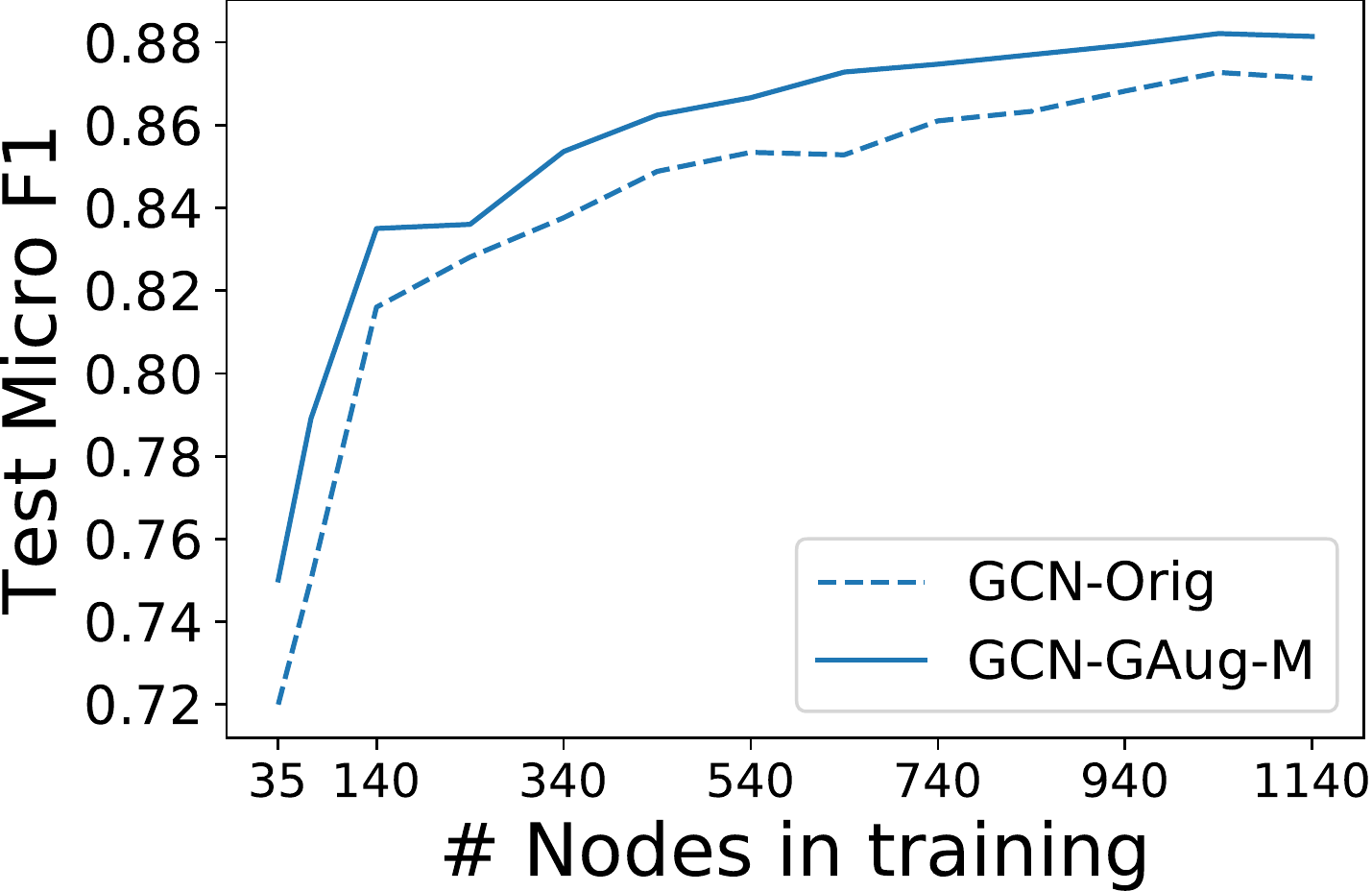}
            \label{fig:}
        \end{subfigure}
        \begin{subfigure}[b]{.23\linewidth}
          \includegraphics[width=\linewidth]{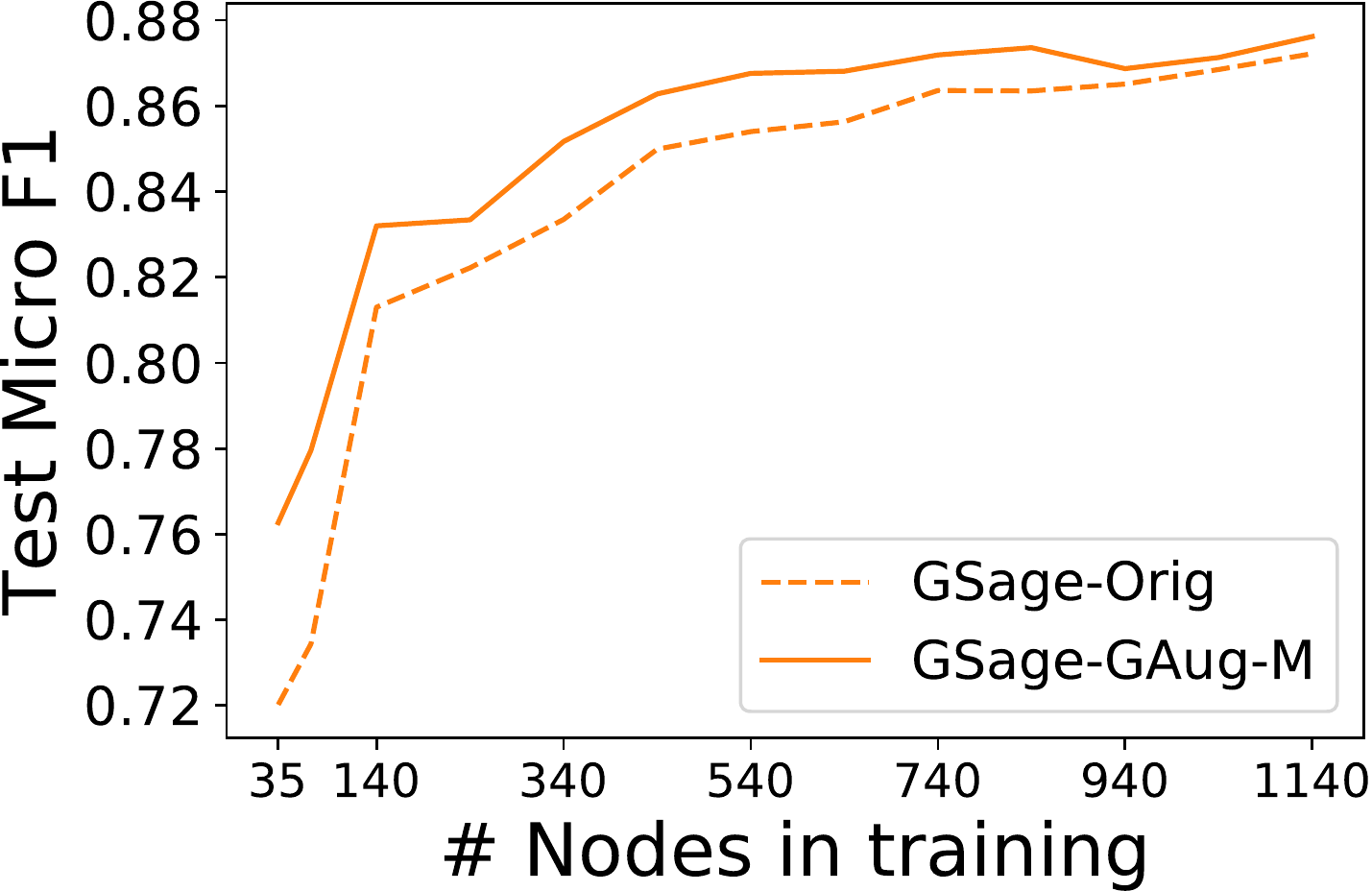}
            \label{fig:}
        \end{subfigure}
        \begin{subfigure}[b]{.23\linewidth}
            \includegraphics[width=\textwidth]{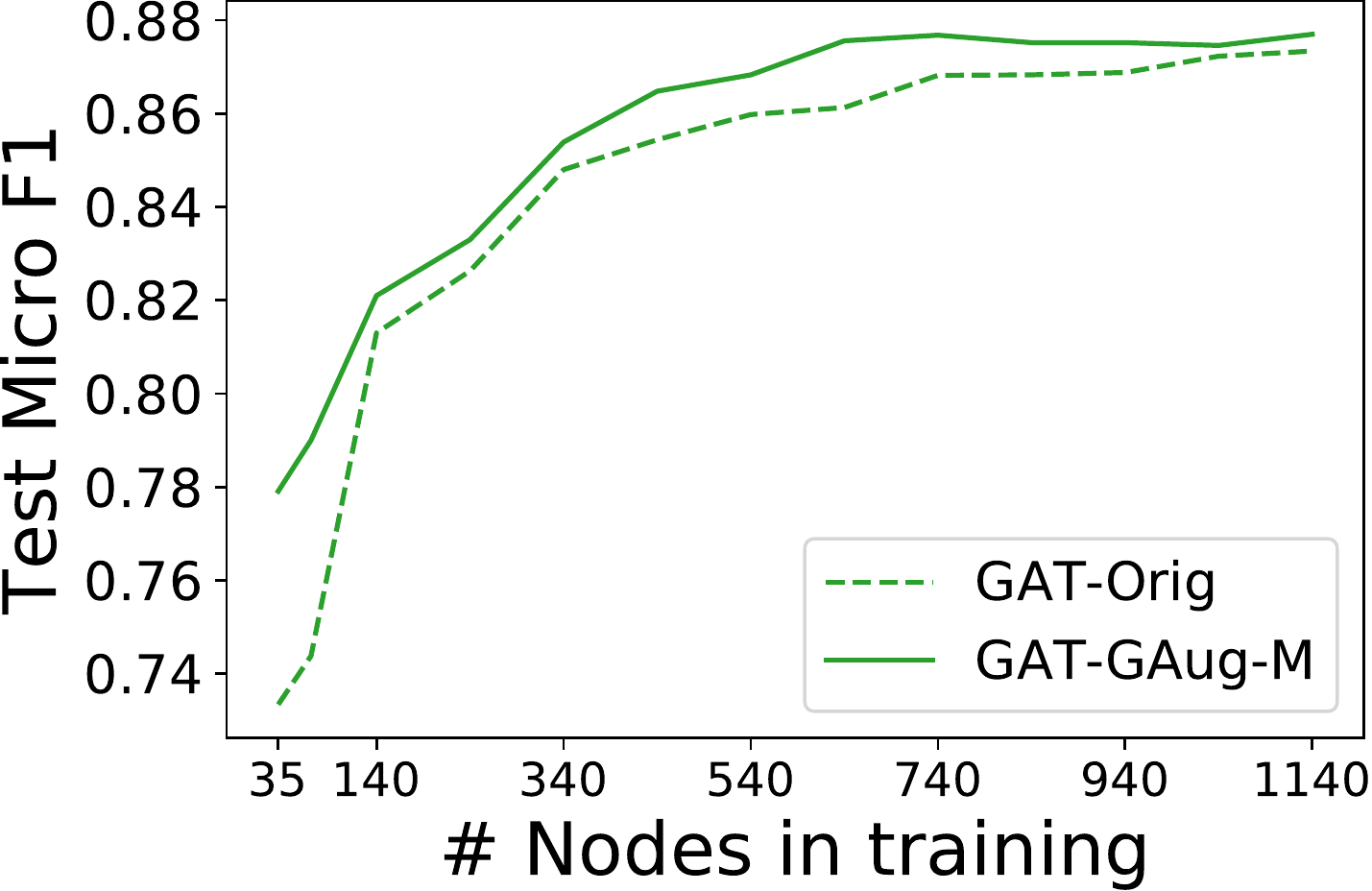}
            \label{fig:}
        \end{subfigure}
        \begin{subfigure}[b]{.23\linewidth}
            \includegraphics[width=\textwidth]{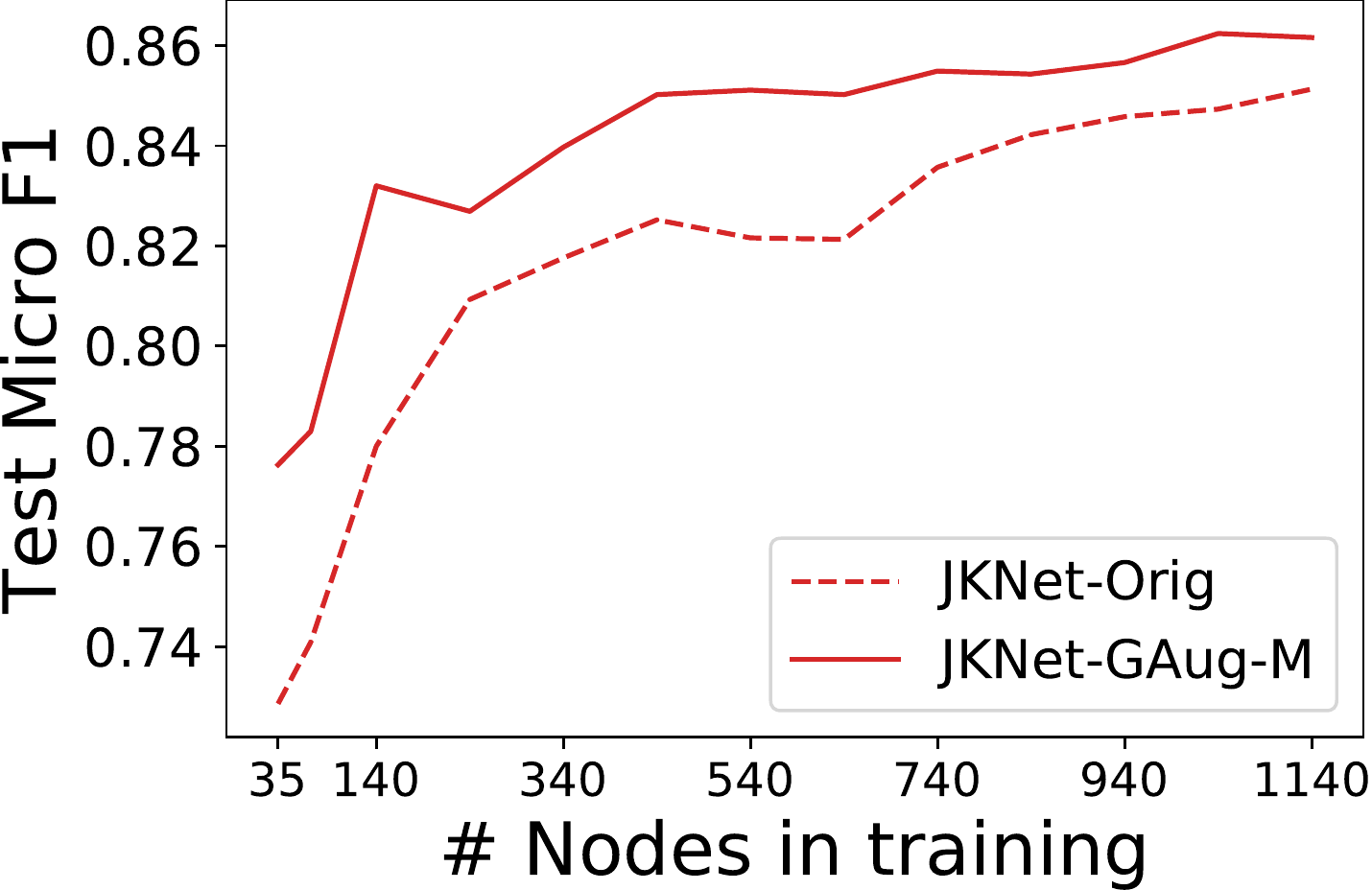}
            \label{fig:}
        \end{subfigure}
        \begin{subfigure}[b]{.23\linewidth}
          \includegraphics[width=\linewidth]{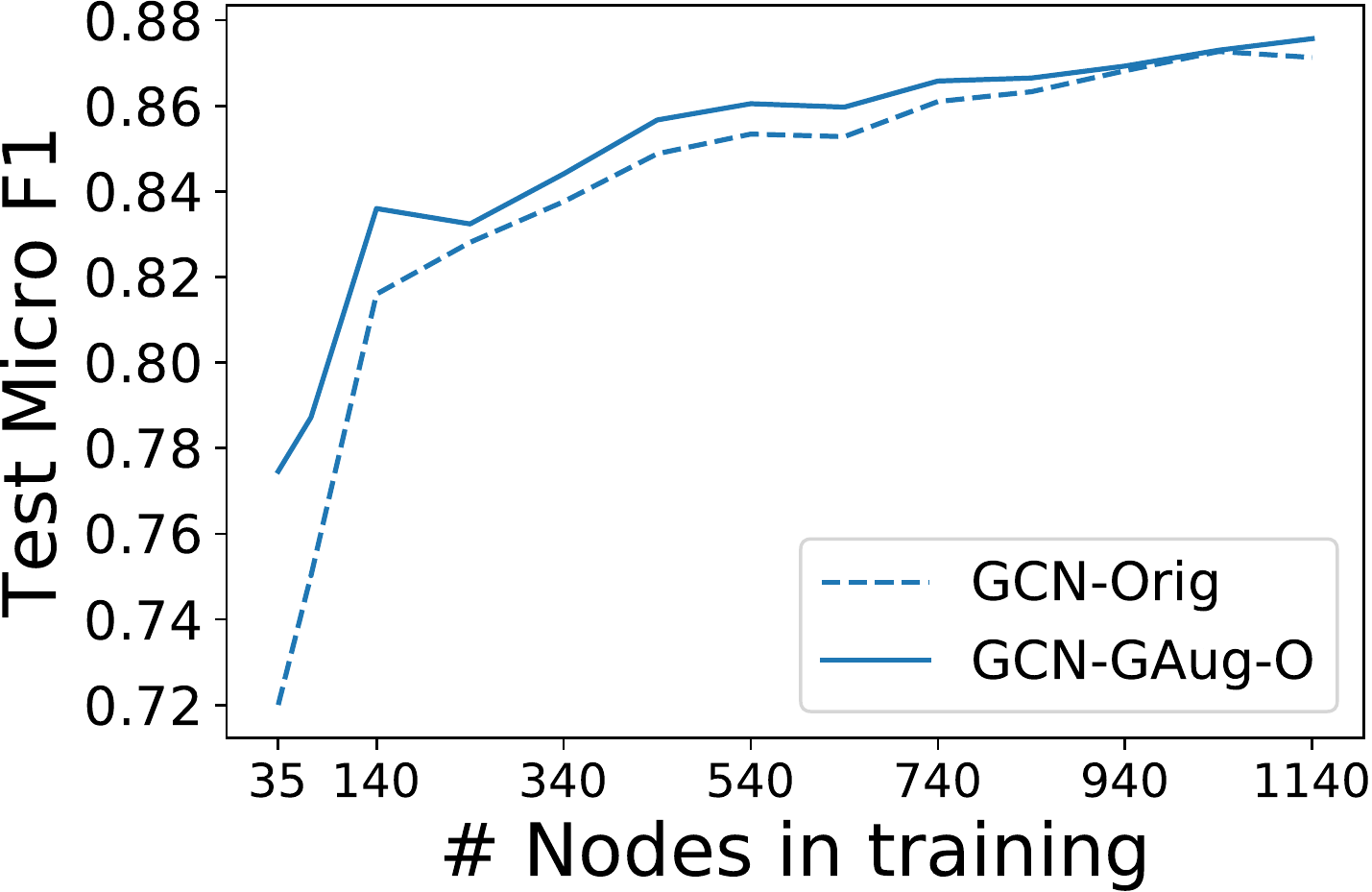}
            \label{fig:}
        \end{subfigure}
        \begin{subfigure}[b]{.23\linewidth}
          \includegraphics[width=\linewidth]{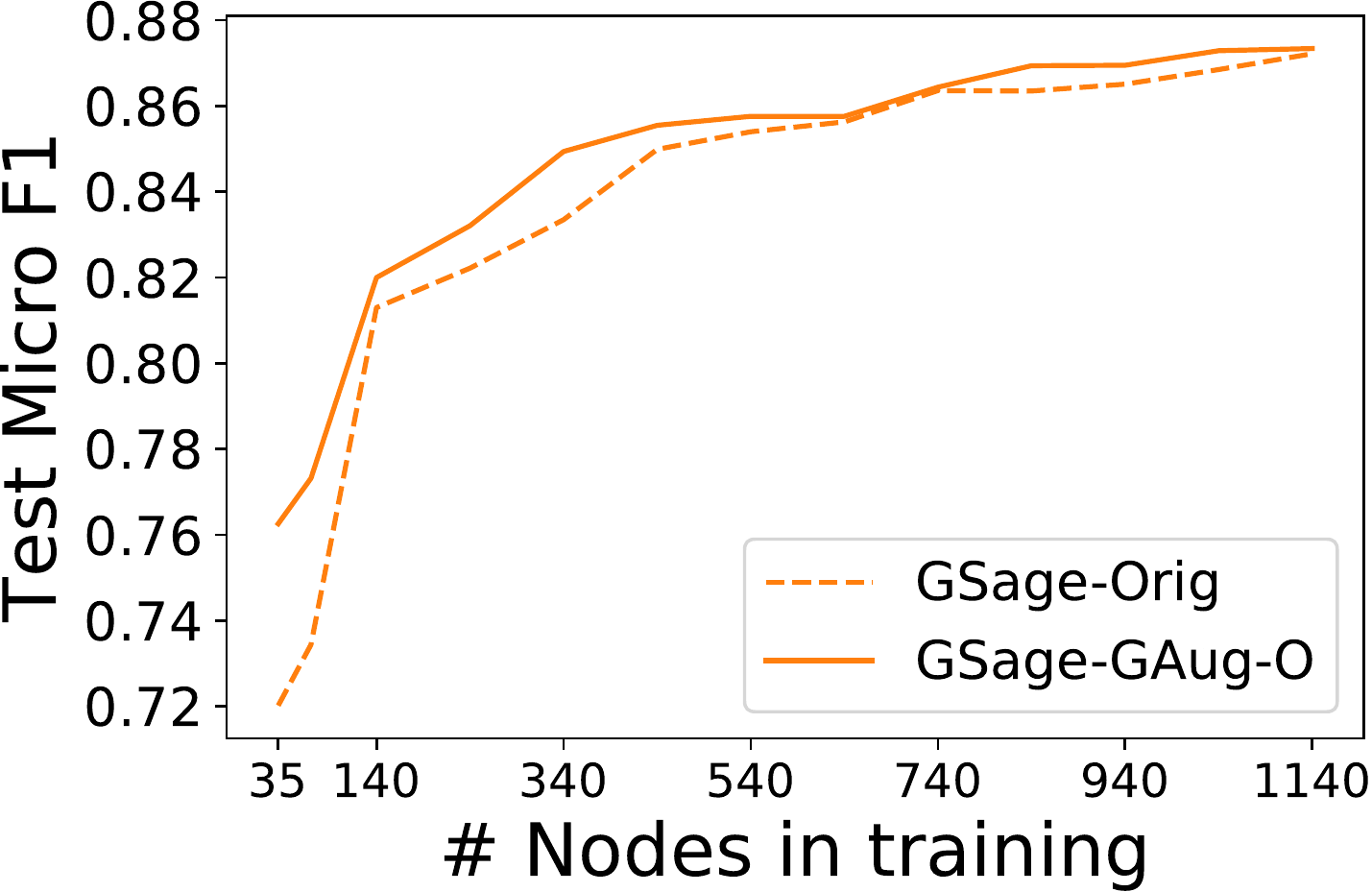}
            \label{fig:}
        \end{subfigure}
        \begin{subfigure}[b]{.23\linewidth}
            \includegraphics[width=\textwidth]{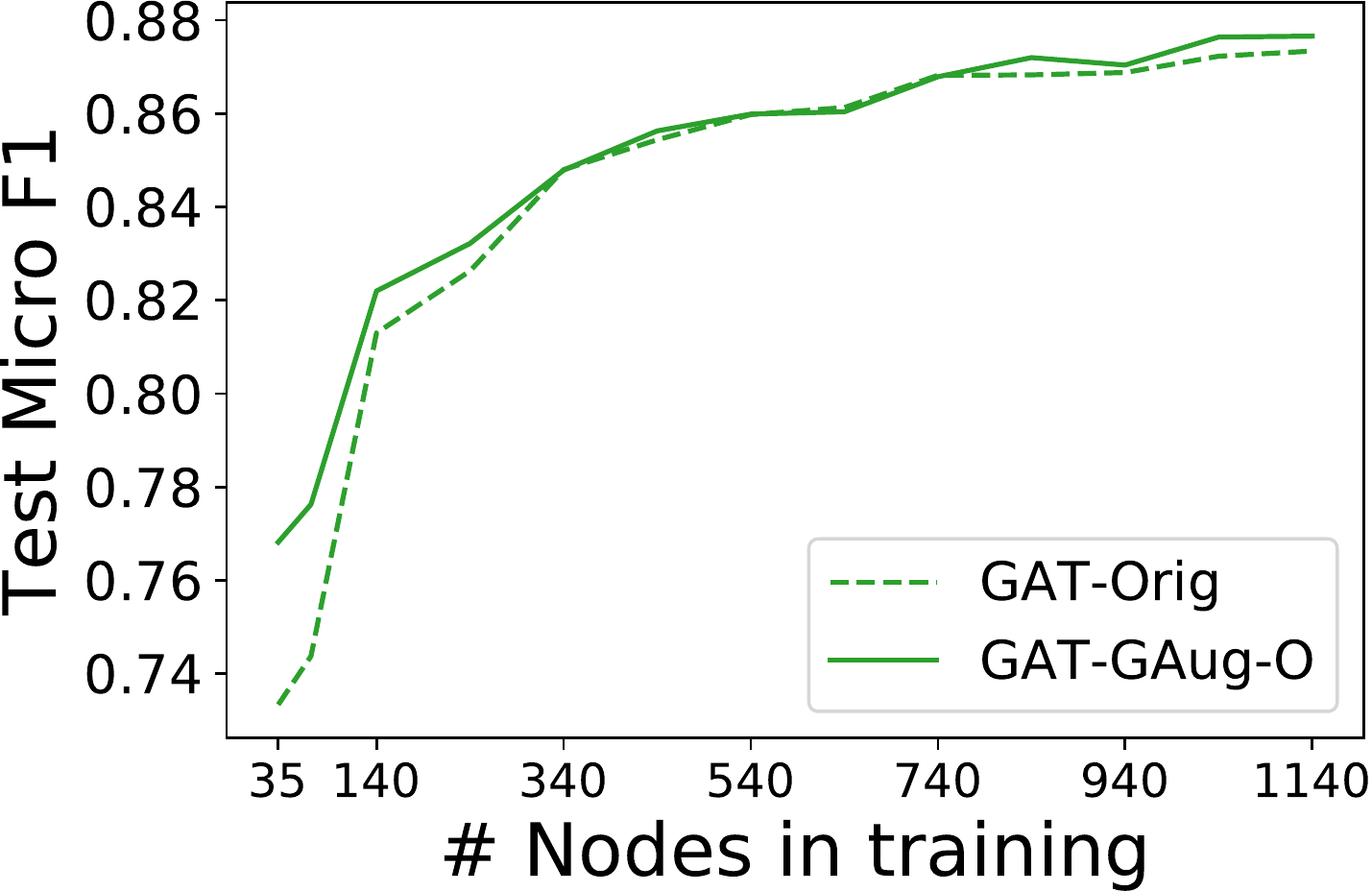}
            \label{fig:}
        \end{subfigure}
        \begin{subfigure}[b]{.23\linewidth}
            \includegraphics[width=\textwidth]{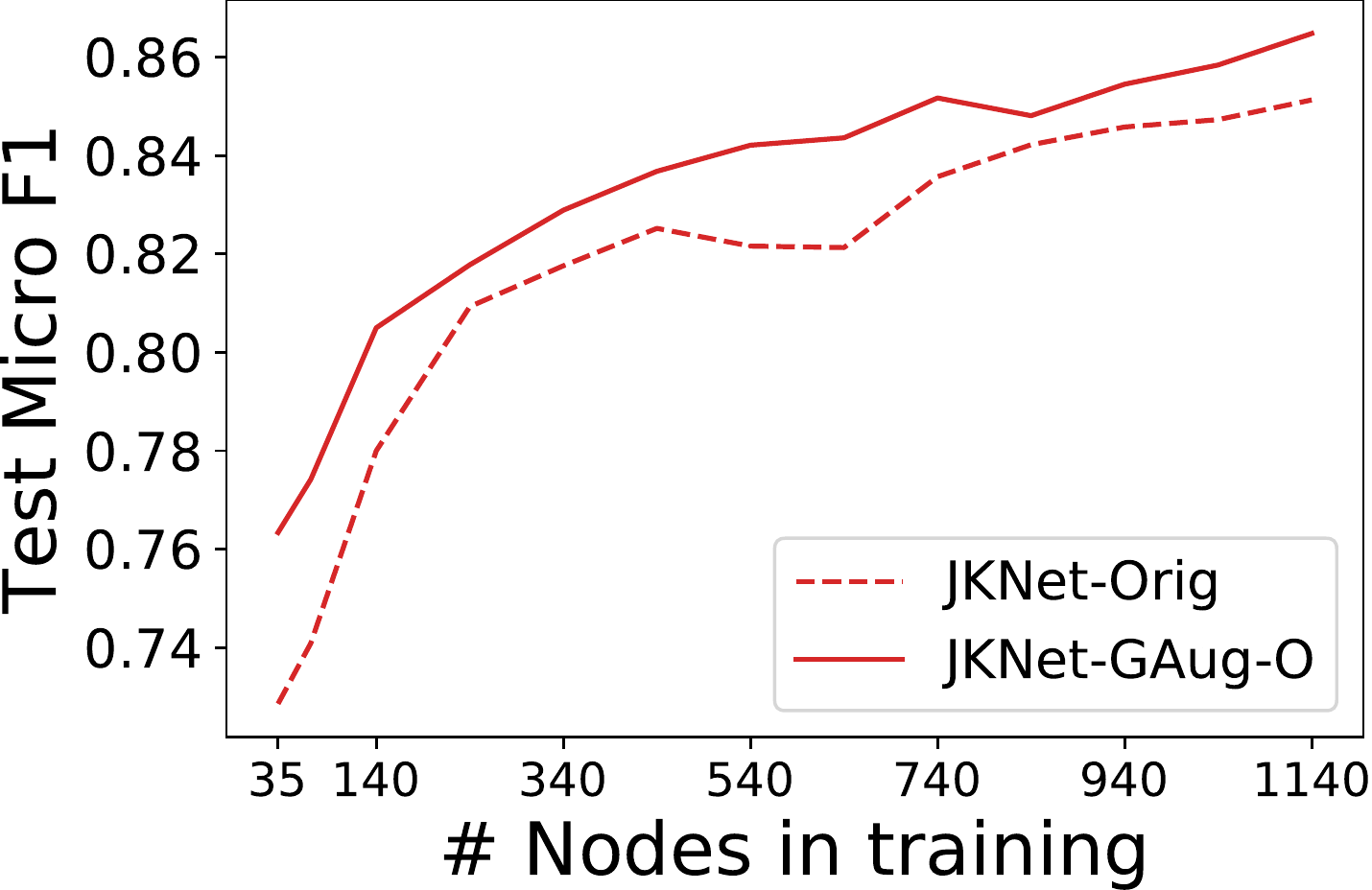}
            \label{fig:}
        \end{subfigure}
    \caption{\methodshared improves performance under weak supervision with each GNN (\gcn, \gsage, \gat and \jknet, left to right) and across augmentation settings (\methodtwo on top, \method on bottom).  Relative improvement is clear even with many training nodes, but is larger with few training nodes.}\label{fig:super_full}
\end{figure*}

\subsection{\methodtwo performance under different edge predictors} 
Although we use GAE as the edge prediction model of choice due to its strong performance in link prediction, \methodtwo can be generally equipped with any edge prediction module. 
In Figure \ref{fig:heatmaps-cora} we show classification performance heatmaps of \methodtwo (with \gcn) on \cora, when adding/removing edges according to different heuristics. Specifically, graph auto-encoder (GAE) \cite{kipf2016variational}, variational graph auto-encoder (VGAE) \cite{kipf2016variational}, the Local Leicht-Holme-Newman Index (LLHN), the Resource Allocation Index (RA), CAR-based Indices (CAR), Local Naive Bayes (LNB) and the Jaccard Index (JI). The first two are \gcn based neural auto-encoder models and the later two are edge prediction methods based on local neighborhoods, which are commonly used and celebrated in network science literature. It is noticeable that even with the same dataset, the performance heatmaps are characteristically different  when using various edge prediction methods, demonstrating the relative importance of edge predictor to \methodtwo's augmentation performance.  Moreover, it supports our findings on the importance strategic edge addition and removal to improve performance of graph augmentation/regularization based methods -- as Table \ref{tab:results} shows, careless edge addition/removal can actually hurt performance.  We do not show results when equipping \method with these different edge predictors, since \method requires the edge predictor to be differentiable for training (hence our choice of GAE).

\subsection{Classification Performance with Deeper GNNs}
In Table \ref{tab:n_layers} we show the performance of our proposed \methodshared framework with different number of layers on the \cora dataset. As mentioned in Appendix \ref{appn:hyperparamter}, \gcn, \gsage and \gat have 2 layers by default, while \jknet has 3 layers due to it's unique design. From Table \ref{tab:n_layers} we can observe that when increasing the number of layers, most GNNs perform worse except for \jknet, which is specifically designed for deep GNNs. \methodtwo shows stable performance improvements over all GNN architectures with different depth; \method shows performance improvements on \gcn, \gsage and \jknet with different depth.  Our results suggest that augmentation can be a tool which facilitates deeper GNN training, as performance improvements for the common practical GNN implementations (\gcn and \gsage) demonstrate quite large performance improvements when compared to standard implementations (e.g. 52.4 point absolute F1 improvement for \gcn, 16.4 point absolute F1 improvement for \gsage at 8 layers).

\begin{figure*}[t!]
    \centering
        \begin{subfigure}[b]{.19\linewidth}
          \includegraphics[width=\linewidth]{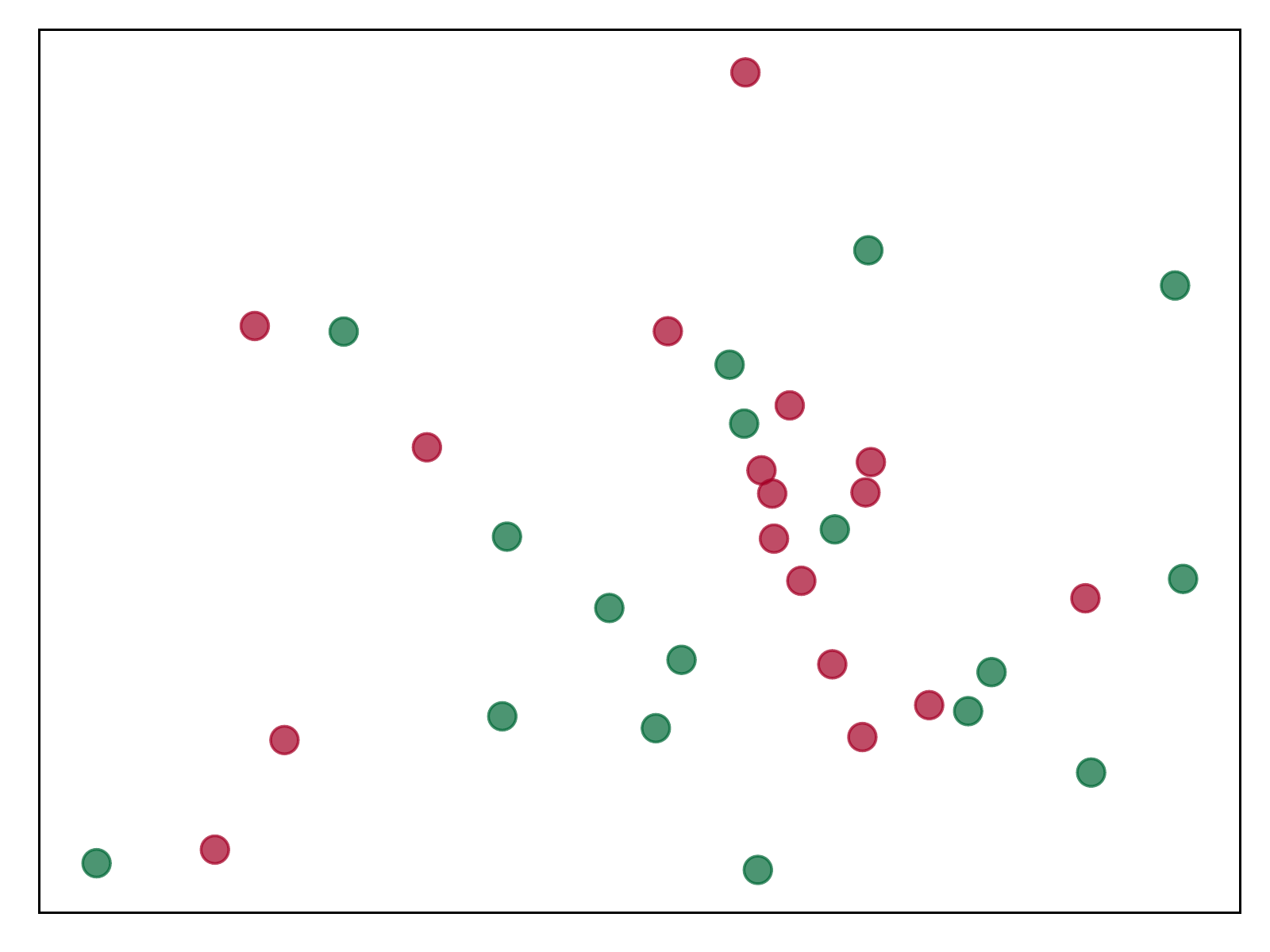}
            \caption{Raw features}\label{fig:emb_raw}
        \end{subfigure}
        \begin{subfigure}[b]{.19\linewidth}
          \includegraphics[width=\linewidth]{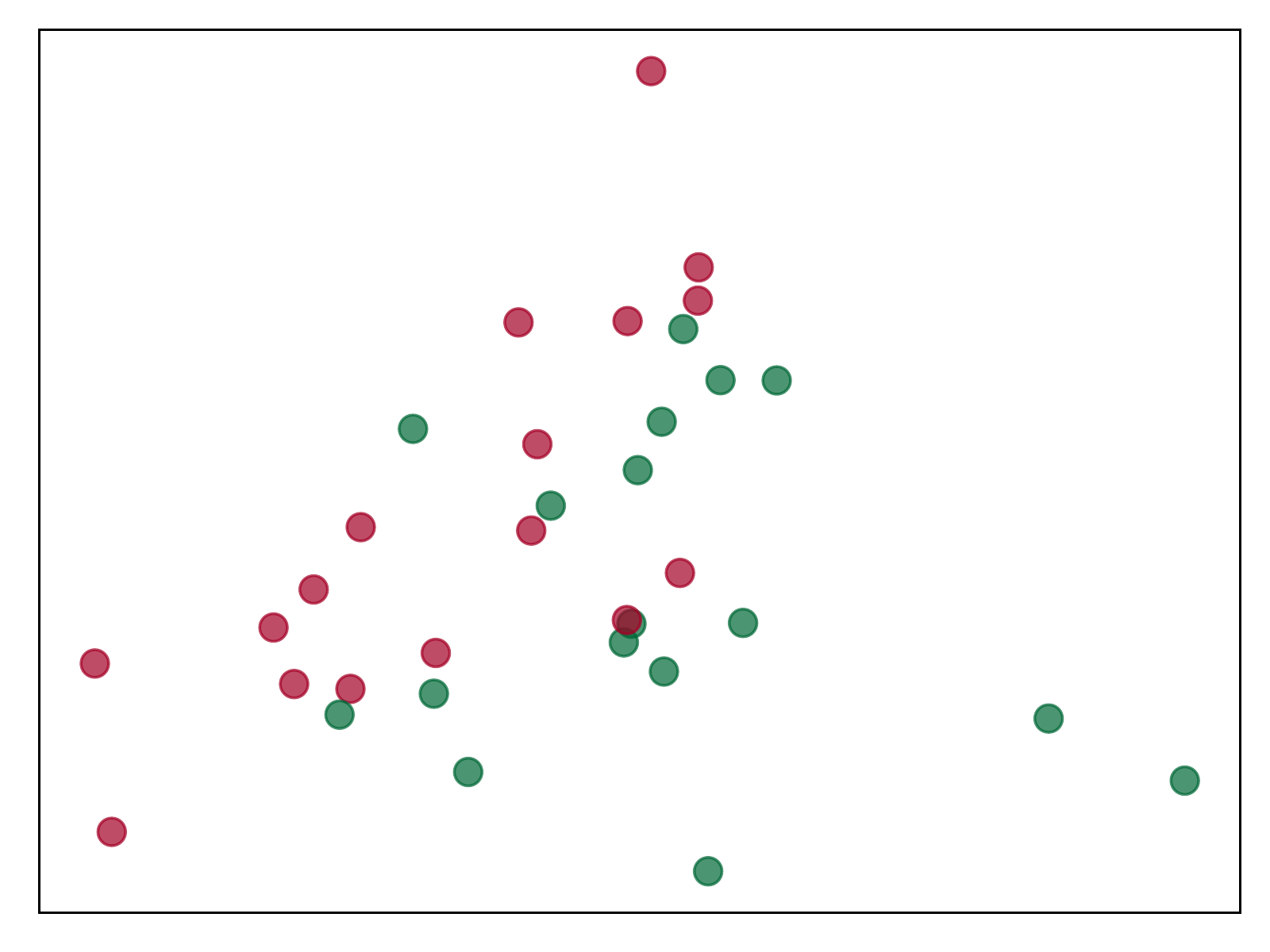}
            \caption{Original graph}\label{fig:emb_orig}
        \end{subfigure}
        \begin{subfigure}[b]{.19\linewidth}
            \includegraphics[width=\textwidth]{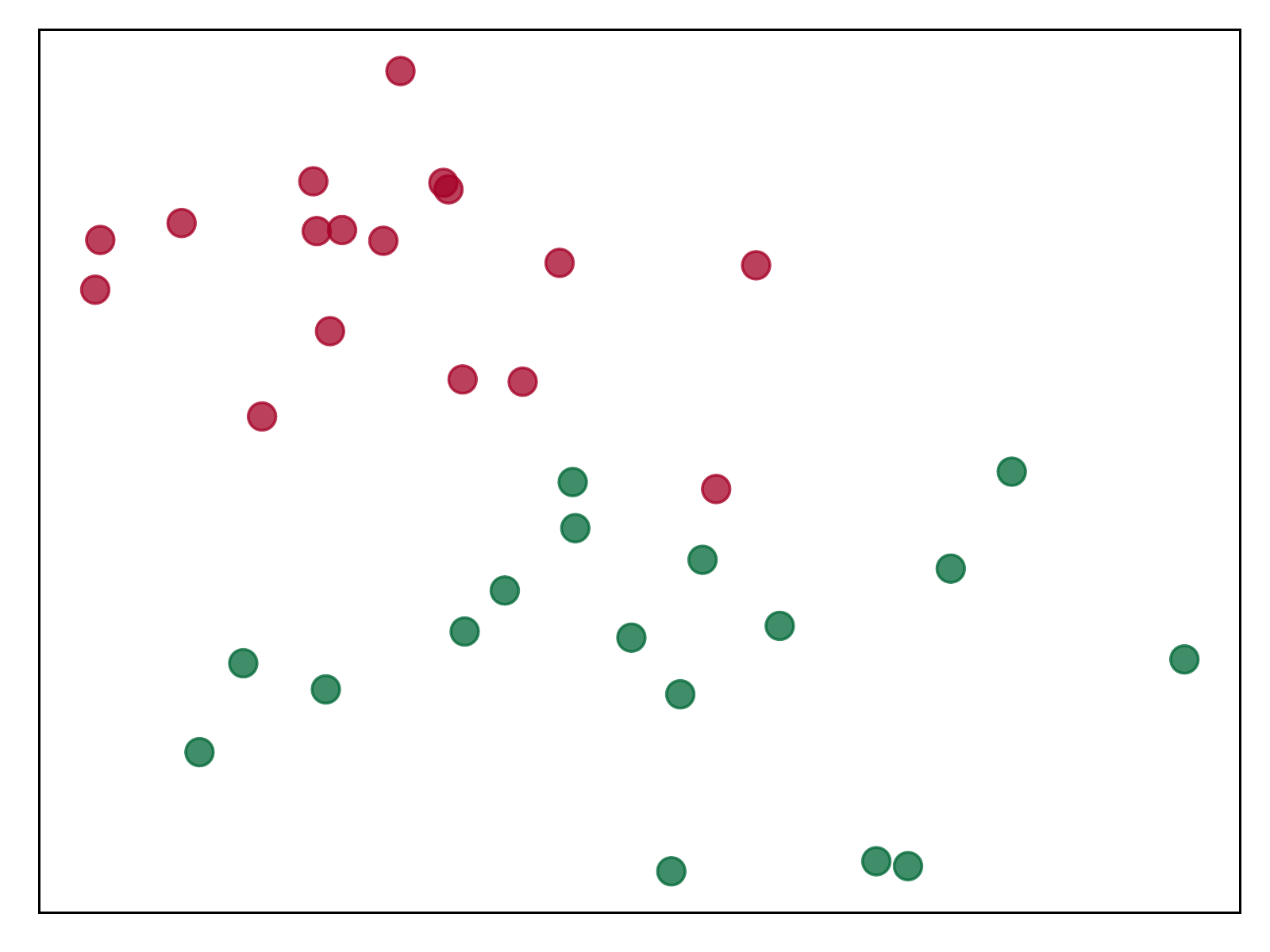}
            \caption{Modified graph}\label{fig:emb_mod}
        \end{subfigure}
        \begin{subfigure}[b]{.19\linewidth}
            \includegraphics[width=\textwidth]{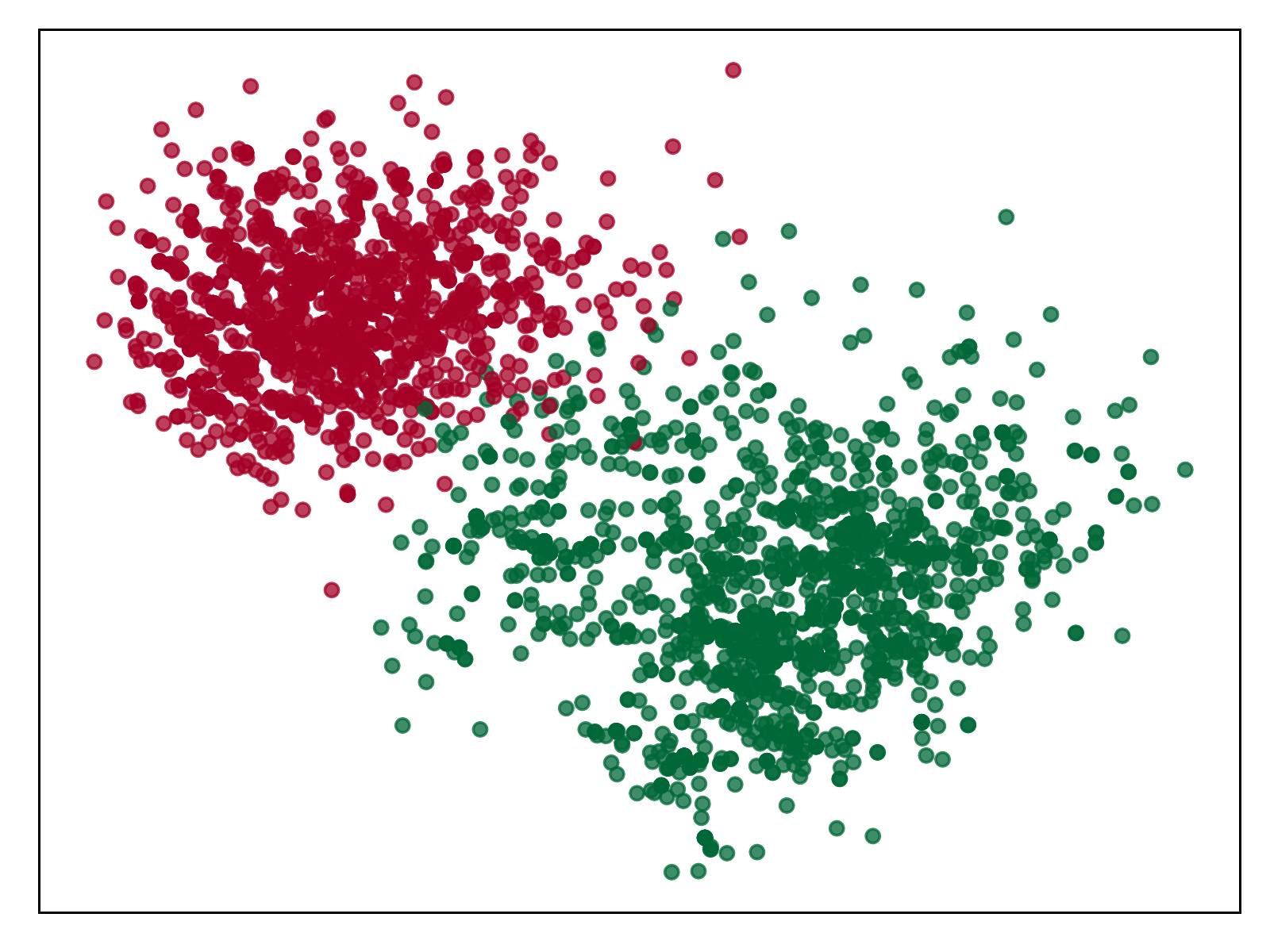}
            \caption{100 Mod. graph }\label{fig:emb_100}
        \end{subfigure}
        \begin{subfigure}[b]{.19\linewidth}
            \includegraphics[width=\textwidth]{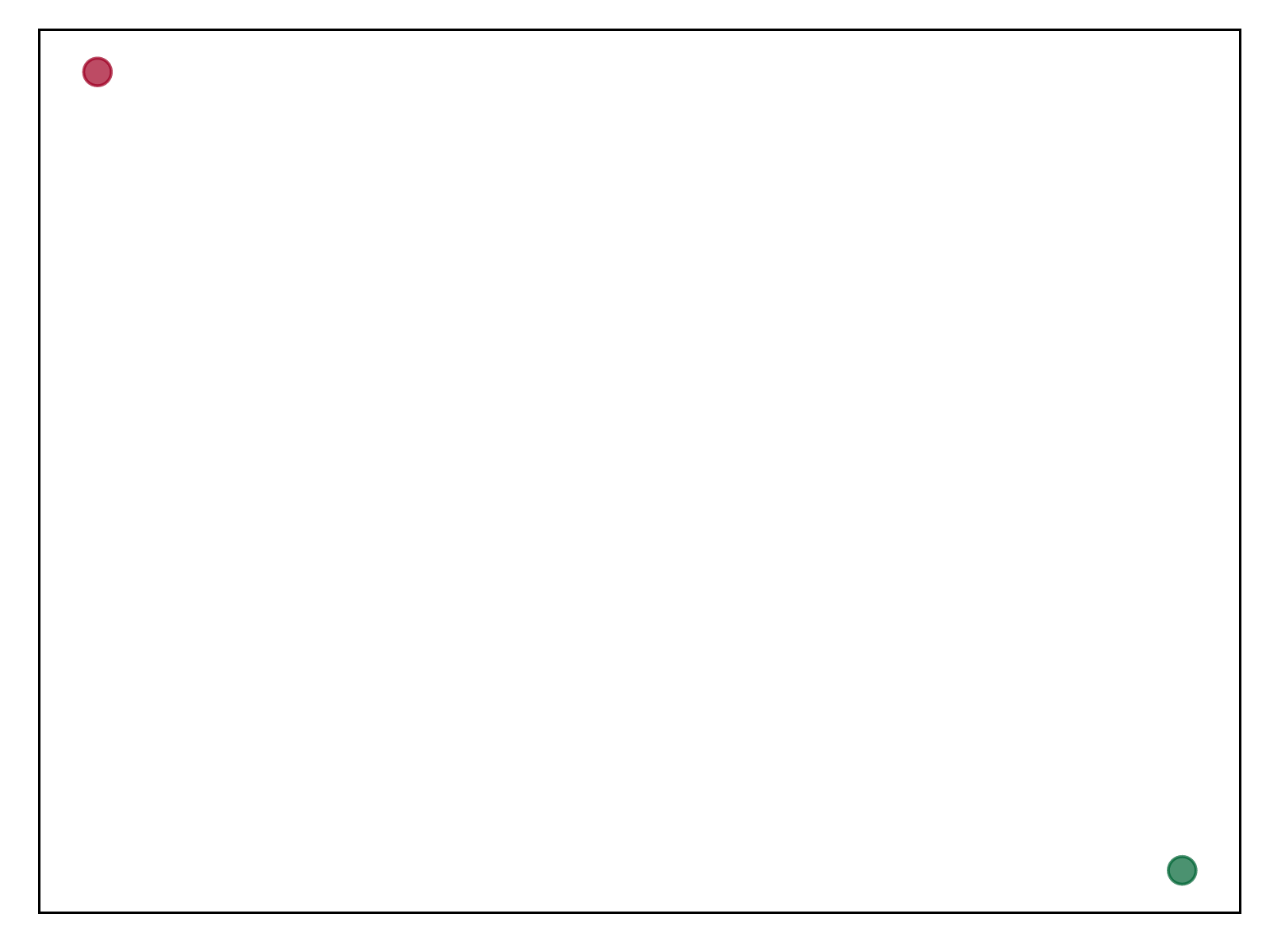}
            \caption{Ideal graph}\label{fig:emb_ideal}
        \end{subfigure}
    \caption{Embeddings after one \gcn layer. (c) and (e) show that augmentation can produce more clear decision boundaries between red and green nodes, compared to raw features (a), and naive \gcn on raw features (b).  (e) shows the effortless classification possible in the ideal graph scenario, where all same-class nodes have the same embedding.}\label{fig:emb}
\end{figure*}

\subsection{\methodshared's sensitivity to supervision}

As previously mentioned in Section \ref{sec:experiment},  which we showed that \methodshared is especially powerful under weak supervision, in Figure \ref{fig:super_full} we detail the sensitivity to supervision for each GNN combined with \methodtwo and \method. We can observe a larger separation between original (dotted) and \methodshared (solid) in each plot when training samples decrease, indicating larger performance gain under weak supervision. In most settings, test F1 score of \methodtwo and \methodshared training with 35 training nodes is at par or better than baseline training with 2 times as many (70) nodes. This suggest that \methodshared is an especially appealing option in cases of weak supervision, like in anomaly detection and other imbalanced learning settings.

\subsection{Embedding Visualization}
\label{}

To further illustrate the case for graph data augmentation in addition to Fig. \ref{fig:tiny} and Section \ref{sec:edgemanip_theory}, we plot the features and embeddings to understand how edge manipulation can contribute to lower-effort (and in the fully class-homophilic/ideal scenario, effortless) classification. In Figure \ref{fig:emb}, we randomly initialize 2-D node features for the Zachary’s Karate Club graph from normal distribution $\mathcal{N}(0, 1)$ (Figure \ref{fig:emb_raw}), and show the results of applying a single graph convolution layer (Eq.~\ref{eq:gcn_layer}) with randomly initialized weights. When the input graph is ``ideal'' (all intra-class edges exist, and no inter-class edges exist) all same-class points project to the same embedding, making discriminating the classes trivial (Figure \ref{fig:emb_ideal}). It is obvious that the original feature and embeddings (Figure \ref{fig:emb_orig}) are harder for a classifier to separate than Figure \ref{fig:emb_mod}, which the graph was modified by adding (removing) intra(inter)-class edges (simulating the scenario which \methodtwo achieves). Figure \ref{fig:emb_100} shows superimposed points resulting from 100 different modifications of the graph (simulating \method), illustrating the much clearer decision boundary between red-class and green-class points.



\end{document}